\PassOptionsToPackage{sort}{natbib}

\documentclass[preprint,12pt]{elsarticle}

\usepackage{amssymb}
\usepackage{amsmath}

\usepackage[x11names, rgb]{xcolor}
\usepackage[table]{xcolor} 
\usepackage{multirow}      
\usepackage{hyperref}
\usepackage{xcolor}
\usepackage{ulem}
\usepackage{wrapfig}

\newcommand{\revision}[2]{{#2}}

\journal{Elsevier}

\begin{document}

\begin{frontmatter}



\title{Implicit Neural Field-Based Process Planning for Multi-Axis Manufacturing: Direct Control over Collision Avoidance and Toolpath Geometry}



\author[labela1]{Neelotpal Dutta \fnref{fn1}} 
\author[labela1]{Tianyu Zhang \fnref{fn1}} 
\author[labela1]{Tao Liu} 
\author[labela1]{Yongxue Chen} 
\author[labela1]{Charlie C.L. Wang \corref{cor1}} 

\affiliation[labela1]{organization={Department of Mechanical and Aerospace Engineering, The University of Manchester},
            city={Manchester},
            postcode={M13 9PL}, 
            country={United Kingdom}}

\cortext[cor1]{Corresponding Author; Email: charlie.wang@manchester.ac.uk}
\fntext[fn1]{Joint First Authors}

\begin{abstract}
Existing curved-layer-based process planning methods for multi-axis manufacturing address collisions only indirectly and generate toolpaths in a post-processing step, leaving toolpath geometry uncontrolled during optimization. We present an implicit neural field-based framework for multi-axis process planning that overcomes these limitations by embedding both layer generation and toolpath design within a single differentiable pipeline. Using sinusoidally activated neural networks to represent layers and toolpaths as implicit fields, our method enables direct evaluation of field values and derivatives at any spatial point, thereby allowing explicit collision avoidance and joint optimization of manufacturing layers and toolpaths. We further investigate how network hyperparameters and objective definitions influence singularity behavior and topology transitions, offering built-in mechanisms for regularization and stability control. The proposed approach is demonstrated on examples in both additive and subtractive manufacturing, validating its generality and effectiveness.
\end{abstract}




\end{frontmatter}



\section{Introduction}

\subsection{Motivation}
Additive Manufacturing (AM), also known as 3D printing, involved building up material as planar layers in selected regions to achieve a target shape, typically requiring only motion in two synchronized degrees-of-freedom (DoFs). Recently, there has been growing interest in introducing additional DoFs, enabling multi-axis material deposition. Several benefits of multi-axis 3D printing have been identified, including support-free fabrication \cite{dai_support-free_2018, zhang_zhangty019s3_deformfdm_2025, liu_neural_2024} and improved surface finish \cite{zhang_zhangty019s3_deformfdm_2025, etienne_curvislicer_2019} (see \cite{yao_comparative_2024} for a comprehensive review). A key advantage of multi-axis printing is its ability to account for the anisotropic properties of materials, enabling enhanced strength in processes such as fused-deposition modelling (FDM) \cite{fang_reinforced_2020, li_vector_2022, zhang_zhangty019s3_deformfdm_2025, liu_neural_2024}, as well as in printing reinforcing materials like continuous carbon fibers along desired directions \cite{fang_exceptional_2024, zhang_toolpath_2025}. However, these capabilities come at the cost of increased planning complexity. Challenges arise from material and process constraints (e.g., maximum overhang angles or path alignment), hardware limitations (e.g., feasible deposition ranges), or their combination (e.g., collision avoidance while meeting design requirements). In general practices, requirements such as support-free printing or directional alignment are usually treated as functional objectives, while issues such as collision avoidance 
are often seen as constraints. In this work, we treat all these requirements uniformly as objectives to be jointly optimized within a computational framework.

\begin{figure}[!t]
    \centering
    \includegraphics[width=1.0\linewidth]{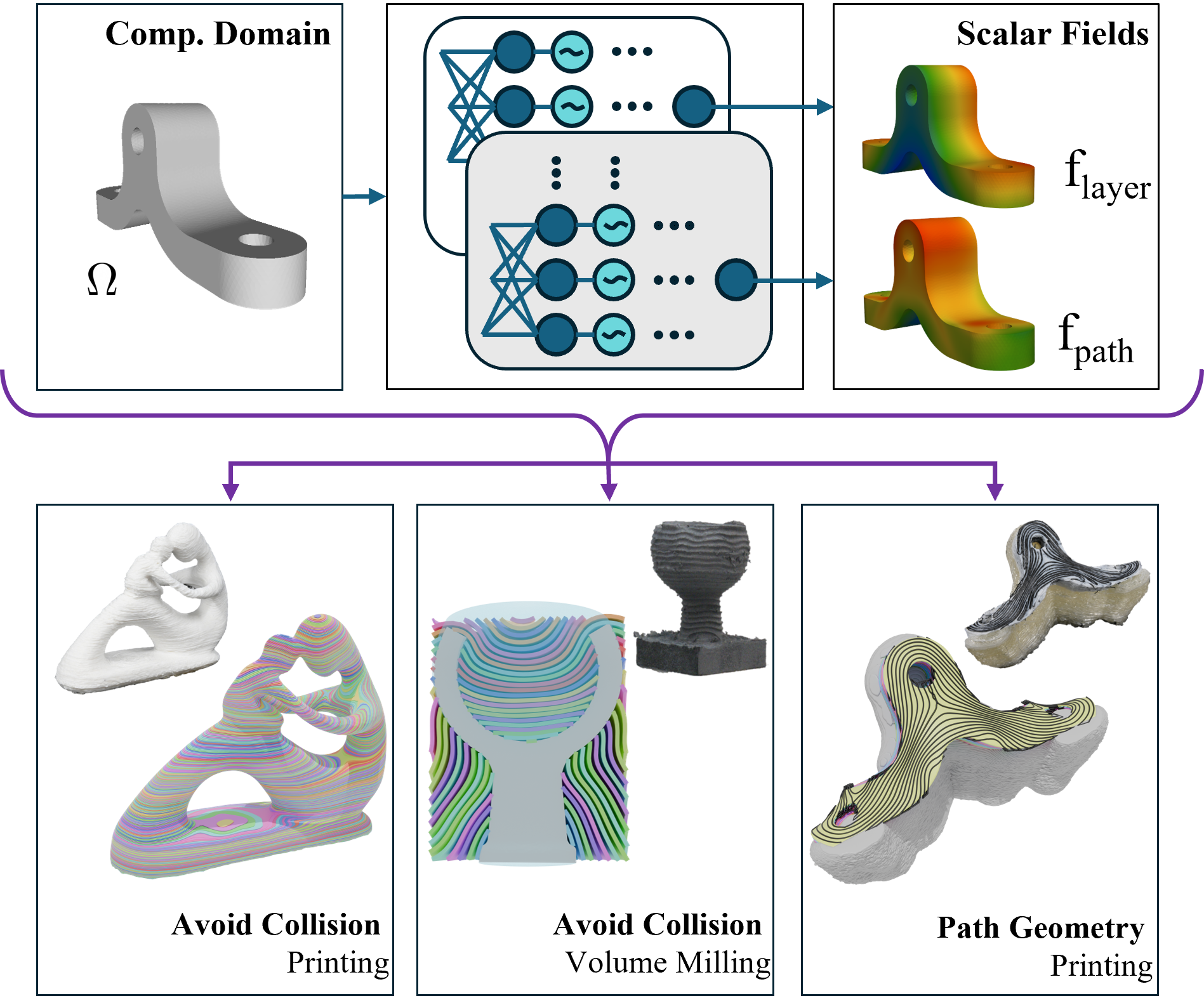}
    \put(-377,204){\scriptsize \color{black}(a)}
    \put(-263,204){\scriptsize \color{black}(b)}
    \put(-118,204){\scriptsize \color{black}(c)}
    \put(-376,8){\scriptsize \color{black}(d)}
    \put(-253,8){\scriptsize \color{black}(e)}
    \put(-128,8){\scriptsize \color{black}(f)}
    \caption{This figure summarizes our work. In the computational domain defined by an input model (a), we employ sinusoidally-activated neural networks (b) to represent implicit scalar-field functions (c), where the level sets of these scalar fields form the basis for generating layers and toolpaths. In our approach, collision-avoidance is directly enforced during layer generation to ensure manufacturability in multi-axis systems. Similarly, toolpath geometry can be simultaneously controlled during layer generation to accommodate applications like continuous carbon-fiber deposition. In (d), we demonstrate support-free fabrication of the Fertility model with integrated collision avoidance. The same framework can be generalized to support 5-axis rough milling -- e.g., the Cup model as shown in (e). Finally, we illustrate the effectiveness of our method in mechanical reinforcement by spatially deposited continuous carbon fiber in (f), enabled by toolpath-level geometry control.}
    \label{fig:teaser}
\end{figure}
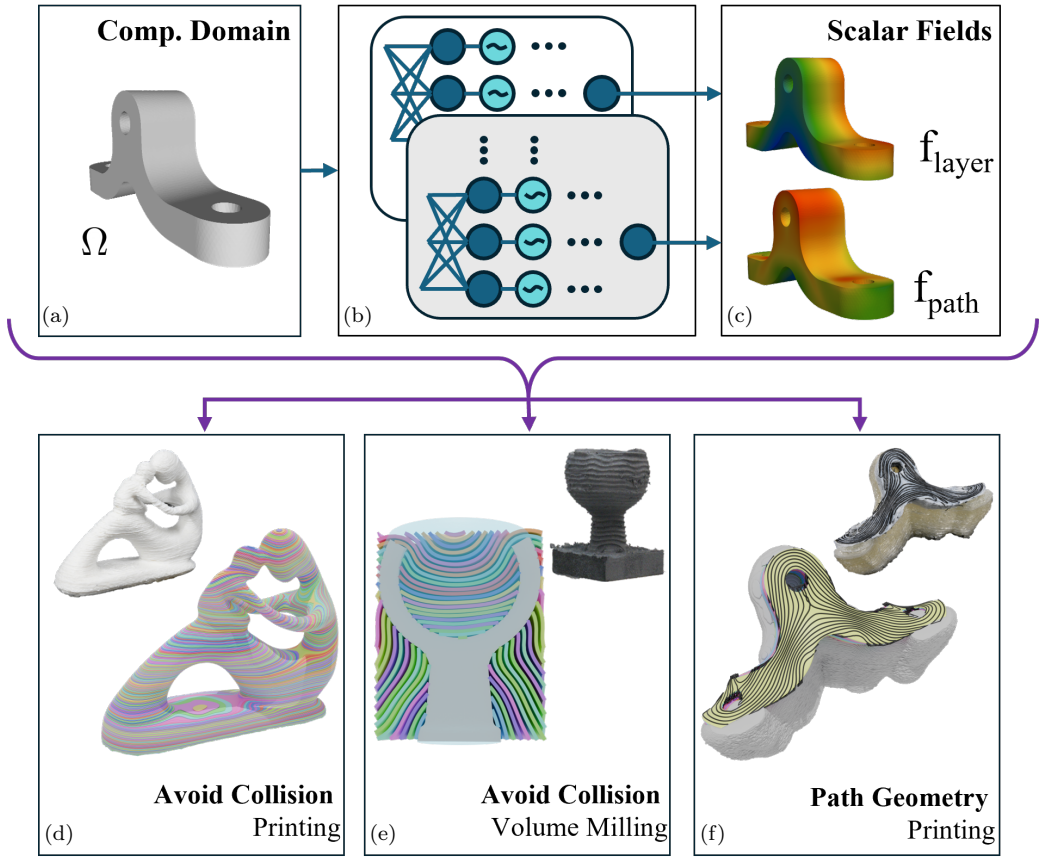

Several recent studies have addressed these challenges. Etienne et al. \cite{etienne_curvislicer_2019} introduced slicing in a deformed space to mitigate the staircase effect, requiring full three-axis motion rather than the conventional 2.5D planar process. This deformation-based concept was further advanced in the $S^3$-slicer \cite{zhang_zhangty019s3_deformfdm_2025}, which employed five-axis motion and incorporated additional objectives such as strength-reinforcement based alignment and support-free printing. Subsequently, Liu et al. \cite{liu_neural_2024} extended this approach using a neural network representation of the deformation field via quaternions and scalings, improving compliance with multiple objectives. In these studies, slicing is derived from a scalar field indirectly obtained through deformation. In contrast, other works \cite{fang_reinforced_2020, li_vector_2022} directly control the gradients of a scalar field to generate deposition layers. A common characteristic of all these methods is their reliance on discrete volumetric structures for optimization, which necessitates indirect estimation of derivatives. Moreover, collision avoidance is typically handled indirectly by limiting local surface curvature. As we show later (e.g., Fig.~\ref{fig:collision_fertility_2}, \ref{fig:results_clip_layer}), this strategy captures local collisions but fails to detect global collisions interfering parts far from the deposition point. Even recent approaches \cite{liu_neural_2025} that provide greater control over manufacturing constraints still address collision indirectly through curvature regulation.

A major requirement for advancing 3D printing beyond prototyping is achieving mechanical robustness in terms of stiffness, strength, and failure resistance. Multi-axis motion facilitates this by depositing material along directions tangential to principal stress fields \cite{fang_reinforced_2020, li_vector_2022, zhang_zhangty019s3_deformfdm_2025, liu_neural_2024}. These properties can be further enhanced through the inclusion of reinforcing materials such as continuous carbon fibers. Various methods have been proposed to optimize fiber orientations, including level-set-based approaches \cite{brampton_new_2015, fernandez_optimal_2019, chen_field-based_2022}. However, they typically compute on planar layers, which limits the achievable strength compared to spatial alignment via multi-axis motion \cite{fang_exceptional_2024}. Multi-axis fiber placement methods \cite{fang_exceptional_2024, zhang_toolpath_2025} generally employ field-based layer generation followed by a separate discrete pipeline for fiber toolpath computation. These methods lack of the direct control over toolpath geometry (e.g., turning angle and spacing) during optimization, and the co-adaptation between layers and fiber paths has not been achieved.

To address these gaps, we propose an implicit pipeline for multi-axis manufacturing processes that enables (1) direct optimization of layers for collision avoidance, (2) explicit control of toolpath geometry, and (3) co-optimization of layers and toolpaths (see Fig.~\ref{fig:teaser}). We employ a sinusoidally activated neural network \cite{sitzmann_implicit_2020} to represent implicit fields, enabling the expression of complex geometric and physical relationships. This implicit formulation allows for direct evaluation of the field and its derivatives at any location in the computational domain, eliminating the need for discrete mesh-based approximations during optimization.

Given the relative intricacies of training sinusoidally activated networks \cite{parascandolo_taming_2016, sitzmann_implicit_2020, yeom_fast_2024}, we provide an analysis of how network hyper-parameters and objective functions can be controlled to implicitly regularize the function space, enabling the network to maintain smooth representations of the fields with reduced local oscillations while still satisfying task-specific requirements. The discussion is further extended to topics such as frequency, topology change, and singularities, enabling users to fully utilize the pipeline according to their needs. We demonstrate that the pipeline can accommodate a wide range of objectives and provide examples to verify its effectiveness.

Finally, to highlight its generality, we extend the pipeline beyond additive manufacturing to collision avoidance in multi-axis rough milling. 
The effectiveness of our approach is validated through physical fabrication and corresponding performance evaluations.

\subsection{Contributions}
The technical contributions of our work are summarized as follows.
\begin{itemize}
    \item A universal implicit scalar-field optimization framework that enables direct control over both layer and toolpath generation in multi-axis manufacturing.
    
    \item A formulation that allows explicit collision avoidance during field-based optimization.
    
    \item Formulating toolpath-level requirements as objectives for joint optimization of layers and toolpaths.

    \item An analytical study of how sinusoidal network architectures and loss definitions influence singularities and topology changes, providing  regularization for controlling the geometry of layers and toolpaths.
\end{itemize}
The effectiveness of our approach has been verified on a variety of examples in both numerical and physical experiments. Examples for both multi-axis 3D printing and CNC milling are employed in these experimental tests. 

\section{Related Works}
\subsection{Multi-Axis Additive Manufacturing}
Traditionally, AM has relied on the deposition of material in successive planar layers \cite{gibson_additive_2015}. Various AM techniques exist, including FDM, Digital Light Processing (DLP), and Selective Laser Sintering (SLS) \cite{gibson_additive_2015}. Despite differences in their underlying process mechanisms, these methods predominantly employ a planar (2.5D), layer-by-layer material accumulation strategy. In this work, unless otherwise specified, the term additive manufacturing or 3D printing specifically refers to FDM technology, which remains the most widely adopted AM method \cite{fang_reinforced_2020}.

In recent years, moving beyond planar deposition has attracted growing attention due to the advantages offered by non-planar material addition. Dai et al.~\cite{dai_support-free_2018} demonstrated support-free printing by introducing multi-axis motion in FDM, while Fang et al. \cite{fang_reinforced_2020} showed that exploiting the anisotropic properties of filaments through non-planar deposition can substantially enhance the mechanical strength of printed parts. Not all forms of non-planar deposition require motion beyond three axes. For example, Etienne et al.~\cite{etienne_curvislicer_2019} utilized 3-axis motion combined with variable deposition rates to improve surface quality. Their method adopted a deformation-based strategy to convert planar slices into non-planar layers for deposition. Building upon this concept, Zhang et al.~\cite{zhang_zhangty019s3_deformfdm_2025} developed the $S^3$-\textit{Slicer}, which uses deformation-based optimization to address multiple objectives such as support-free printing, strength reinforcement, and surface-quality improvement. This approach was further extended in \textit{Neural Slicer} \cite{liu_neural_2024}, which employs an implicit neural field to represent deformation, thereby overcoming several limitations of mesh-based representations. These methods extract deposition layers as iso-scalar surfaces (level sets) of a scalar field derived from a height field in the deformed space.

In a related direction, Fang et al.’s Reinforced-FDM \cite{fang_reinforced_2020} aligns layers with stress fields by leveraging gradients of scalar functions, and Li et al.~\cite{li_vector_2022} incorporated multiple functional objectives into the slicing process using a similar formulation. While these methods embed multiple objectives into the optimization loop, geometric properties such as curvature are typically controlled only indirectly (e.g., through Laplacian smoothing). A major limitation of these approaches is the lack of explicit collision handling during optimization. Most rely on indirect measures such as curvature reduction to mitigate local collisions or employ post-processing steps for correction.

Beyond differences in optimization formulation, these approaches can also be distinguished by their underlying computational data structures. Dai et al.~\cite{dai_support-free_2018} employed a voxel-based representation, whereas most subsequent methods \cite{etienne_curvislicer_2019, fang_reinforced_2020, li_vector_2022, zhang_zhangty019s3_deformfdm_2025} utilized tetrahedral meshes. Such discrete representations store field values only at fixed elements (vertices, faces, or cells), requiring interpolation for arbitrary points and approximation for derivative computation. The recently developed Neural Slicer \cite{liu_neural_2024} combines a continuous neural field with a tetrahedral cage, while Liu et al.~\cite{liu_neural_2025} employed a fully continuous neural-field representation. Nevertheless, none of these approaches directly incorporate global collision constraints. Moreover, they do not provide explicit control over toolpath geometry, and the layer-slicing process remains decoupled from geometric constraints at the toolpath level.

More recently, Chermain et al.~\cite{chermain_atomizer_2025} proposed a different approach that moves away from conventional layer-based toolpath generation. Their Atomizer method controls deposition at the level of individual points, marking a significant departure from traditional layer-based slicing. However, its current application remains limited to surface-quality optimization for parts with fixed orientation.

\subsection{Collision Avoidance in Multi-axis Manufacturing}

While the introduction of multi-axis motion in additive manufacturing (AM) is relatively recent, subtractive manufacturing processes such as CNC milling have a long history of employing multi-axis motion. Consequently, the problem of collision avoidance associated with non-planar motion has been extensively studied in prior works \cite{cho_generation_1997, balasubramaniam_automatic_2001, zhang_optimal_2016, liang_review_2021, zaragoza_chichell_collision-free_2024}. \revision{}{Very recently, Chichell et al. \cite{chichell2025evolution} proposed a method that can simultaneously optimize the milling path as well as tool configuration while ensuring collision-free machining.} However, most of these studies focus on the final-finishing stage, where the surface to be machined is fixed. In contrast, rough milling, which operates on the material volume similar to AM, has largely remained restricted to planar milling due to the relative simplicity of planning \cite{he_geodesic_2021}. Mahdavi et al.~\cite{mahdavi-amiri_vdac_2020} proposed an accessible-volume milling strategy that automatically partitions the machining volume into regions with fixed accessibility directions. Material removal in their approach still relies on 2.5D motion, and it has not been demonstrated how additional constraints on tool orientation can be incorporated. He et al.~\cite{he_geodesic_2021} introduced a non-planar rough milling strategy by growing a scalar field outward from the convex hull of the part, thereby inherently improving accessibility. Similarly, Dutta et al.~\cite{dutta_vector_2023} employed a vector-field formulation and presented an anchor-based strategy to flexibly modify the resulting scalar fields for enhanced accessibility. Although these methods do not explicitly address tool–collision avoidance during layer generation (or volume-peeling), they provide valuable insights into how gradients of scalar fields can be modified to improve accessibility.

In the domain of multi-axis additive manufacturing, collision avoidance has predominantly been addressed indirectly through curvature-based control strategies \cite{fang_reinforced_2020, li_vector_2022, zhang_zhangty019s3_deformfdm_2025, liu_neural_2024, liu_neural_2025} and/or within the post-processing phase \cite{fang_reinforced_2020, li_vector_2022, zhang_zhangty019s3_deformfdm_2025, liu_neural_2024, liu_neural_2025, chermain_atomizer_2025, kubalak_simultaneous_2025}. However, as we have previously noted and further demonstrate later, curvature-based methods can only ensure local collision resolution. Post-processing approaches, on the other hand, require modifications to the tool orientations that were originally optimized for specific functional objectives, without any guarantee of obtaining a globally valid collision-free configuration.

Lau et al. \cite{lau_partition-based_2023} proposed a partition-based strategy to select a collision-free sequence of iso-scalar surfaces. Their method, however, requires an input scalar field that is not optimized during the process and assumes that the parts are of genus-0. In contrast, Jayakody et al. \cite{jayakody_topological_2024} introduced a topological analysis-based approach to generate a tool-orientation vector field for collision avoidance. Although the topology of the part provides useful guidance for determining potentially accessible orientations, collisions also depend on geometric properties, which cannot be fully captured by their method.

Kubalak et al. \cite{kubalak_simultaneous_2025} and Chermain et al. \cite{chermain_atomizer_2025} adopted an alternative strategy based on re-ordering rods and points to ensure accessibility. In their approaches, collision handling is performed as a separate step after toolpaths and orientations have already been determined through a preceding optimization cycle. The re-ordering procedure, while potentially effective, can be computationally demanding, often accounting for a substantial portion of the overall process time, and is best suited to scenarios where only a single iteration is required. Moreover, the discrete nature of such re-ordering makes it difficult to formulate in a differentiable manner, posing challenges for integration into gradient-based co-optimization frameworks.

Very recently, Qu et al. \cite{qu2025inf3dpimplicitneuralfields} introduced an approach using implicit neural fields to represent tool configurations, enabling adjustments of both deposition sequence and tool orientation to reduce tool collisions. However, their method does not incorporate global-collision avoidance directly during the initial field (or layer) generation stage.

\revision{}{The importance of incorporating accessibility considerations within the optimization process has been highlighted by \cite{mirzendehdel2020topology}, who introduced a convolution-based continuous \textit{inaccessibility measure field} in topology optimization to ensure manufacturability of the resulting designs. They also provide a brief review of prior work addressing accessibility constraints in topology optimization. However, the incorporation of continuous accessibility formulations into toolpath and motion planning for multi-axis manufacturing processes remains limited and requires further investigation. In related domains, signed-distance fields, extensively used in robotics \cite{oleynikova2016signed} and computer graphics \cite{macklin2020local}, demonstrate the effectiveness of implicit distance-based representations for efficient collision querying.}

Building upon this line of research, we propose a fundamentally different strategy. Specifically, our method integrates collision avoidance directly into the field generation stage, rather than addressing it as a post-processing step. This integration enables the simultaneous co-optimization of collision-free fields alongside other design and process requirements. As a result, our framework provides a unified formulation for achieving feasible, efficient, and practically deployable solutions that respect both geometric constraints and functional objectives.

\subsection{Planning for Continuous Fibers}
Significant research attention has been directed toward the optimization of fiber orientations in fiber-reinforced composites, motivated by their enhanced strength and stiffness along the fiber direction. Recent studies have focused on developing orientation strategies that maximize structural performance while ensuring manufacturable fiber paths \cite{brampton_new_2015, fernandez_optimal_2019, liu_stress_2024, wang_load-dependent_2021, chen_field-based_2022, luo_spatially_2023, kubalak_simultaneous_2025}. These methods primarily aim to align fiber with principal stress directions to exploit the superior mechanical properties along the fiber axis. Two commonly employed approaches include stress-line tracing \cite{liu_stress_2024, wang_load-dependent_2021} and level-set methods using scalar fields (periodic or non-periodic) aligned with principal stress directions \cite{brampton_new_2015, fernandez_optimal_2019, chen_field-based_2022, xu_concurrent_2022, ren_concurrent_2024}. 

Stress-line tracing approaches are particularly suitable for single-step sequential processes, whereas level-set methods are more commonly applied in iterative optimization schemes. While alignment with principal stress directions is crucial for enhancing mechanical performance, the geometry and distribution of fibers are also important from a manufacturability perspective. Sharp turns, for instance, not only pose challenges for manufacturing hardware but may also introduce defects \cite{brooks_manufacturing_2018, qu_placement_2021, halbritter_leveraging_2023}. For example, Xiao et al. \cite{xiao_field-based_2025} addressed this issue by controlling the geodesic curvature of planned fiber paths. Despite these advances, most existing methods have been applied primarily to planar surfaces or relatively simple geometries.

Fang et al. \cite{fang_exceptional_2024} extended stress-line tracing to non-planar surfaces, demonstrating superior performance relative to planar approaches. However, their method did not explicitly control the spacing between fibres, an aspect later addressed by the high-density toolpath method of Zhang et al. \cite{zhang_toolpath_2025}. While this approach significantly improves mechanical performance, its reliance on discontinuous segments introduces sharp turns and numerous tool jumps within layers, which can compromise manufacturability and partially offset the benefits.

Therefore, there remains a need for methods that optimize both fiber distribution and geometric properties, such as geodesic curvature, while maintaining alignment with the desired directional field on non-planar surfaces. Liu et al. \cite{liu_neural_2025} co-optimize both the structural layout and the layers based on path orientation as a new framework for topology optimization of structures reinforced by continuous fibers (ref.~\cite{xu_concurrent_2022, luo_spatially_2023, ren_concurrent_2024}).  However, their approach only controls layer geometry and relies on post-processing, similar to Zhang et al. \cite{zhang_toolpath_2025}, for toolpath generation. Such post-processing makes the layer-generation stage largely agnostic to toolpath-level geometric constraints.

Furthermore, directly incorporating toolpaths during the layer-generation stage requires attention to toolpath-topology and direct toolpath-geometric control. Fang et al. \cite{fang_exceptional_2024} demonstrated that an indirect control on the path-scalar field as done in Chen et al. \cite{chen_field-based_2022}, can lead to deviation of the paths from the desired direction field. In iterative methods such as gradient-based optimization, the influence of the initial design \cite{fernandez_optimal_2019, ren_concurrent_2024} should also be carefully considered. As a result, a differentiable method that directly controls toolpath geometry can enable seamless integration with broader design optimization frameworks and also maintain its generality, avoiding restriction to specific applications.

\subsection{Implicit Representation of Fields}
Deep neural networks are well known for their universal function approximation property \cite{liang_why_2017}, and have therefore been widely employed to represent complex signals or fields, such as geometric objects or even entire object classes \cite{liao_deep_2018, gropp_implicit_2020, liu_learning_2022}. Sitzmann et al.~\cite{sitzmann_implicit_2020} recently demonstrated that networks using periodic (sinusoidal) activation functions, referred to as SIRENs, offer significant advantages in representing complex signals and their derivatives. While sinusoidally activated networks provide clear representational benefits, they are known to be sensitive to initialization and hyperparameter choices, and may become trapped in local minima \cite{parascandolo_taming_2016, yeom_fast_2024, sitzmann_implicit_2020}. Subsequent studies have shown that strategies such as frequency tuning and weight scaling can further improve their stability and accuracy \cite{parascandolo_taming_2016}. To address these challenges, Sitzmann et al.~\cite{sitzmann_implicit_2020} proposed a dedicated initialization scheme that preserves the activation characteristics consistently across layers and within their derivatives, demonstrating its effectiveness across a wide range of applications. Motivated by these findings, we adopt the SIREN-based representation for all fields considered in the present work. In addition, we systematically study the influence of key hyperparameters on our results and provide practical guidance for their selection.

\section{Overview}

\subsection{Representation and Pipeline}

\begin{figure}[t]
    \centering
    \includegraphics[width=0.95\linewidth]{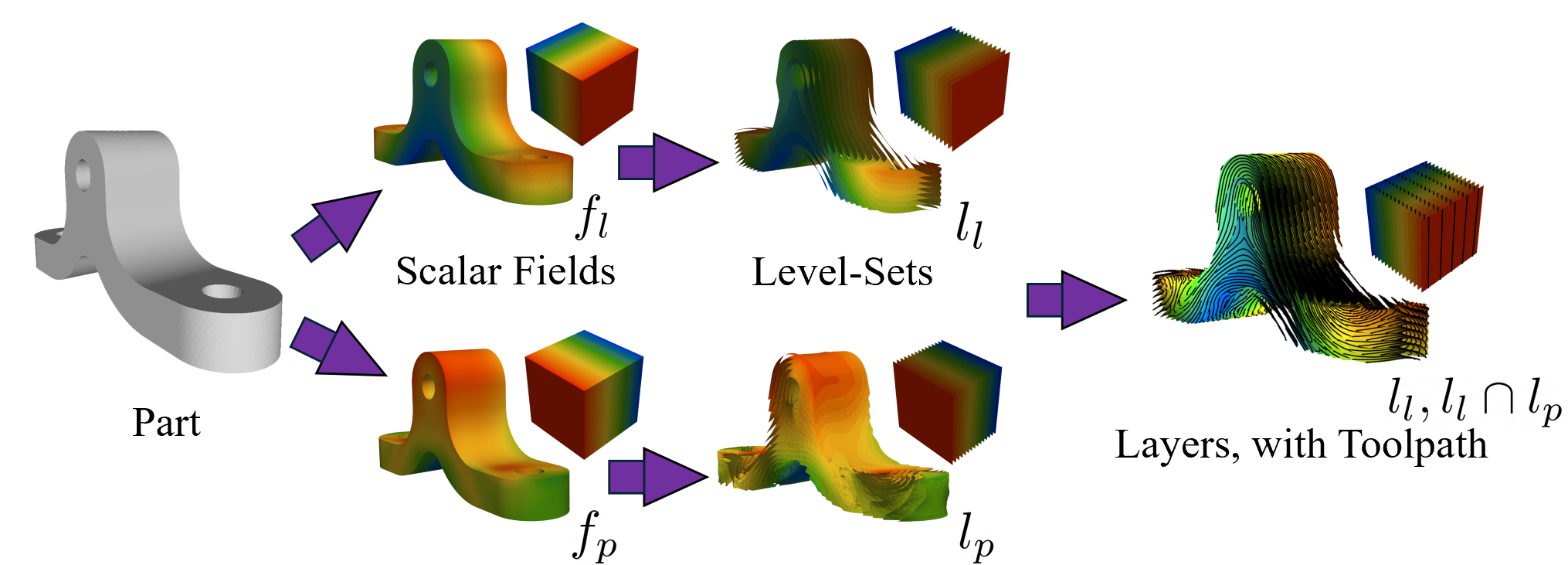}
    \put(-377,44){\footnotesize \color{black}(a)}
    \put(-298,121){\footnotesize \color{black}(b)}
    \put(-206,121){\footnotesize \color{black}(d)}
    \put(-298,2){\footnotesize \color{black}(c)}
    \put(-206,2){\footnotesize \color{black}(e)}
    \put(-103,44){\footnotesize \color{black}(f)}
\caption{The figure illustrates our unified representation of both layers and toolpaths in multi-axis manufacturing for a part (a). We define (b) a layer scalar field $f_l$ and (c) a path scalar field $f_p$ over the part geometry. The corresponding level sets of these fields are denoted by $l_l$ and $l_p$, respectively (d, e). Layers are represented by the level set $l_l$, while toolpaths are obtained as contours formed by the intersection of $l_l$ and $l_p$ located on $l_l$ as shown in (f).
}\label{fig:represnetaiton}
\end{figure}
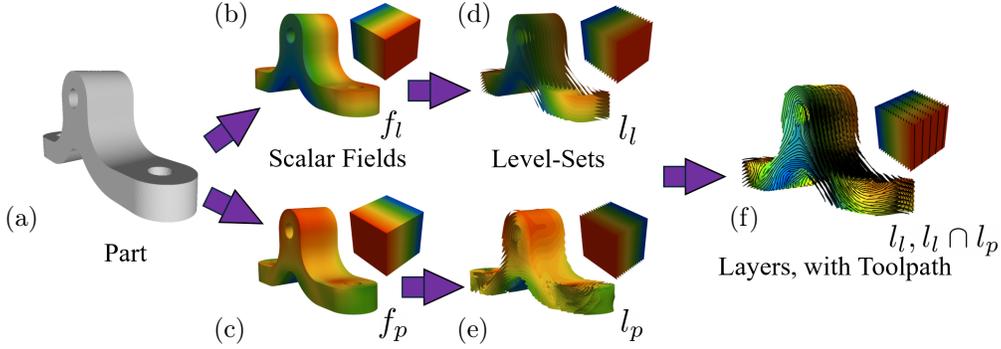

Considering a solid model $\Omega$ to be printed, we represent both the layers and the toolpaths within the layers through the level sets of scalar fields defined over the domain. Specifically, two scalar fields are introduced: the layer field $f_l(\mathbf{x})$ and the path field $f_p(\mathbf{x})$.

The level set of each field is defined as
\begin{equation}
l_i(c) = \{\mathbf{x} \mid f_i(\mathbf{x}) = c, \; \mathbf{x} \in \Omega\}, \quad i \in \{l, p\}.
\end{equation}
Both $f_l$ and $f_p$ are functions of the three spatial coordinates $\mathbf{x} = (x, y, z)$, and therefore their level sets correspond to surfaces, referred to here as iso-surfaces in three dimensions. The iso-surfaces of the field $f_l$ represent the layers for printing (or milling), while the iso-contours of $f_p$ on each layer define the toolpaths. Hence, the toolpaths are obtained as the intersection curves of the iso-surfaces of $f_l$ and $f_p$ (see Fig.~\ref{fig:represnetaiton} for an illustration).

At each point on a layer, the tool orientation is aligned with the surface normal of the corresponding iso-surface of $f_l$. Two neural networks are employed to implicitly represent the scalar fields $f_l$ and $f_p$. Thus, the general form of each field is expressed as $f(\mathbf{x}, \Theta)$, where $\Theta$ denotes the set of network parameters serving as the variables to be determined via self-learning (i.e., optimization). For simplicity, we often omit one or both arguments when their meaning is clear from the context.

The spatial derivatives of these fields, including the gradients ($\nabla f_l$, $\nabla f_p$) and second-derivatives ($\mathbf{H}_{f_l}$, $\mathbf{H}_{f_p}$), are readily obtained through automatic differentiation \cite{paszke_automatic_2017}. These quantities are then used to define various manufacturing and functional objectives. For example, collision avoidance is formulated as a function of $f_l$ and $\nabla f_l$ (Sec.~\ref{subsec:collision}), whereas control over the geodesic curvature of toolpaths depends on both the first and second derivatives of $f_p$ and $f_l$ (Sec.~\ref{subsec:curvatures}).

Each objective is expressed as a function (referred to as a loss function or simply loss), and the overall formulation is cast as a multi-objective optimization problem. The total loss is minimized to obtain the final representations of $f_l$ and $f_p$, using a stochastic gradient-based solver that leverages the differentiability of both the neural-field representations and the loss function definitions.

The basic computational pipeline can be summarized as follows:
\begin{enumerate}
\item Evaluate the scalar fields $f_l$ and $f_p$ and their spatial derivatives for the current network parameters $\Theta$;
\item Compute the individual loss functions and the resulting total loss function;
\item While the total and individual loss values remain outside the desired tolerance, update the parameters $\Theta$ of $f_l$ and $f_p$ and repeat from Step~1;
\item Extract the layers as the iso-surfaces of $f_l$, and obtain the toolpaths as iso-contours of $f_p$ on each layer.
\end{enumerate}

\begin{figure}
    \centering
    \includegraphics[width=1.0\linewidth]{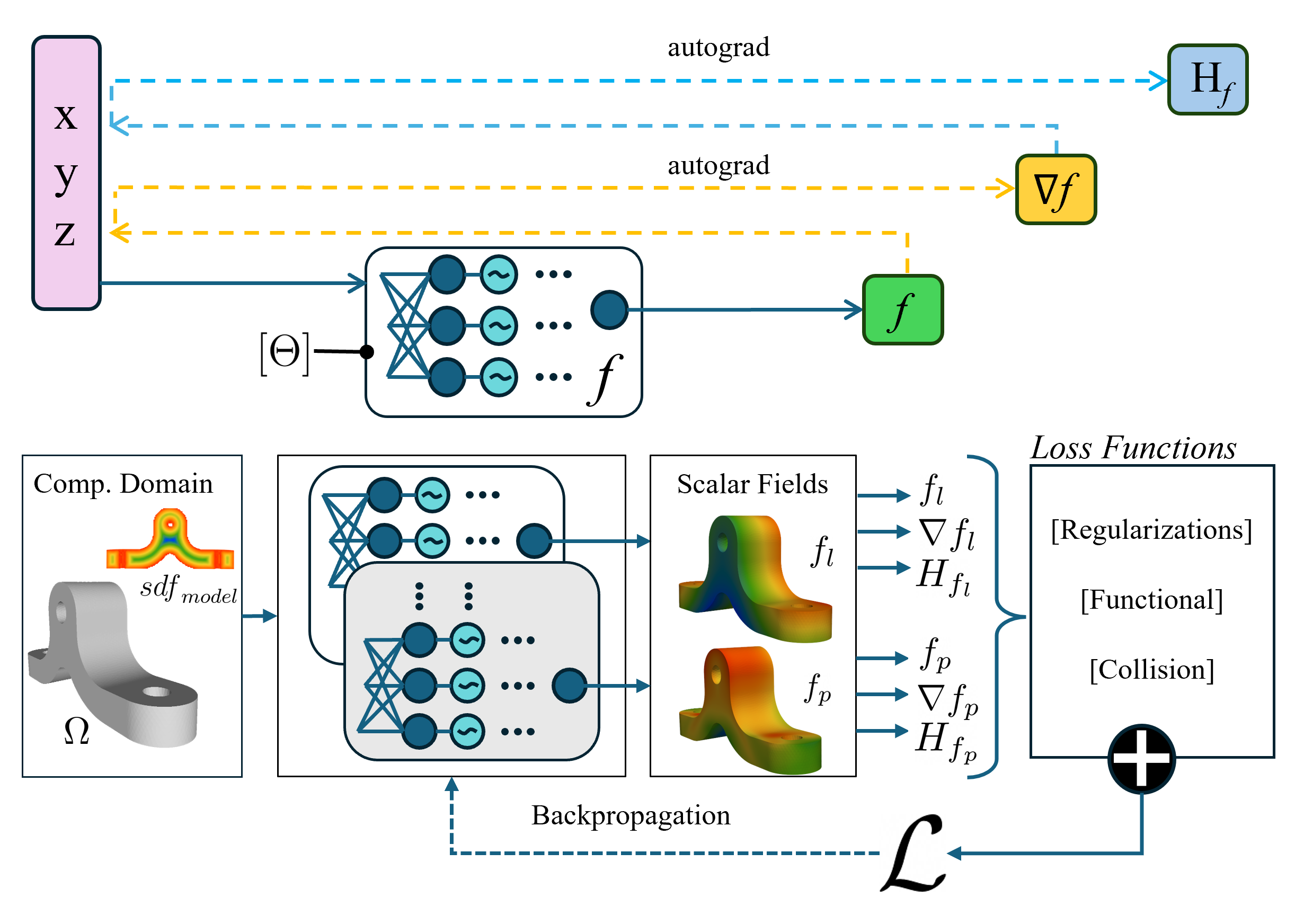}
    \put(-390,160){\footnotesize \color{black}(a)}
    \put(-390,15){\footnotesize \color{black}(b)}
    \vspace{-15pt}
    \caption{Pipeline of the algorithm for neural field-based process planning: (a) illustrates the structure of the neural network used for all field representations in our work. Each unit $f$ is a multilayer perceptron (MLP) with sinusoidal activation, commonly referred to as a SIREN \cite{sitzmann_implicit_2020}. The input to the network is a 3-dimensional vector representing a coordinate in $\mathbb{R}^3$ , and the output is a scalar value corresponding to the field at that location. The spatial derivatives such as the gradient ($\nabla f$) and the Hessian ($\mathbf{H}_f$) can be computed directly at any point. These quantities form the standard three-part output of our network: [$f, \nabla f, \mathbf{H}_f$]. $\Theta$ denotes the network parameters that are optimized during the process. We also hyper-parameterize the network depth, the frequency-scaling inside the activation sine functions, which are chosen differently for the layer, toolpath, and model networks. (b) illustrates the overall pipeline of our method. For a given part ($\Omega$), which itself can be represented as a signed distance field (${sdf}_{model}$), we employ two field networks corresponding to layers and toolpaths. From these networks, we obtain the field values and their derivatives, which are then used to define a set of requirements expressed as loss functions. The loss functions are categorized into regularization losses (e.g., curvature), functional losses (e.g., directional alignment), and collision losses. The combined sum of these individual losses constitutes the total loss, which is minimized to determine the network parameters representing the layers and toolpaths.}
    \label{fig:pipeline}
\end{figure}

A graphical overview of our computational pipeline is illustrated in Fig.~\ref{fig:pipeline}.
In addition, we employ another neural-network, denoted as $sdf_{model}$, for the implicit  representation of the parts as a \textit{signed distance field} (SDF). This field facilitates efficient point-membership queries, with negative distances assigned to points inside the object and positive distances to those outside.

Before proceeding further, we clarify a notational convention used throughout this work. Symbols written without curly braces (e.g., $\mathcal{D}$) denote continuous (sub-)domains in 3D space, whereas the corresponding symbols enclosed in curly braces (e.g., $\{\mathcal{D}\}$) denote discrete representations of the same domains, obtained by sampling points from them. Additionally, throughout this text, we use the terms optimization and learning to mean the optimization process of the fields to minimize the total loss.


\subsection{Gradients and Beyond}
\label{subsec:overview_derivatives}
\begin{figure}
    \centering
    \includegraphics[width=0.95\linewidth]{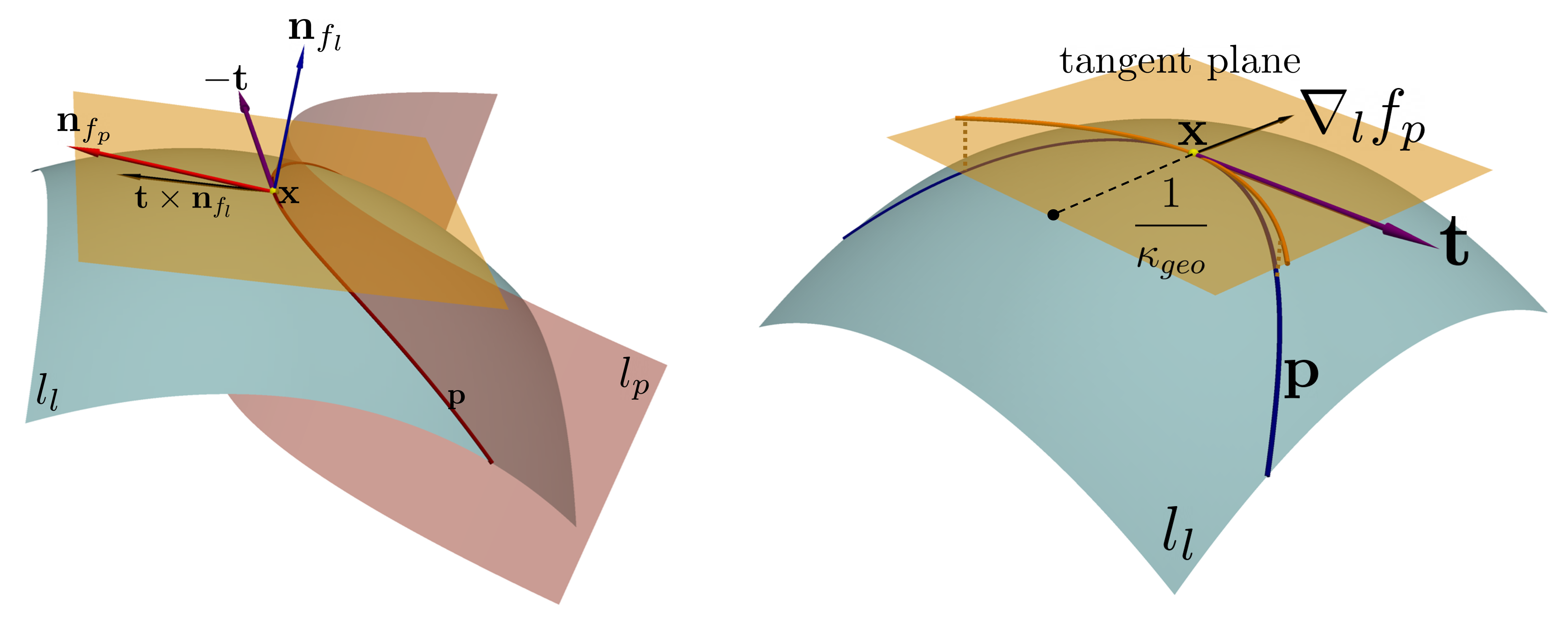}
    \put(-270, -8){\small \color{black}(a)}
    \put(-100, -8){\small \color{black}(b)}
    \caption{Illustration of the geometric meaning of spatial derivatives of scalar fields and their derived quantities.}
    \label{fig:derivativeGeometry}
\end{figure}

In the previous subsection, we referred to the use of derivatives of the field functions. Here, we introduce these derivatives and discuss their geometric interpretation. The spatial gradient of a scalar field at any point is normal to the iso-surface passing through that point. Accordingly, the unit normal to a layer surface at position $\mathbf{x}$ can be expressed as:
\begin{equation}
{\mathbf{n}}_{f_l}(\mathbf{x}) = \frac{\nabla f_l(\mathbf{x})}{\|\nabla f_l(\mathbf{x})\|}
\label{eqn:def_layer_normal}
\end{equation}
Throughout this work, the operator $\nabla$ denotes differentiation with respect to the spatial coordinates $\mathbf{x}$, unless stated otherwise. The norm $\|\nabla f_l(\mathbf{x})\|$ represents the local rate of change of the field and can therefore serve as a measure of the spacing between adjacent layers.

Similarly, we define the unit normal to the iso-surfaces of the path field $f_p$ as ${\mathbf{n}}_{f_p}$. 

Using these quantities, a Darboux frame (see Fig.~\ref{fig:derivativeGeometry}(a)) can be constructed at any point. The tangent vector to the toolpath curve $\mathbf{p}$ is defined as:
\begin{equation}
\mathbf{t} = \frac{d\mathbf{p}}{ds} = \frac{{\mathbf{n}}_{f_l} \times {\mathbf{n}}_{f_p}}{\|{\mathbf{n}}_{f_l}\times {\mathbf{n}}_{f_p}\|}
\label{eq:def_tangent}
\end{equation}
where $s$ denotes the arc-length parameter of the toolpath $\mathbf{p}$. 

The local spacing between iso-contour toolpaths can be characterized by the magnitude of the projected gradient of $f_p$ on the layer surface defined by $f_l$:
\begin{equation}
\nabla_l f_p = \nabla f_p - (\nabla f_p \cdot \mathbf{n}_{f_l})\mathbf{n}_{f_l}
\label{eqn:def_projected_gradient}
\end{equation}
The magnitude $\|\nabla_l f_p\|$ provides a measure of the path spacing, while its normalized direction $\frac{\nabla_l f_p}{\|\nabla_l f_p\|}$ serves as the third vector of the Darboux frame when both the layer and the path fields are evaluated at a point.

The Mean curvature $K_M$ and the Gaussian curvature $K_G$ of a layer at a given point are computed following \cite{goldman_curvature_2005}:
\begin{equation}
    K_M = \frac{\nabla f_l^T\cdot \mathbf{H}_{f_l} \cdot \nabla f_l - \|\nabla f_l\|^2(trace(\mathbf{H}_{f_l}))}{2\|\nabla f_l\|^3}
    \label{eqn:def_mean_curv}
\end{equation}

\begin{equation}
    K_G = \frac{\nabla f_l^T\cdot \bar{\mathbf{H}}_{f_l} \cdot \nabla f_l }{\|\nabla f_l\|^4}
    \label{eqn:def_gaussian_curv}
\end{equation}
where $\mathbf{H}_{f_l}$ denotes the Hessian of $f_l$ and $\bar{\mathbf{H}}_{f_l}$ its adjugate (adjoint) matrix. The Mean and Gaussian curvatures correspond to the mean and product of the two principal curvatures, respectively, from which the principal curvatures can be directly evaluated.

The tangent vector $\mathbf{t}$, defined in Eq.~\eqref{eq:def_tangent}, allows the geodesic curvature of the toolpath to be expressed as:
\begin{equation}
\kappa_{geo}(\mathbf{x}) = \|\frac{d\mathbf{t}}{ds} - \Big(\frac{d\mathbf{t}}{ds} \cdot \mathbf{n}_{f_l}\Big)\mathbf{n}_{f_l}\|
\label{eq:geodesic_curv}
\end{equation}
where $s$ denotes the arc-length parameterization. A full derivation of $\kappa_{geo}$ in terms of the first and second derivatives of the field functions is provided in ~\ref{appendix:geodesic}. 

Geometrically, $\kappa_{geo}$ quantifies the in-surface turning of the toolpath (see Fig.~\ref{fig:derivativeGeometry}(b)). The normal curvature can similarly be defined, if required.

Hence, all relevant geometric quantities of the layers and toolpaths, such as normals, curvatures, and spacings, can be directly computed from the implicit field representations. In the following section, we describe how these geometric measures are used to formulate differentiable loss functions for various objectives in multi-axis processes.

\section{Objectives for Field Optimization}
\label{sec:losses_objectives}

\begin{figure}
    \centering
    \includegraphics[width=0.95\linewidth]{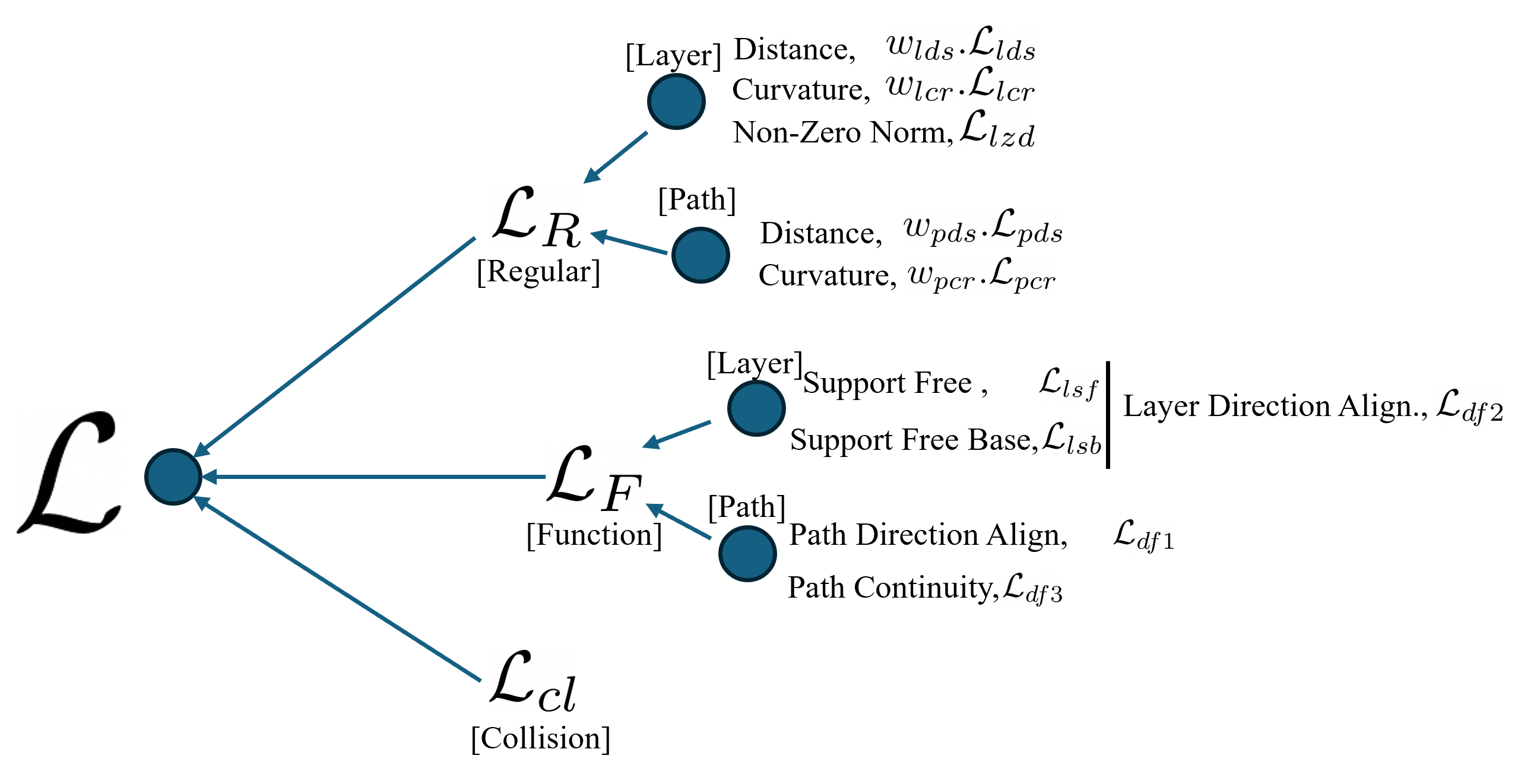}
    \caption{Illustration of the total loss function framework used across different applications. Individual loss components are categorized as: (i) regularization losses ($\mathcal{L}_R$), which enforce geometric properties such as spacing and curvature; (ii) functional losses ($\mathcal{L}_F$), which promote specific objectives including direction-field alignment or support-free layers; and (iii) collision losses ($\mathcal{L}_{cl}$), which account for potential collisions with layers, the part itself in milling operations, or other obstacles.}
    \label{fig:loss_tree}
\end{figure}
\revision{}{Fig.~\ref{fig:loss_tree} illustrates the different loss functions that constitute the total loss, whose minimization defines the main objective of the optimization process. This section introduces the individual loss functions corresponding to different requirements, while Sec.~\ref{sec:total_losses} presents their application-specific composition into the total loss.}


\subsection{Collision Detection and Avoidance}
\label{subsec:collision}
\subsubsection{Implicit Formulation}

We define a collision as any configuration when the tool volume overlaps with another object, with the primary concern in additive manufacturing being collision between the tool and the partially fabricated part. A naive approach is to sample the part with $N$ points and test whether any lie inside the tool when it is placed at any other point, leading to a computational cost of $O(N^2 t_\mathrm{in})$ per step, where $t_\mathrm{in}$ is the cost of an individual inclusion test. Moreover, it offers no explicit guidance on the direction in which the tool or the layers should be modified to eliminate the collision.

Instead, we sample the tool itself with $K$ points and test their inclusion within the part. Assuming each test takes $O(t_\mathrm{col})$ (which is effectively $O(1)$ in our implementation), the total cost becomes $O(KN t_\mathrm{col})$. Since $K \ll N$  (the tool being smaller and/or also can be sampled only near its boundary)
this approach is considerably more efficient and suitable for iterative optimization where frequent collision checks are required.

Importantly, this collision condition can be expressed directly in terms of the implicit scalar field representing the layers. Doing so allows collision avoidance to be formulated as a differentiable objective for field optimization, influencing both local tool orientations and the global field structure, thus yielding more coherent collision-free configurations.

\begin{figure}[!t]
    \centering
    \includegraphics[width=0.5\linewidth]{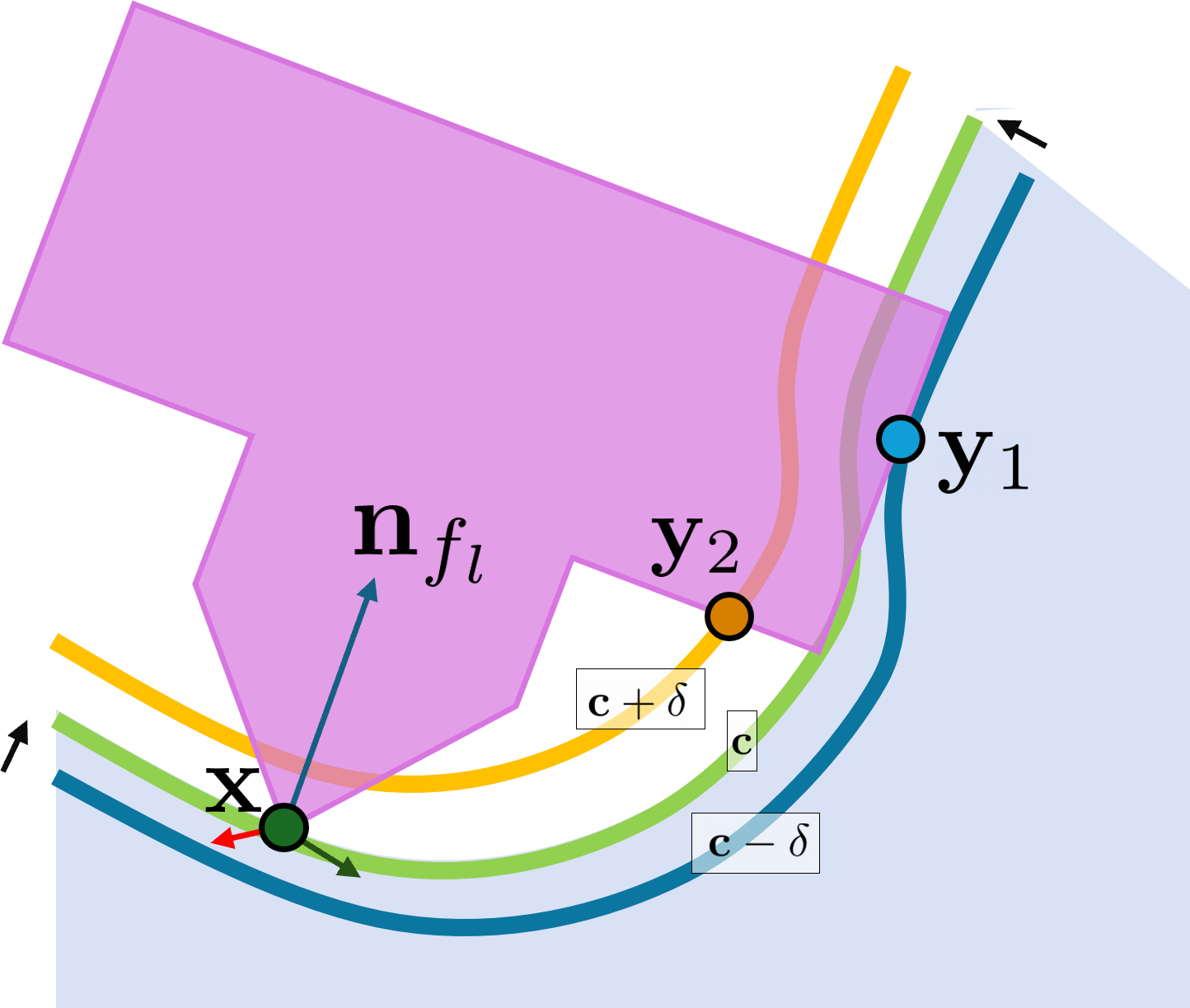}
    \caption{Illustration of the proposed collision detection scheme. The shaded gray region represents the printed part, while the coloured curves (blue, green, yellow) denote successive layers. The tool (pink) is positioned at a point $\mathbf{x}$ on the current layer ($f_l(\mathbf{x}) = c$) and oriented along the layer normal $\mathbf{n}_{f_l}$. Sampling points $\mathbf{y}_1$ and $\mathbf{y}_2$ on the tool correspond to the cases of collision ($f_l(\mathbf{y}_1) < c$) and non-collision ($f_l(\mathbf{y}_2) > c$), respectively. We denote the tool configuration at $\mathbf{x}$ as $\mathcal{T}_{\mathbf{x}}$, with $\mathbf{y}_1, \mathbf{y}_2 \in \{\mathcal{T}_{\mathbf{x}}\}$.}
    \label{fig:collision_illustrate}
\end{figure}

For any point $\mathbf{x}$ within the part, the layer field $f_l(\mathbf{x})$ provides a local normal direction $\mathbf{n}_{f_l}(\mathbf{x})$. Using this normal, we construct a local reference frame and place the tool, represented by the point set $\{\mathcal{T}\}$, within it (see Fig.~\ref{fig:collision_illustrate}). The condition for collision-free placement of the tool at $\mathbf{x}$ can then be expressed as:
\begin{equation}
    f_l(\mathbf{y}) > f_l(\mathbf{x}), \quad \forall \mathbf{y} \in \{\mathcal{T}_{\mathbf{x}}\} \;\&\; \Tilde{\Omega}_{\mathrm{part}}(\mathbf{y}) = 1,
    \label{eqn:collision_main}
\end{equation}
where $\{\mathcal{T}_{\mathbf{x}}\}$ is the set of points in $\{\mathcal{T}\}$, placed and oriented at the local frame at $\mathbf{x}$.  $\Tilde{\Omega}_{\mathrm{part}}(\mathbf{y})$ is an indicator function defined using the implicit signed-distance field $sdf_{model}$ (see Sec.~\ref{subsec:implicit_model_rep}):
\begin{equation}
    \Tilde{\Omega}_{\mathrm{part}}(\mathbf{y}) =
    \begin{cases}
        1, & sdf_{model}(\mathbf{y}) \le 0, \\
        0, & sdf_{model}(\mathbf{y}) > 0.
    \end{cases}
    \label{eqn:omega_part}
\end{equation}
That is, a point $\mathbf{y}$ on the tool positioned at $\mathbf{x}$ collides with the printed region if $f_l(\mathbf{y}) \le f_l(\mathbf{x})$ and $sdf_{model}(\mathbf{y}) \le 0$; the condition in Eq.~\eqref{eqn:collision_main} prevents this.

Since the local frame depends on $\nabla f_l(\mathbf{x})$, the tool point set $\{\mathcal{T}_{\mathbf{x}}\}$ becomes a function of the layer-field gradient:
\begin{equation}
    \mathcal{T}_{\mathbf{x}} = \mathcal{T}(\mathbf{x},\nabla f_l(\mathbf{x})).
    \label{eqn:toolpoint}
\end{equation}
A general collision-free condition for any configuration can then be written as:
\begin{equation}
    F(\mathbf{y}) > 0, \quad \forall \mathbf{y} \in \{\mathcal{T}_{\mathbf{x}}\},
    \label{eqn:general_collision}
\end{equation}
where $F$ is a scalar function that must remain positive for all points $\mathbf{y}$ on or inside the tool.  
For internal collisions within layers, $F(\mathbf{y}) = f_l(\mathbf{y}) - f_l(\mathbf{x})$, as in Eq.~\eqref{eqn:collision_main}. For external obstacles, $F$ can be represented as a signed-distance field that is negative within the obstacle volume.

The condition in Eq.~\eqref{eqn:general_collision} can be formulated as a differentiable loss function defined on the layer-field network $f_l$:
\begin{equation}
    \mathcal{L}_{cl} = \frac{1}{|\{\Omega\}|\,|\{\mathcal{T}\}|} \sum_{\mathbf{x} \in \{\Omega\}} \sum_{\mathbf{y} \in \{\mathcal{T}_{\mathbf{x}}\}} 
    \left( 10\, \mathrm{ReLU}(-F(\mathbf{y}) + \delta) \right)^2,
    \label{eqn:general_collision_loss}
\end{equation}
where $|\cdot|$ denotes set cardinality. The ReLU function \cite{nair2010rectified} is used to clip negative values to zero. The small offset $\delta$ enforces strict positivity in the constraint. The factor of $10$ in the loss function scales the ReLU slope, ensuring the gradient magnitudes are within a suitable range within the total loss. This intrinsic scaling can be combined with any additional task-specific weighting when needed.

For additive manufacturing, substituting with the Eq.~\eqref{eqn:collision_main} yields the specific loss for collision avoidance:
\begin{equation}
    \mathcal{L}_{cla} = \frac{1}{|\{\Omega\}|\,|\{\mathcal{T}\}|} 
    \sum_{\mathbf{x} \in \{\Omega\}} \sum_{\mathbf{y} \in \{\mathcal{T}_{\mathbf{x}}\}} 
    \left( 10\, \mathrm{ReLU}(f_l(\mathbf{x}) - f_l(\mathbf{y}) + \delta)\,
    \Tilde{\Omega}_{\mathrm{part}}(\mathbf{y}) \right)^2.
    \label{eqn:main_collision_loss}
\end{equation}

\subsubsection{Variant for CNC Milling}
\label{subsub:collision_milling}

The general collision-avoidance formulation introduced in Eq.~\eqref{eqn:general_collision} can be modified to other manufacturing processes beyond additive fabrication, specifically, to those involving sequential removal or addition of material, such as CNC milling.
In rough (or volume) milling, the objective is to remove bulk material from an initial stock of material to approach a target shape. Here, the tool operates in the region surrounding, but external to, the target part.

Let the target part be represented by $\mathcal{H}$ and the initial material block by $\mathcal{M}$. The domain where the layers are computed is then defined as $\Omega = \mathcal{M} - \mathcal{H}$.
While the loss function preventing collision with remaining material layers during milling remains the same as Eq.~\eqref{eqn:main_collision_loss}, we must ensure that the tool does not intersect regions already milled \revision{}{or any other obstacle}, and the cutting surface does not penetrate the target geometry.

This can be achieved by defining the general function $F$ in Eq.~\eqref{eqn:general_collision_loss} with the signed-distance field of the target part, $sdf_{model}$. The resulting constraint becomes:
\begin{equation}
F(\mathbf{y}) = sdf_{model}(\mathbf{y}) > 0, \quad \forall \mathbf{y} \in {\mathcal{T}_{\mathbf{x}}},
\label{eqn:collision_milling1}
\end{equation}
which ensures that all sampled points of the tool, when positioned at $\mathbf{x}$, lie outside the target surface. \revision{}{For additional obstacles, they can either be incorporated into the existing $sdf_{model}$ or represented by separate functions $F$ for each obstacle}.

The corresponding loss function, denoted by $\mathcal{L}_{clm}$, is then expressed as:
\begin{equation}
\mathcal{L}_{clm} = \frac{1}{|\{\Omega\}||\{\mathcal{T}\}|}\sum_{\mathbf{x} \in {\Omega}}\sum_{\mathbf{y} \in {\mathcal{T}{\mathbf{x}}}}
\left(10 ReLU(-sdf_{model}(\mathbf{y}) + \delta)\right)^2,
\label{eqn:loss_collision_milling}
\end{equation}
indicating that any sampled point of the tool must remain outside the target part throughout the milling process.


\subsection{Direction Optimization for Toolpaths}
\label{subsec:direction_loss}

In many cases, the manufacturing toolpath must align with a prescribed direction field. For instance, following the field of maximum principal stresses can enhance the stiffness of a printed structure \cite{zhang_toolpath_2025}. Formally, this requirement can be expressed as a constraint aligning the toolpath tangent $\mathbf{t}$ to a given direction field $\mathbf{d}$ defined on $\Omega_d \subseteq \Omega$. We enforce them to be parallel at each point, implying that their cross-product vanishes:
\begin{equation}
\mathbf{t} \times \mathbf{d}
=
\left(
\frac{\mathbf{n}_{f_l} \times \mathbf{n}_{f_p}}
{\|\mathbf{n}_{f_l} \times \mathbf{n}_{f_p}\|}
\right)
\times
\mathbf{d}
= 0,
\quad \forall \mathbf{x} \in \Omega_d
\label{eqn:direction_follow_main}
\end{equation}

The normalization term in Eq.~\eqref{eqn:direction_follow_main} can introduce numerical instabilities and hinder topological changes, since singularities, i.e., points of ambiguity in the gradient field (see Sec.~\ref{subsec:singularity}), may in fact be desirable. Furthermore, the solution becomes highly sensitive to the initial guess. To resolve these issues, we propose a modified formulation that omits the normalization term:
\begin{equation}
(\mathbf{n}_{f_l} \times \nabla f_p) \times \mathbf{d} = 0,
\quad \forall \mathbf{x} \in \Omega_d
\label{eqn:direcion_follow_true}
\end{equation}
Removing normalization allows convergence towards $\nabla f_p = 0$, thereby enabling singularities and topological transitions.

For layer surfaces, we further require that the prescribed directions lie within the tangent space of the layers:
\begin{equation}
\mathbf{n}_{f_l}\cdot \mathbf{d} = 0,
\quad \forall \mathbf{x} \in \Omega_d
\label{eqn:direction_layer_only}
\end{equation}
Similarly, for toolpaths, we enforce continuity along the prescribed directions (when non-zero):
\begin{equation}
\nabla f_p \cdot \mathbf{d} = 0,
\quad \forall \mathbf{x} \in \Omega_d
\label{eqn:direction_path_only}
\end{equation}

By the properties of vector multiplication, when $\nabla f_p \neq 0$, Eq.~\eqref{eqn:direcion_follow_true} implicitly satisfies Eqs.~\eqref{eqn:direction_layer_only} and \eqref{eqn:direction_path_only}. However, when $\nabla f_p = 0$, Eq.~\eqref{eqn:direcion_follow_true} no longer guarantees Eq.~\eqref{eqn:direction_layer_only}. Consequently, allowing a zero gradient may hinder convergence to the desired layer field $f_l$. To ensure robustness, we explicitly include Eqs.~\eqref{eqn:direction_layer_only} and \eqref{eqn:direction_path_only} as independent constraints for stable convergence.

In summary, the following loss functions are introduced to impose direction guidance on toolpaths:
\begin{equation}
\mathcal{L}_{df1}
=
\frac{10}{|{\Omega_d}|}
\sum_{\mathbf{x} \in {\Omega_d}}
\|
(\mathbf{n}_{f_l} \times \nabla f_p) \times \mathbf{d}
\|^2
\label{eqn:loss_df_main}
\end{equation}
\begin{equation}
\mathcal{L}_{df2}
=
\frac{10}{|{\Omega_d}|}
\sum_{\mathbf{x} \in {\Omega_d}}
(\mathbf{n}_{f_l} \cdot \mathbf{d})^2
\label{eqn:loss_df_layer}
\end{equation}
\begin{equation}
\mathcal{L}_{df3}
=
\frac{10}{|{\Omega_d}|}
\sum_{\mathbf{x} \in {\Omega_d}}
(\nabla f_p \cdot \mathbf{d})^2
\label{eqn:loss_df_tp}
\end{equation}
Here, the factor of 10 acts as an intrinsic normalization weight, similar with the formulation in the previous sub-section.


\subsection{Curvature}
\label{subsec:curvatures}
As discussed in Sec.~\ref{subsec:overview_derivatives}, our implicit representation enables the computation of higher-order derivatives, allowing direct evaluation of the Mean curvature (Eq.~\eqref{eqn:def_mean_curv}) and the Gaussian curvature (Eq.~\eqref{eqn:def_gaussian_curv}). Consequently, the principal curvatures of the iso-surface at any point can be computed as
\begin{equation}
\kappa = {\kappa_1, \kappa_2} = K_M \pm \sqrt{K_M^2 - K_G}.
\label{eqn:principal_curv}
\end{equation}

To ensure manufacturability and maintain print quality, we constrain the surface curvature to remain below a threshold $\kappa_T$. Regions of high curvature can cause abrupt directional changes in the tool motion, which may approach the kinematic limits of the machine~\cite{conway2013experimental, chen2025co}, leading to non-uniform acceleration and inconsistent material extrusion. High curvatures can also result in printing artifacts such as gaps or discontinuities in the toolpath~\cite{huang2026force, lim2016modelling}. Collectively, these effects degrade both the geometric fidelity and the mechanical integrity of the printed part~\cite{huang2026force}. Furthermore, highly concave regions with large local curvature can introduce the risk of local gouging or collision in both milling and printing operations~\cite{liang_review_2021, liu_neural_2025}. To mitigate these issues, we introduce a curvature-based loss term that penalizes violations of the prescribed curvature threshold:

\begin{equation}
\mathcal{L}_{lcr} = \frac{1}{|{\Omega}|}\sum_{\mathbf{x} \in {\Omega}} \left(30 ReLU(|\kappa| - \kappa_T)\right)^2.
\label{eqn:loss_curvature_Surf}
\end{equation}

For the toolpath, it is also desirable that direction-aligned fibers avoid sharp turns along the surface, as these can induce printing defects~\cite{brooks_manufacturing_2018, qu_placement_2021, halbritter_leveraging_2023, xiao_field-based_2025}. To capture this, we employ the geodesic curvature ($\kappa_{geo}$, Eq.\eqref{eq:geodesic_curv} in Sec.~\ref{subsec:overview_derivatives}), which measures the bending of a curve along a surface more accurately than other curvature metrics like Laplacian. Similar to the surface-based term, we define a corresponding toolpath curvature loss:
\begin{equation}
\mathcal{L}_{pcr} = \frac{1}{|{\Omega'}|}\sum_{\mathbf{x} \in {\Omega'}} \left(10 ReLU(|\kappa_{geo}| - \kappa_T)\right)^2.
\label{eqn:loss_curvature_TP}
\end{equation}

In Eq.~\eqref{eqn:loss_curvature_TP}, the domain is denoted by $\Omega'$ instead of $\Omega$ to filter out regions of singularity, as will be discussed in Sec.~\ref{subsec:singularity}. The factors of 30 and 10 in Eq.~\eqref{eqn:loss_curvature_Surf} and Eq.~\eqref{eqn:loss_curvature_TP}, respectively, act as intrinsic scaling terms that modulate the gradient magnitudes of the loss functions, consistent with the normalization strategy used in our previous loss formulations. In all our experiments, we set $\kappa_T = \frac{1}{35}\text{mm}^{-1}$ for $\mathcal{L}_{lcr}$ and $\kappa_T = \frac{1}{5}\text{mm}^{-1}$ for $\mathcal{L}_{pcr}$, unless otherwise specified. These values were determined empirically based on fabrication trials on our devices. \revision{}{For example, imposing sharp turns (e.g., with a radius of $5$ mm) led to deviations of the fiber from its intended trajectory during fabrication.} \revision{but they}{These} can be adjusted to suit different manufacturing setups.


\subsection{Distance Control}
\label{subsub:layer_density}
The variation in layer thickness (or path spacing along a layer) is constrained by both hardware and material properties. Moreover, due to the inherently non-linear nature of the extrusion process, maintaining \revision{}{as uniform as possible} layer thickness is desirable for improved \revision{}{extrusion} control \revision{and higher print quality}{}~\cite{zhang_zhangty019s3_deformfdm_2025}.

Since the scalar field $f_l$ encodes the order of material deposition, the norm of its gradient, $\|\nabla f_l\|$, represents the local rate of growth and thus determines the spacing between iso-surfaces at any point. To encourage consistent layer spacing, we constrain the spatial derivative of this gradient norm by:
\begin{equation}
\mathcal{L}_{lds} = \frac{1}{|{\Omega}|}
\sum_{\mathbf{x} \in {\Omega}}\sum_{k \in {x,y,z}}
\frac{1}{\|\nabla f_l\|}\left(\frac{d(\|\nabla f_l\|)}{dk}\right)^2.
\label{eqn:layer_density_loss}
\end{equation}

\begin{figure}
    \centering
    \includegraphics[width=0.95\linewidth]{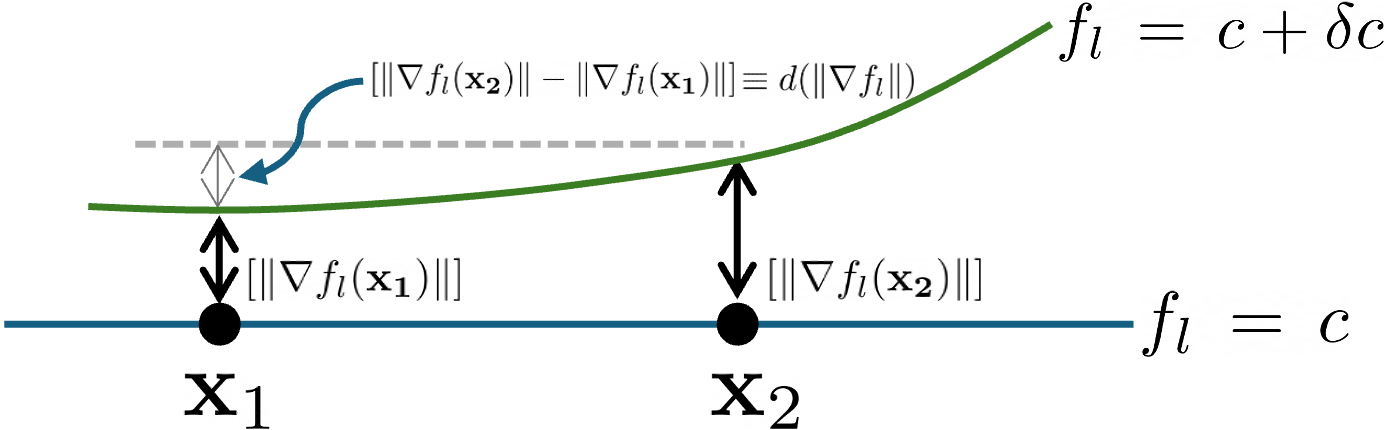}
    \caption{An illustration of our inter-layer distance control strategy. The blue and green lines represent two iso-surfaces with scalar values $c$ and $c+\delta c$ respectively. The magnitude of the gradient is the rate of growth of the field and can be used to estimate inter-layer spacing. Minimizing the differences in gradient magnitude encourages uniform layer spacing. This illustration is intended for conceptual clarity and is not mathematically exact. The values are used inside the square brackets (e.g. $[\|\nabla f_l\|]$) to indicate that the expressions inside the brackets can be used as a measure of those distances and does not mean the exact value.}
    \label{fig:gradNormLossIllustrate}
\end{figure}

An illustration of how this relates to local changes in layer spacing is shown in Fig.~\ref{fig:gradNormLossIllustrate}.
However, since this loss alone does not penalize a trivial gradient solution (i.e., all-zero), we introduce an additional term to explicitly discourage vanishing gradients:
\begin{equation}
\mathcal{L}_{lzd} = \frac{1}{|{\Omega}|}
\sum_{\mathbf{x} \in {\Omega}} e^{-100\|\nabla f_l\|^2}.
\label{eqn:loss_small_grad}
\end{equation}

\revision{}{The factor $100$ in the exponent controls the width of the region around zero over which the loss has a significant effect. For a negative exponent, smaller values of this factor result in a slower decay, thereby allowing the loss to influence a broader range of $\|\nabla f_l\|$ values away from zero.}

To control the spacing between adjacent toolpaths, each toolpath, represented as an iso-curve of $f_p$, is encouraged to maintain a uniform distribution over the surface. This will help us to introduce higher continuous-fiber coverage \cite{zhang_toolpath_2025}. As discussed in Sec.~\ref{subsec:overview_derivatives}, the projected gradient of $f_p$, denoted $\nabla_l f_p$, captures the variation of the path field along the tangent directions of the layer surface (i.e., within the iso-surfaces of $f_l$). Therefore, we constrain the magnitude of this projected gradient to promote consistent toolpath spacing across the surface:
\begin{equation}
    \mathcal{L}_{pds} = \frac{1}{|\{\Omega\}|}\sum_{\mathbf{x} \in 
 \{\Omega\}} (1-\| \nabla f_p - (\nabla f_p \cdot\mathbf{n}_{f_l})\mathbf{n}_{f_l}\|)^2
    \label{eqn:toolpath_density_loss}
\end{equation}

This constraint encourages the norm of $\nabla_l f_p$ to approach $1$ everywhere, where the value $1$ is chosen for convenience; its exact magnitude does not affect the qualitative behavior or meaning of the loss.


\subsection{Support-Free}

During the printing process, material must be deposited onto an underlying substrate. An overhang refers to any portion of the print that extends outward beyond the layer beneath it, lacking direct material support. In the absence of such support, the deposited material can sustain only a limited overhang, determined by material, hardware, and process characteristics \cite{zhang_zhangty019s3_deformfdm_2025}. Any overhang exceeding this limit necessitates additional support structures. To minimize such requirements, we impose the following condition to ensure that each new layer is deposited on underlying material along the printing orientation, thereby making it self-supporting:
\begin{equation}
\mathbf{n}_{f_l} \cdot \mathbf{N_s}(\mathbf{x}) > \cos(90^\circ + \alpha)
\quad \forall \mathbf{x} \in \partial \Omega \setminus \mathcal{B}(\Omega)
\label{eqn:support_free}
\end{equation}
\begin{wrapfigure}{r}{0.35\textwidth}
\centering
\includegraphics[width=0.35\textwidth]{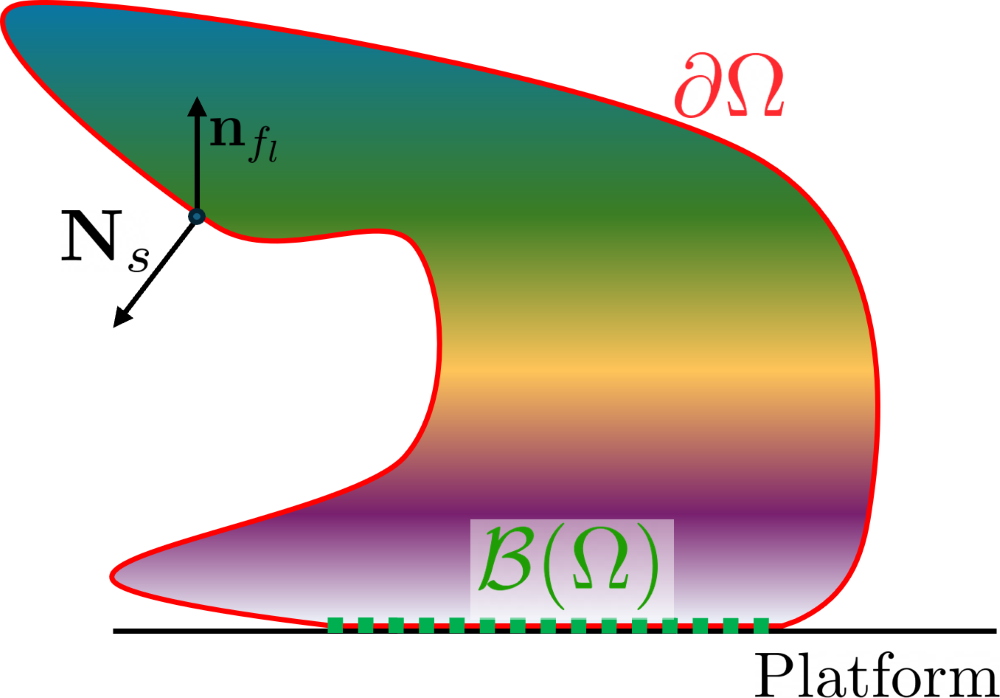}
\end{wrapfigure}
where $\mathbf{N_s}(\mathbf{x})$ denotes the surface normal at point $\mathbf{x}$, and $\alpha$ is the maximum allowable overhang angle, dependent on the material and process parameters. This condition is evaluated over the part’s boundary $\partial \Omega$, excluding the base region $\mathcal{B}(\Omega)$ in contact with the build platform (see adjacent figure in the right).

This constraint is then formulated as the following loss:
\begin{equation}
\mathcal{L}_{lsf}=
\frac{1}{|\{\partial \Omega \setminus \mathcal{B}(\Omega)\}|}
\sum_{\mathbf{x} \in {\partial \Omega \setminus \mathcal{B}(\Omega)}}
\left(25~ ReLU\left(\cos(90^\circ + \alpha - \delta) -
\mathbf{n}_{f_l} \cdot \mathbf{N_s}(\mathbf{x})\right)\right)^2
\label{eqn:loss_support_free}
\end{equation}

In practice, we found an overhang limit of $\alpha = 45^\circ$ suitable for our setups. To strictly enforce the “inequality” condition of Eq.~\eqref{eqn:support_free}, the threshold is slightly reduced by $\delta$, resulting in an effective support-free angle of $(\alpha - \delta) = 43^\circ$ in Eq.~\eqref{eqn:loss_support_free}.

Additionally, the base of the part in contact with the build platform should align with the platform’s normal. This is encouraged through the following loss:
\begin{equation}
\mathcal{L}_{lsb} =
\frac{1}{|\{\mathcal{B}(\Omega)\}|}
\sum_{\mathbf{x}\in{\mathcal{B}(\Omega)}}
\left\|2.5\left(\mathbf{n}_{f_l}(\mathbf{x})-\mathbf{n}_\mathcal{P}\right)\right\|^2
\label{eqn:loss_support_dir}
\end{equation}
where $\mathbf{n}_\mathcal{P}$ denotes the platform normal. The weighting constants 25 and 2.5 in these two loss functions serve as intrinsic scaling factors to control the gradient magnitude, in the same manner as the other loss functions presented before.

Together, the losses in Eq.~\eqref{eqn:loss_support_free} and Eq.~\eqref{eqn:loss_support_dir} encourage self-supporting layer orientations for a fixed part placement, consistent with prior works \cite{zhang_zhangty019s3_deformfdm_2025, li_vector_2022, liu_neural_2024}. Here, the user explicitly specifies the region of contact with the  platform ($\mathcal{B}(\Omega)$). An alternative formulation seeks to minimize the amount and improve the ease of constructing the required support structures by optimizing the part’s orientation and placement, when a support-free condition is not explicitly introduced. The relevant formulation will be discussed in Sec.~\ref{subsec:platfrom_loss}.

\subsection{Setup Orientation and Position}
\label{subsec:platfrom_loss}
When a support-free condition is not explicitly enforced (e.g., if the primary concern is the toolpath alignment), we allow the use of support structures to enable printing. However, we still aim to minimize the amount of support material and improve the ease of constructing such structures. In this context, the orientation and placement of the part relative to the build platform are crucial, as support structures generally connect the part to the platform \cite{zhang_support_2023}. Another important consideration is the potential collision of the tool with the platform during printing. Given the geometry of the printing layers, we therefore seek a part configuration that avoids any collisions with the platform throughout the printing process.

We represent the platform as a plane ($\mathcal{P}$) defined by a point ($\mathbf{o}_\mathcal{P}$) and a surface normal ($\mathbf{n}_\mathcal{P}$):
\begin{equation}
    \mathcal{P} \equiv \{\mathbf{o}_\mathcal{P}, \mathbf{n}_\mathcal{P}\}
    \label{eqn:eq_patform_rep}
\end{equation}

The most basic requirement is that the part lies entirely above the platform. At the same time, the part should not be positioned excessively far from the platform, as this would lead to unnecessary use of support material. This requirement is introduced through the following loss:

\begin{equation}
    \mathcal{L}_{sdp} =(\min_{\mathbf{x} \in \{\partial \Omega\}} ((\mathbf{x} - \mathbf{o}_\mathcal{P}) \cdot \mathbf{n}_\mathcal{P})-\delta)^2
    \label{eqn: loss_platform_pos}
\end{equation}
which pushes the lowest point of the part to a distance $\delta$ above the platform. A small positive $\delta$ prevents numerical errors and ensures that the part remains strictly above the platform. This can also be used to control the minimum clearance required for constructing support structures using algorithms such as those proposed by Zhang et al. \cite{zhang_support_2023}.

The support-generation method compatible with curved-layer printing \cite{zhang_support_2023} constructs support material that converges toward the platform. In such cases, it is desirable for the regions requiring support to face the platform, improving the ease of support construction. We therefore define a loss to orient the platform appropriately with respect to the part:
\begin{equation}
    \mathcal{L}_{sor} = \frac{1}{|\{ST(\partial\Omega)\}|}\cdot\sum_{\{ST(\partial\Omega)\}} (\|\mathbf{n}_\mathcal{P} - \mathbf{n}_{f_l}(\mathbf{x})\|)^2
    \label{eqn: loss_platform_orient}
\end{equation}
where $ST(\partial \Omega)$ represent the region of $\partial \Omega$ requiring support.

For collision, the platform can be considered as an external collision with the function $F$ (equation \ref{eqn:general_collision}) defined as following to ensure that all the tool points lie above the platform:
\begin{equation}
    F(\mathbf{y}) = (\mathbf{y} - \mathbf{o}_\mathcal{P}) \cdot \mathbf{n}_\mathcal{P}
    \label{eqn: platform_collision}
\end{equation}
Hence the platform collision loss is defined using Eq.~\eqref{eqn:general_collision_loss} as:
\begin{equation}
    \mathcal{L}_{cls} = \frac{1}{|\{\Omega\}| |\{{\mathcal{T}}\}|}\sum_{\mathbf{x} \in \{\Omega\}}\sum_{\mathbf{y} \in \{\mathcal{T}_{x}\}}(10 ReLU(-(\mathbf{y} - \mathbf{o}_\mathcal{P}) \cdot \mathbf{n}_\mathcal{P} + \delta))^2
    \label{eqn:loss_platform_pos_reduce}
\end{equation}

All these loss functions determine the orientation and position of the platform with respect to the part. These constraints are applied at the platform level rather than directly on the implicit field, and therefore do not represent a direct application of our implicit representation. Nevertheless, they play a crucial role in ensuring manufacturability. We include them here to demonstrate the completeness of our pipeline and to illustrate how the proposed framework can be extended through complementary applications.


\section{Total Loss for Optimization}
\label{sec:total_losses}

Following the definition of the individual loss functions for various objectives presented in Sec.~\ref{sec:losses_objectives}, we now define the total loss as a combination of the relevant losses corresponding to specific application requirements. In this work, we demonstrate four distinct applications as examples. The overall structure of the loss formulation is illustrated in Fig.~\ref{fig:loss_tree}, where the losses are categorized into three groups: (1) regularization losses, which control geometric properties such as layer and toolpath spacing as well as curvature; (2) functional losses, which enforce directional alignment and support-free printing; and (3) collision losses. The combined total loss for each of our four applications explored in this work is presented below.


\subsection{Support \& Collision Free Printing}
\begin{equation}
    \begin{split}
        \mathcal{L}_{SCF} = [w_{lds}\mathcal{L}_{lds} +  w_{lcr}\mathcal{L}_{lcr} + \mathcal{L}_{lzd}] \\
        + [\mathcal{L}_{lsf} +\mathcal{L}_{lsb}]\\
        + w_{cl}\mathcal{L^*}_{cl} \\
    \end{split}
    \label{eqn:total_loss_sf}
\end{equation}

Eq.~\eqref{eqn:total_loss_sf} presents the total loss for the Support and Collision-Free (SCF) case. Since all the requirements in this formulation are defined over the layers, we include the layer regularization losses like the layer spacing loss ($\mathcal{L}{lds}$, Eq.~\eqref{eqn:layer_density_loss}), the layer curvature loss ($\mathcal{L}_{lcr}$, Eq.~\eqref{eqn:loss_curvature_Surf}), and the layer zero-gradient loss ($\mathcal{L}_{lzd}$, Eq.~\eqref{eqn:loss_small_grad}), which collectively control the geometric regularity of the layers.
For the functional group of losses, we include the support-free inclination loss ($\mathcal{L}_{lsf}$, Eq.~\eqref{eqn:loss_support_free}) and the base orientation loss ($\mathcal{L}_{lsb}$, Eq.~\eqref{eqn:loss_support_dir}).
Finally, the collision avoidance requirement is incorporated through the collision loss $\mathcal{L}_{cl}$, implemented as $\mathcal{L}_{cla}$ (Eq.~\eqref{eqn:main_collision_loss}).
Next, we present the loss formulation for the Direction Alignment and Collision-Free (DAC) case.


\subsection{Direction Alignment with Collision Avoidance}
\begin{equation}
    \begin{split}
        \mathcal{L}_{DAC} = [w_{lds}\mathcal{L}_{lds} +  w_{lcr}\mathcal{L}_{lcr} + \mathcal{L}_{lzd}] \\
         + \mathcal{L}_{df2}\\
        + w_{cl}\mathcal{L^*}_{cl}
    \end{split}
    \label{eqn:total_loss_str}
\end{equation}

In this case, the functional objective is to align the layers with a prescribed input direction field. Therefore, the support-free losses are replaced by the layer–direction alignment loss, $\mathcal{L}_{df2}$ (Eq.~\eqref{eqn:loss_df_layer}). The remaining components, such as the layer regularization and collision losses, are retained as defined in the previous case to ensure smoothness and collision-free deposition.


\subsection{Toolpath Geometry Control}

The case of Toolpath Geometry Control (Eq.~\eqref{eqn:total_loss_tp}) builds upon the previous formulation but additionally aims to generate continuous-fiber toolpaths. To this end, we introduce toolpath regularization losses ($\mathcal{L}_{pcr}$ and $\mathcal{L}_{pds}$) along with toolpath direction-alignment losses ($\mathcal{L}_{df1}$ and $\mathcal{L}_{df3}$), which collectively enforce geometric requirements and directional consistency of the toolpaths:
\begin{equation}
    \begin{split}
        \mathcal{L}_{TPD} = [(w_{lds}\mathcal{L}_{lds} +  w_{lcr}\mathcal{L}_{lcr} + \mathcal{L}_{lzd})\\
        +(w_{pcr}\mathcal{L}_{pcr}  + w_{pds}\mathcal{L}_{pds})] + \\
        + [\mathcal{L}_{df1} + 
         \mathcal{L}_{df2}+
        \mathcal{L}_{df3}]\\ 
    \end{split}
    \label{eqn:total_loss_tp}
\end{equation}


\subsection{Collision-Free Milling}

In the final case, we extend the implicit-layer formulation to multi-axis milling, aiming to automatically generate rough-milling layers that prevent tool collisions. Specifically, we introduce both the tool–part collision loss ($\mathcal{L}_{clm}$, Eq.~\eqref{eqn:collision_milling1}) and the inter-layer collision loss ($\mathcal{L}_{cla}$, Eq.~\eqref{eqn:main_collision_loss}), ensuring that the milling tool remains clear of both the target part and intermediate surfaces during material removal. The combined loss function is expressed as:

\begin{equation}
    \begin{split}
        \mathcal{L}_{MLC} = [w_{lds}\mathcal{L}_{lds} +  w_{lcr}\mathcal{L}_{lcr} + \mathcal{L}_{lzd}] \\
        + w_{cl}\mathcal{L^*}_{cl}
    \end{split}
    \label{eqn:total_loss_mill}
\end{equation}

\subsection{Weighting Scheme}
\label{subsec:weights}
In all the loss functions, weights are applied to the regularization terms such as distance and curvature. This approach allows our pipeline to remain flexible across different multi-axis manufacturing tasks, where the desired balance between curvature control and distance uniformity may vary depending on the process, hardware, material, and functional requirements. The weights used for all cases are summarized in Table~\ref{tab:weights}. We assign a higher $w_{lds}$ for all cases except the SCF configuration. Unlike the SCF case, where the functional requirement is defined only on the boundary, the other cases impose functional constraints within the volume (e.g., directional alignment), which necessitates stronger internal regularization. Furthermore, in these cases, support layers may be required for printing, and maintaining better layer-distance uniformity facilitates the generation of compatible support layers using scalar-field-based methods \cite{zhang_support_2023}.

The path curvature loss employs a variable weighting scheme (consistent across all cases), as shown in the Table~\ref{tab:weights} with a detailed explanation provided in Sec.~\ref{subsec:singularity}. Finally, the collision loss carries the highest relative weight, as any collision directly compromises manufacturability. To prevent the optimization from being dominated by the collision-free term only, we employ a gradual weight-increasing strategy during the optimization process.

\begin{table}[]
    \centering
    \begin{tabular}{|c|c|c|c|c|c|}
    \hline
         Case& $w_\mathrm{lds}$& $w_\mathrm{lcr}$& $w_\mathrm{pds}$& $w_\mathrm{pcr}$&$w_\mathrm{cl}$\\ \hline
         $\mathcal{L}_\mathrm{SCF}$& 0.02& 0.2& -& -& 8000.0 $\to$ 40000.0\\ \hline
         $\mathcal{L}_\mathrm{DAC}$& 2.0& 0.2& -& -& 8000.0 $\to$ 40000.0\\ \hline
         $\mathcal{L}_\mathrm{TPD}$& 2.0& 0.2& 2.0& 0.001$\to$1.0& -\\ \hline
         $\mathcal{L}_\mathrm{MLC}$& 2.0& 0.2& -& -& 8000.0 $\to$ 40000.0\\ \hline
    \end{tabular}
    \caption{Summary of the combinations of regularization and collision-loss weights employed for the different applications on which our pipeline was tested. In certain cases, the arrow($\to$) symbol indicates that the weights are gradually increased during the optimization process to facilitate faster convergence. The rationale underlying the choice of these values is provided in Sec.~\ref{subsec:weights} and further discussed in relation to the results.}
    \label{tab:weights}
\end{table}


\subsection{Setup Optimization}

In certain examples that require support structures, we additionally optimize the printing setup. To achieve this, we introduce the setup loss defined in Sec.~\ref{subsec:platfrom_loss}:

\begin{equation}
\mathcal{L}_{ST} = \mathcal{L}_{sor} + 0.1\mathcal{L}_{spd} + 0.05\mathcal{L}_{cls},
\label{eqn:loss_setup_combined}
\end{equation}

Since the setup primarily depends on the final layer field, we incorporate this loss only during the later stages of the optimization process to improve computational efficiency.

\section{Implementation Details}
\label{sec:implementation}


\subsection{Network Hyperparameters}
\label{subsec:network}
We adopt the SIREN network architecture \cite{sitzmann_implicit_2020} owing to its proven capability to represent signals with high-frequency as well as the higher-order derivatives, both of which are critical to our formulations. The overall pipeline with the network employed in our implementation has been illustrated in Fig.~\ref{fig:pipeline}.

\revision{}{A consistent network structure and pipeline are utilized to represent the layer field ($f_l$), the toolpath field ($f_p$), and the implicit surface representation of the model ($sdf_{model}$). For the manufacturing field networks, each layer comprises 128 neurons. The network depth is set to 10 layers for the layer field ($f_l$) and 15 layers for the toolpath field ($f_p$). The rationale behind these architectural choices will be discussed in Sec.~\ref{subsec:singularity}. Following the configuration proposed by Sitzmann et al. \cite{sitzmann_implicit_2020}, the implicit model network ($sdf_{model}$) consists of 5 layers with 256 neurons each.}

The SIREN formulation introduces two additional hyperparameters: (1) $\omega_0$, the activation frequency scaling applied to the input layer, and (2) $\omega$, the activation frequency scaling for all subsequent layers. These parameters govern the representational frequency of the network and thus determine the rate at which the scalar field and its spatial derivatives can vary. Their influence on the network’s performance will be analyzed in Sec.~\ref{subsec:singularity}. In summary, these parameters regulate the smoothness of the learned fields and helps the representation of high-frequency geometric features, such as sharp curvatures or singularities, when required. The specific values used in this work are selected using empirical formula, which will also be discussed in Sec.~\ref{subsec:singularity}.


\subsection{Implicit Solid Representation}
\label{subsec:implicit_model_rep}
\begin{figure}
    \centering
    \includegraphics[width=0.99\linewidth]{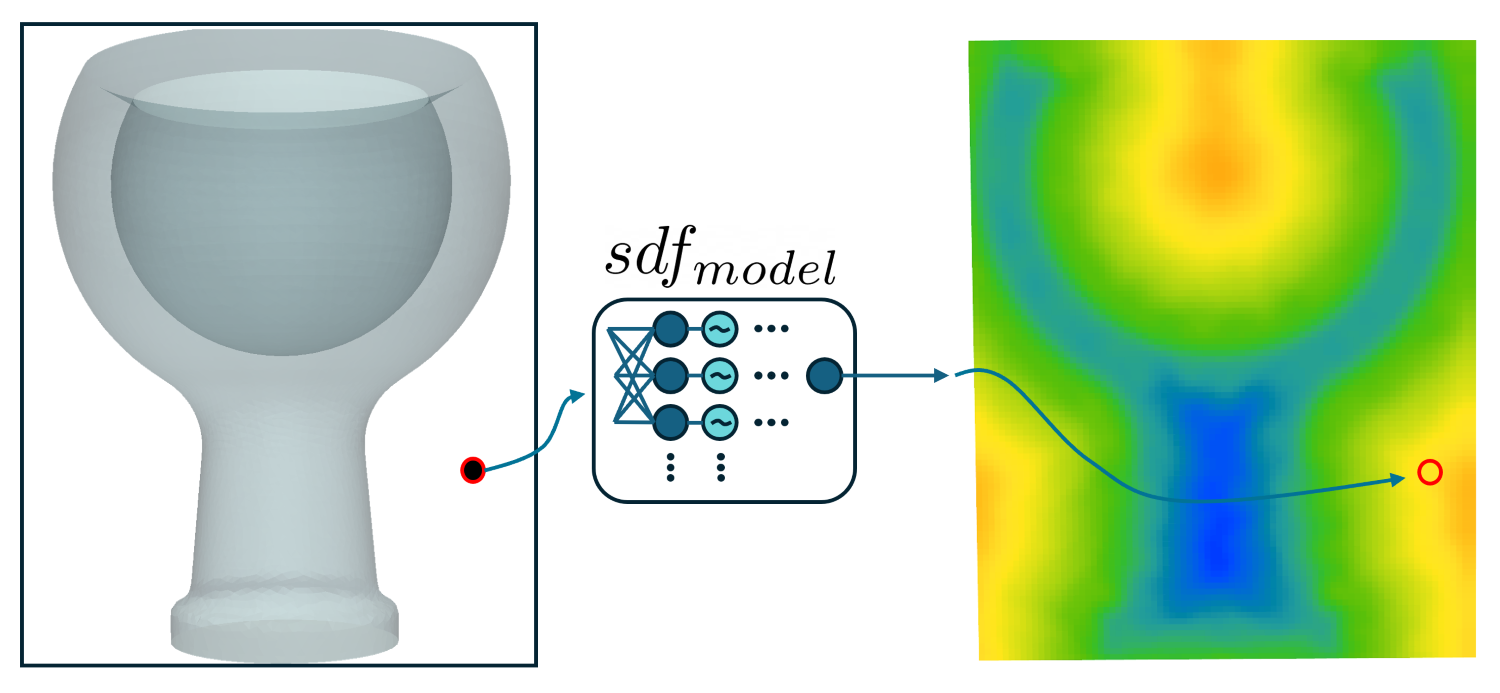}
    \put(-325,-9){\footnotesize \color{black}(a)}
    \put(-200,-9){\footnotesize \color{black}(b)}
    \put(-70,-9){\footnotesize \color{black}(c)}
    \caption{Figure showing our neural approximation of the SDF for the model in (a). We use a SIREN network (b) and optimize its parameters to output the SDF for any input point during the forward pass. (c) shows the SDF field over the space generated using the optimized $sdf_{model}$ network.}
    \label{fig:append_sdf1}
\end{figure}

As mentioned above, we also employ the use of an implicit representation of the part. This is especially useful for computing the collision loss as it allows us to make a very fast point-inclusion query (O(1)). We follow the approach of Sitzmann et al. \cite{sitzmann_implicit_2020} but modify the process slightly to suit our requirements. Specifically, we add loss to control the value of distance field at regions away from the surface as discussed below. We use a mesh representation as the input and convert it into an implicit SDF representation (Fig.~\ref{fig:append_sdf1} illustrates the process).

 We refer to the function based on the learned neural-network to be $sdf_{model}$ and $SDF$ to be the true signed distance field.  Our domain of consideration is the axis-aligned bounding box of the model. In order to train the network to represent the SDF, we use different loss functions defined on the scalar-field and gradient data. The losses used are described below. For brevity, we have omitted the $model$ subscript in $sdf_{model}$.

\begin{equation}
    \mathcal{L}_{s.scalar} = \frac{1}{|\{\partial\Omega\}|}\sum_{\mathbf{x}\in\{\partial\Omega\}}(sdf(\mathbf{x}))^2
\end{equation}
\begin{equation}
    \mathcal{L}_{s.gradient} = \frac{1}{|\{\partial\Omega\}|}\sum_{\mathbf{x}\in\{\partial\Omega\}}\|\nabla sdf(\mathbf{x}) - {\nabla SDF(\mathbf{x})} \|^2
\end{equation}
\begin{equation}
    \mathcal{L}_{d.scalar} = \frac{1}{|\{\mathcal{D}\}|}\sum_{\mathbf{x}\in\{\mathcal{D}\}}(sdf(\mathbf{x}) - SDF(\mathbf{x}) )^2
\end{equation}
\begin{equation}
    \mathcal{L}_{d.norm} = \frac{1}{|\{\mathcal{D}\}|}\sum_{\mathbf{x}\in\{\mathcal{D}\}}(1-\|\nabla sdf(\mathbf{x})\|)^2
\end{equation}

The total training loss is then defined as:

\begin{equation}
    \begin{split}
            \mathcal{L}= w_{s.scalar}.\mathcal{L}_{s.scalar}+\mathcal{L}_{s.gradient}\\+w_{d.scalar}.\mathcal{L}_{d.scalar}+\mathcal{L}_{d.norm}
    \end{split}
\end{equation}
In the above expressions, the $sdf$ is the learned model and $SDF$ is the true sdf function that is not known to us but we can determine its value at the sampled points. The values of ${\nabla SDF(\mathbf{x})}$ are required only on the model surface and are equal to the surface normals at those points. The domain $\mathcal{D}$ represent the entire bounding-box of the model (referred to as general-points) and the set $\partial \Omega$, represent the boundary surface of the model. The loss $\mathcal{L}_{s.scalar}$  is the scalar error on the surface points, $\mathcal{L}_{s.gradient}$  is the gradient error on the surface points, $\mathcal{L}_{d.scalar}$  is the scalar error on the general points and $\mathcal{L}_{d.norm}$  is the Eikonal error\cite{sitzmann_implicit_2020} on general points.

\begin{figure}[]
    \centering
    \includegraphics[width=0.80\linewidth]{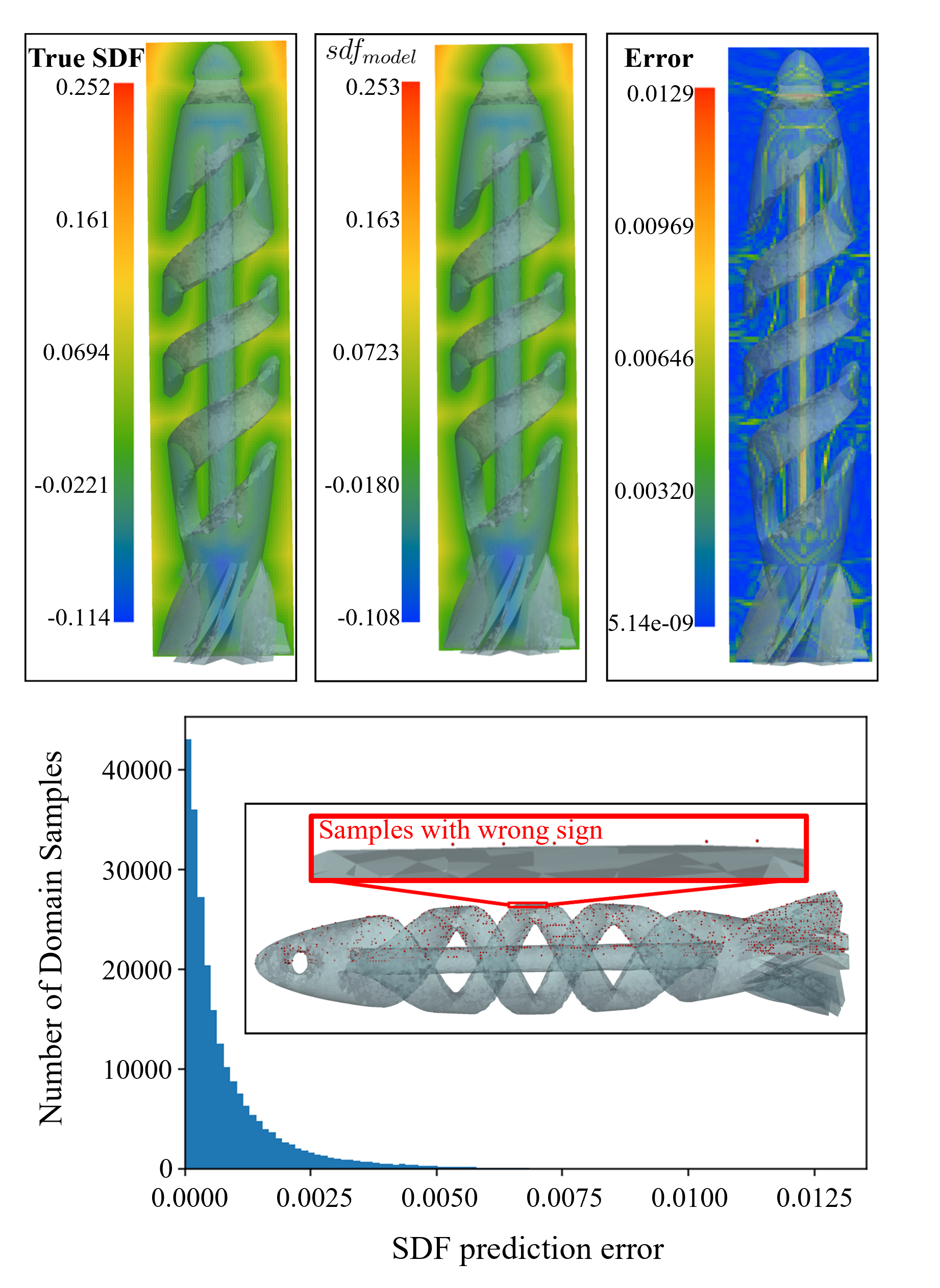}
    \put(-260,194){\small \color{black} (a)}
    \put(-158,194){\small \color{black} (b)}
    \put(-56,194){\small \color{black} (c)}
    \put(-150,-5){\small \color{black} (d)}
    \caption{Evaluation of the signed distance field network ($sdf_{model}$) for the Spiral-Fish geometry. (a) Ground-truth SDF and (b) corresponding predictions by the trained $sdf_{model}$ for point samples within the domain defined by the part’s bounding box. (c) Spatial distribution of the absolute error between the predicted and true SDF values. The error is most pronounced near regions corresponding to local extrema of the distance field, where the gradient magnitude vanishes, violating the Eikonal condition expected for a true distance field. (d) Histogram of the error distribution, with the inset highlighting the samples with incorrect sign. Note that such sign-inconsistent points remain confined to a narrow region near the boundary surface.}
    \label{fig:append_sdf2}
\end{figure}

 We use an Adam optimizer that we run for 150 epoches. The values of the weights are chosen as $w_{s.scalar}$ = $w_{d.scalar}$ = 100.0. Figure \ref{fig:append_sdf1} illustrates the implicit field for the Cup model, while Figure \ref{fig:append_sdf2} presents the accuracy of the approximated $sdf_{\text{model}}$ for the Spiral-Fish model.

\subsection{Differentiability}
\label{subsec:differentiability}
Here we analyze the differentiability of the proposed loss function for collision-avoidance. We begin with the case of layer–tool collisions in 3D printing, which corresponds to proving that Eq.~\eqref{eqn:main_collision_loss} is differentiable with respect to the network parameters at every point of evaluation. Specifically, this requires showing that the term $f_l(\mathbf{x}) - f_l(\mathbf{y})$ admits a computable derivative for all $\mathbf{x} \in \{\Omega\}$ and $\mathbf{y} \in \{\mathcal{T}_{x}\}$.

Let the network parameters be denoted by $\Theta$. Then, the layer field at any spatial location $\mathbf{x}$ can be expressed as $f_l(\mathbf{x}, \Theta)$. Accordingly,
\begin{equation}
    \frac{d (f_l(\mathbf{x}, \Theta)-f_l(\mathbf{y}, \Theta))}{d \Theta} = \frac{d (f_l(\mathbf{x}, \Theta)}{d \Theta} - (\frac{\partial (f_l(\mathbf{y}, \Theta))}{\partial \Theta} + \frac{\partial (f_l(\mathbf{y}, \Theta))}{\partial \mathbf{y}} \cdot\frac{d \mathbf{y}}{d \Theta} )
    \label{eqn:differentiability_collision1}
\end{equation}

\begin{figure}
    \centering
    \includegraphics[width=0.55\linewidth]{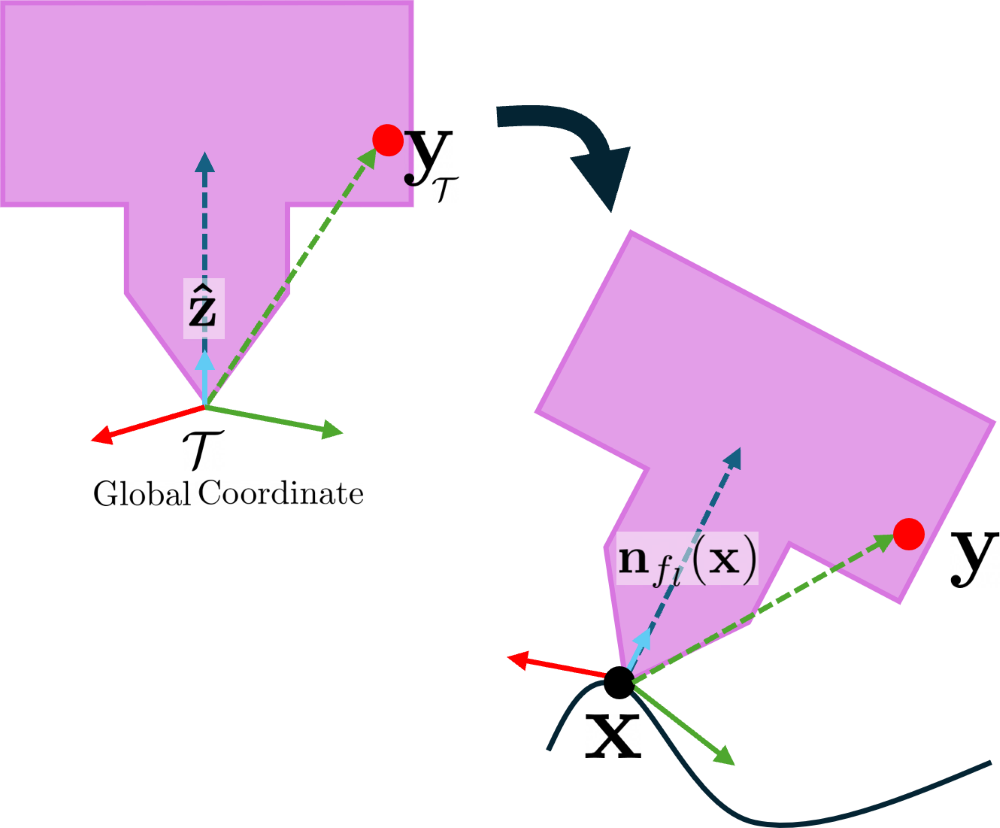}
    \caption{Transformation of tool to a local frame for collision-checking}
    \label{fig:inset_differentiability}
\end{figure}

Consider a point $\mathbf{y}$ in the tool configuration $\{\mathcal{T}_{x}\}$, which corresponds to $\mathbf{y}_{\mathcal{T}}$ in the tool’s home configuration $\mathcal{T}$ (coinciding with the world coordinates). Its position can be expressed as (see Fig.~\ref{fig:inset_differentiability}):
\begin{equation}
\begin{split}
    \mathbf{y} = \mathbf{x} + R(\mathbf{n}_{f_l})\cdot\mathbf{y}_{\mathcal{T}}\\
    \implies \mathbf{y} = \mathbf{x} + R(\frac{{\nabla{f_l(\mathbf{x},\Theta)}}}{\|\nabla{f_l(\mathbf{x},\Theta)}\|})\cdot\mathbf{y}_{\mathcal{T}},\\
\end{split}
\label{eqn:differentiability_toolPoint}
\end{equation}
where $R$ is the rotation function aligning the tool’s axis with the layer normal $\mathbf{n}_{f_l}(\mathbf{x})$. Taking derivatives with respect to the network parameters $\Theta$ gives:

\begin{equation}
\begin{split}
    \frac{d \mathbf{y}}{d \Theta} = \frac{\partial R}{\partial{\mathbf{n}_{f_l}}}\cdot\frac{d}{d \Theta}(\frac{{\nabla{f_l(\mathbf{x},\Theta)}}}{\|\nabla{f_l(\mathbf{x},\Theta)}\|})\cdot\mathbf{y}_{\mathcal{T}},
\end{split}
\label{eqn:differentiability_toolPoint_Derivative}
\end{equation}

Since $f_l$ is represented by a sinusoidally-activated neural network with parameters $\Theta$, we can compute $\frac{d}{d \Theta}(\nabla{f_l(\mathbf{x},\Theta)})$ via automatic differentiation. Consequently, $\frac{d \mathbf{y}}{d \Theta}$ and all other terms in Eq.~\eqref{eqn:differentiability_collision1} are differentiable with respect to $\Theta$. Therefore, the additive manufacturing collision loss is fully differentiable with respect to the network parameters.

Next, we show the differentiability of the loss for collision with the target part in milling. This amounts to proving that $sdf_{model}(\mathbf{y})$ is differentiable with respect to the network parameters $\Theta$:
\begin{equation}
    \begin{split}
        \frac{d (sdf_{model}(\mathbf{y}))}{d \Theta} = \frac{\partial (sdf_{model}(\mathbf{y}))}{\partial \mathbf{y}}\cdot \frac{d \mathbf{y}}{d \Theta},
    \end{split}
\end{equation}
where $\frac{\partial (sdf_{model}(\mathbf{y}))}{\partial \mathbf{y}}$ is the spatial gradient of the signed distance field. Since we represent the target part using an implicit neural network, this gradient can be computed directly via automatic differentiation. In the absence of an implicit representation, this term can be approximated by the direction toward the nearest surface point, or any suitable outward-pointing approximation.

All other loss functions in our formulation involve simple algebraic operations (products, dot/cross products) of the scalar field and its spatial derivatives. Therefore, they are inherently differentiable with respect to the network parameters $\Theta$.


\subsection{Sampling and Optimization Scheme}
\label{subsub:optimisationScheme}

As detailed in the network architecture, all field-function networks take three-dimensional coordinates $\{x,y,z\}$ as input. Manufacturing objectives, however, are not uniformly defined across the model. Global regularization constraints, such as layer or path spacing, curvature, and collision avoidance, apply throughout the domain, while support-free conditions are restricted to the model boundary, and direction-following requirements are enforced only at designated locations. During optimization, if a constraint is specified over a subset $\mathcal{R}$, the corresponding field function $f_i(\mathbf{x})$ or its derivatives must satisfy it for all $\mathbf{x} \in \mathcal{R}$. In practice, objectives are evaluated at a discrete set of sampled points, as enforcing constraints at every point is computationally prohibitive. Due to the inherent regularization and continuity of the networks \revision{}{(See the discussion on Network Frequency in Sec.~\ref{subsec:singularity})}, satisfying constraints at these finite points \revision{generally}{} ensures compliance across the intervening regions, supporting the validity of sparse sampling.

For objectives defined over the whole model, such as regularization and collision, we generate samples $\{\Omega\}$ on a grid within the axis-aligned bounding box and restricted to points inside the model. A point interval of 2 to 3.5 mm sufficiently resolves geometric features while remaining consistent with the network-frequency definition (Sec.~\ref{subsec:singularity}). Boundary-specific objectives, such as the support-free condition, use uniform surface sampling to capture relevant features. For direction-following objectives, we use maximum principal stress directions from \revision{}{Finite Element Analysis} (FEA). These dense, non-uniform data are downsampled via a farthest-point strategy to match the sampling-rate of $\{\Omega\}$, after which points in the region of interest are selected. All input points are centered and normalized by the maximal extent of the bounding box to restrict the domain to $[-1,1]^3$.

For collision detection, all points occupied by the tool must remain outside the part and obstacles (Section~\ref{subsec:collision}). Since tool rotation about its axis is unknown, we approximate it using axi-symmetric enclosure units (cylinders or cones) covering the relevant regions. Points within these units are sampled using fixed and perturbed positions to ensure coverage with fewer points, as illustrated in Fig.~\ref{fig:tool_sampling}; additional implementation details are provided in the \ref{appendix:tool_sampling} for completeness.

We used the Adam optimizer \cite{kingma2015adam}, training for up to 1100 epochs. An initial learning rate of around 4e-6 was observed to produce stable convergence and was reduced to 40\% every 150 epochs. The Milling case with fewer constraints converged in roughly 150 epochs, while Toolpath-Geometry control cases, optimizing two fields simultaneously, converged faster with a higher initial learning rate of 5e-5. A summary of the different parts that we used in our experiments is provided in Table \ref{tab:result_summary}. \revision{}{The optimization process was implemented using Python with the PyTorch \cite{paszke2019pytorch} framework utilized for the neural-network related implementation. The code was run on a Desktop PC with NVIDIA RTX 3070 graphics card, Intel i9-14900K (3.20 GHz) processor and 64.0 GB RAM. The PyVista \cite{sullivan2019pyvista} library was used to read and store the inputs (in form of mesh) and display the results.} The source code of our implementation will be released upon the acceptance of this paper.

\begin{figure}[!t]
    \centering
    \includegraphics[width=0.95\linewidth]{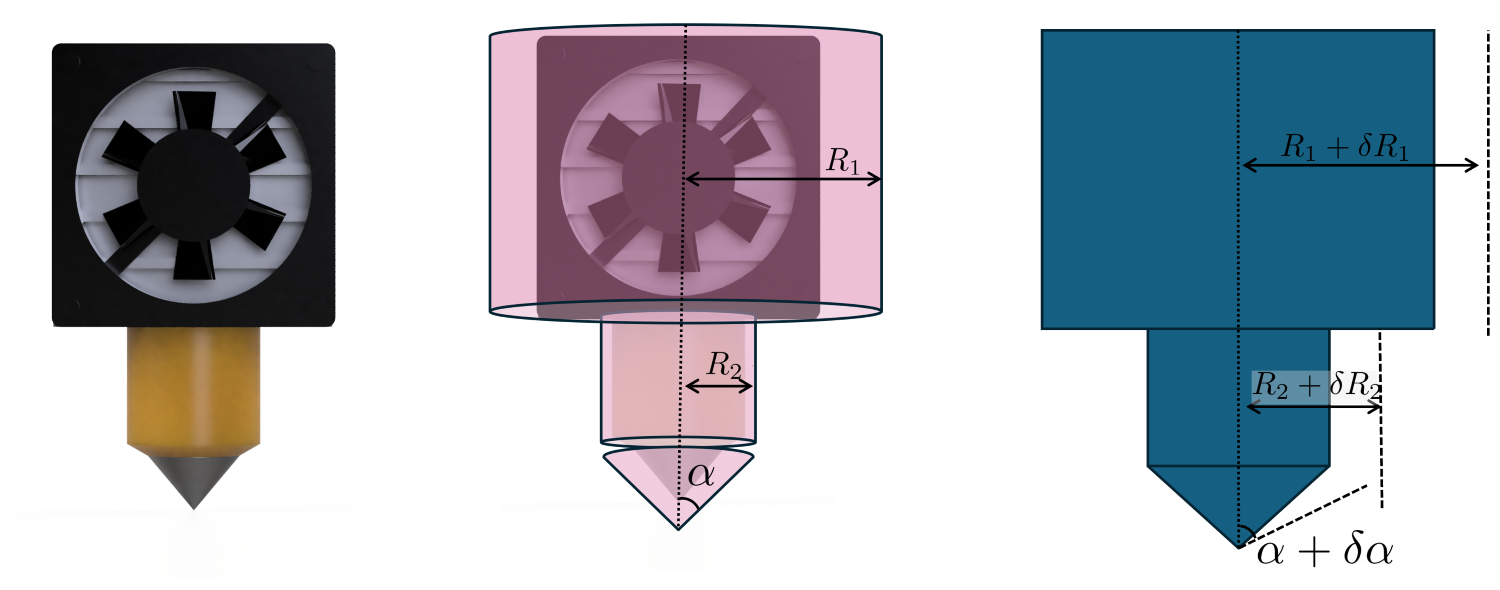}
    \put(-333,-7){\footnotesize \color{black}(a)}
    \put(-207,-7){\footnotesize \color{black}(b)}
    \put(-70,-7){\footnotesize \color{black}(c)}
    \caption{This figure presents the tool-geometry used in collision avoidance within our computational pipeline. (a) depicts a representation of the print head. To accelerate the process, 
    the tool is approximated using conical and cylindrical enclosure units, as illustrated in (b). To ensure safe collision avoidance, the dimensions of these units are slightly offset, and the resulting constraint is formulated as a soft constraint (c). Sampling is performed from these offset enclosure units to evaluate potential collisions}
    \label{fig:tool_sampling}
\end{figure}


\subsection{Physical Fabrication Setup}
To validate our proposed pipeline, we physically fabricated selected models covering the various applications discussed in this paper. The multi-axis motion was realized using a 6-DoF robotic arm (ABB IRB 2600) combined with a 2-DoF positioner (ABB A250). The material extrusion system was mounted as the end-effector of the robotic arm. The setup includes a dual-extrusion FDM (Fused Deposition Modeling) printer capable of depositing two materials, typically the base (matrix) and the support material. In addition, a continuous carbon-fiber printing head was integrated into the system.

\begin{figure}
    \centering
    \includegraphics[width=0.95\linewidth]{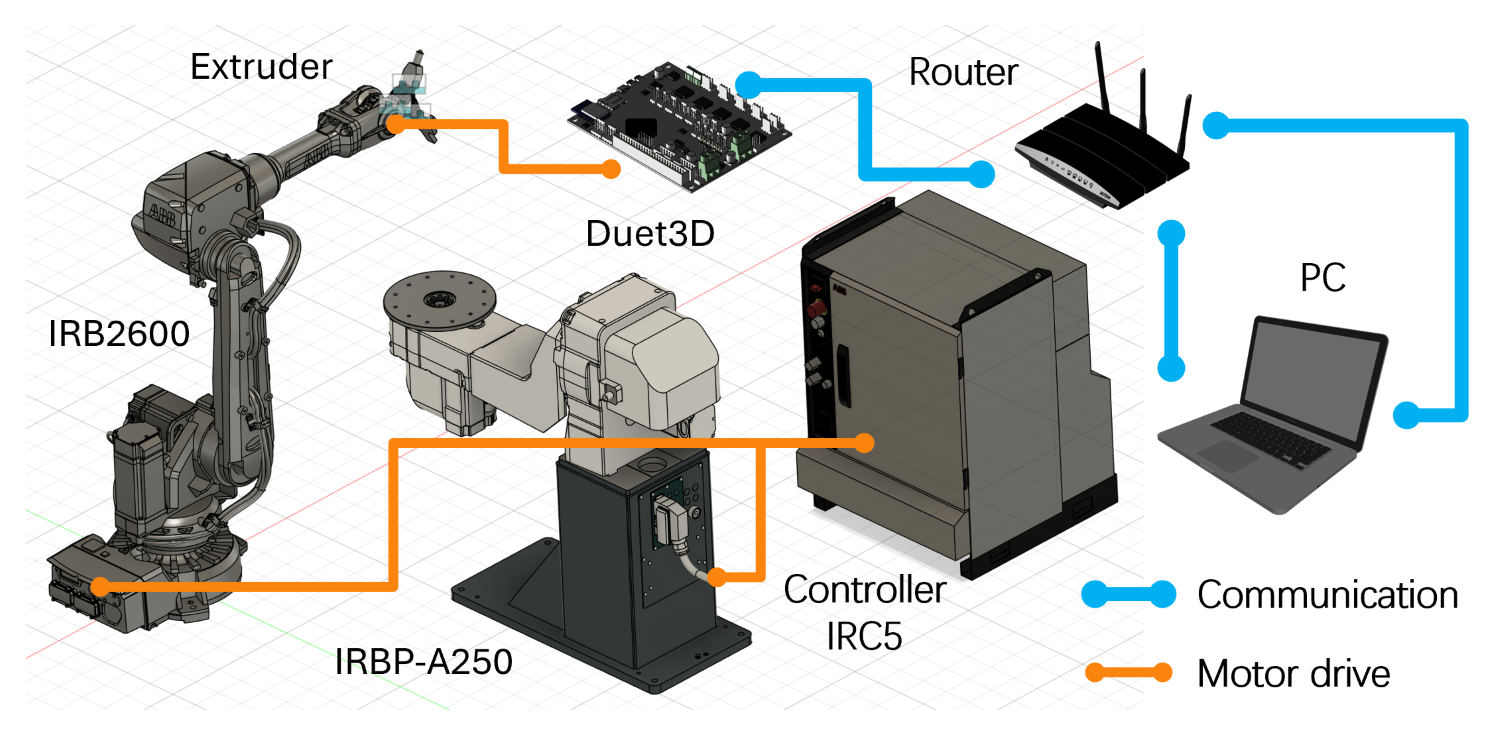}
    \caption{Illustration of the multi-axis additive manufacturing (printing) setup. A 6-DoF robotic arm (ABB IRB 2600) equipped with material extruders operates in conjunction with a 2-DoF positioner (ABB A250) that supports the printing platform. Motion coordination is managed by the IRC5 controller, which communicates with a PC and the Duet3D control-board for the extruders to ensure synchronized deposition during printing. }
    \label{fig:physical_setup}
\end{figure}

The motions of the robotic arm and positioner were coordinated through an ABB IRC5 controller, while the extrusion processes were managed by a Duet3D control board. Synchronization between the arm motion and the extrusion rate was achieved through custom control scripts. An illustration of our FDM-based setup is shown in Fig.~\ref{fig:physical_setup}.

Polylactic acid (PLA) was used as the primary printing material (and as the polymer matrix when used with carbon fiber), while polyvinyl alcohol (PVA) served as the support material in cases where the support-free constraint was not enforced. The continuous carbon fiber used was Markforged CF-FR-50 with a nominal diameter of 0.37 mm.

For the milling example, we employed a modified version of the 5AxisMaker 5XM600XL machine equipped with a rotary base. The milling process was demonstrated on foam material to validate collision-free volume milling.

\section{Case Studies and Results}
\begin{table}[]
    \centering
    \begin{tabular}{|c||c|c|c|c|c|c|}
        \hline
         & & & B.Box Size&\multicolumn{2}{c|}{Points \#}&Opt. Time\\ 
         \cline{5-6} 
         Case& Model& Fig.&(mm)&Reg.&Func.&(mins)\\ \hline \hline
         \multirow{3}{*}{$\mathcal{L}_{SCF}$}& Fertility & \ref{fig:result_fertility_main}& 150.00 &9897&18744&161.8
         \\ \cline{2-7} 
         & 4C &\ref{fig:results_4c} &50.00&4932&12931&67.05
         \\ \cline{2-7} 
         & Spiral-Fish& \ref{fig:result_spiral_norm}&300.0&9707&19925&146.4
         \\ \hline

         \multirow{1}{*}{$\mathcal{L}_{DAC}$}& Fork & \ref{fig:results_clip_layer}& 143.00&12336&2077&128.7
         \\ \hline

         \multirow{3}{*}{$\mathcal{L}_{TPD}$}& T-Bracket & \ref{fig:result_t_complete}& 130.00&10459&8865& 52.5
         \\ \cline{2-7} 
         & S-Curve& \ref{fig:result_tp_cooptimisation}&150.00&11079&11040&45.1
         \\ \cline{2-7} 
         & Flat Bracket & \ref{fig:curvatureFiltering}& 210.00 &16725&14654&92.7
         \\ \hline

         \multirow{1}{*}{$\mathcal{L}_{MLC}$}& Cup & \ref{fig:result_milling}& 100.00 &37970&-&45.9
         \\ \hline
         
    \end{tabular}
    \caption{List of tested models along with their corresponding figure references, organized according to their multi-axis manufacturing requirements. The maximum bounding box(B.Box) dimension of each model is reported, as this parameter determines the selection of frequency-scaling factors ($\omega_i$, see Sec.~\ref{subsec:singularity}) for the SIREN networks, which play a critical role in the optimization process. The number of sampled points used for evaluating the regularization and collision losses (Reg.) and the functional losses (Func.) are also provided for reference. The time for the standard cycles of optimization are also reported.}\label{tab:result_summary}
\end{table}

The following subsections present the results for different application cases of the pipeline. A supplementary video offering a brief overview of the pipeline and demonstrations of selected manufacturing processes is also provided with this paper.


\subsection{Support and Collision Free}
\label{subsec:result_support}

\begin{figure}[!t]
    \centering
    \includegraphics[width=1.0\linewidth]{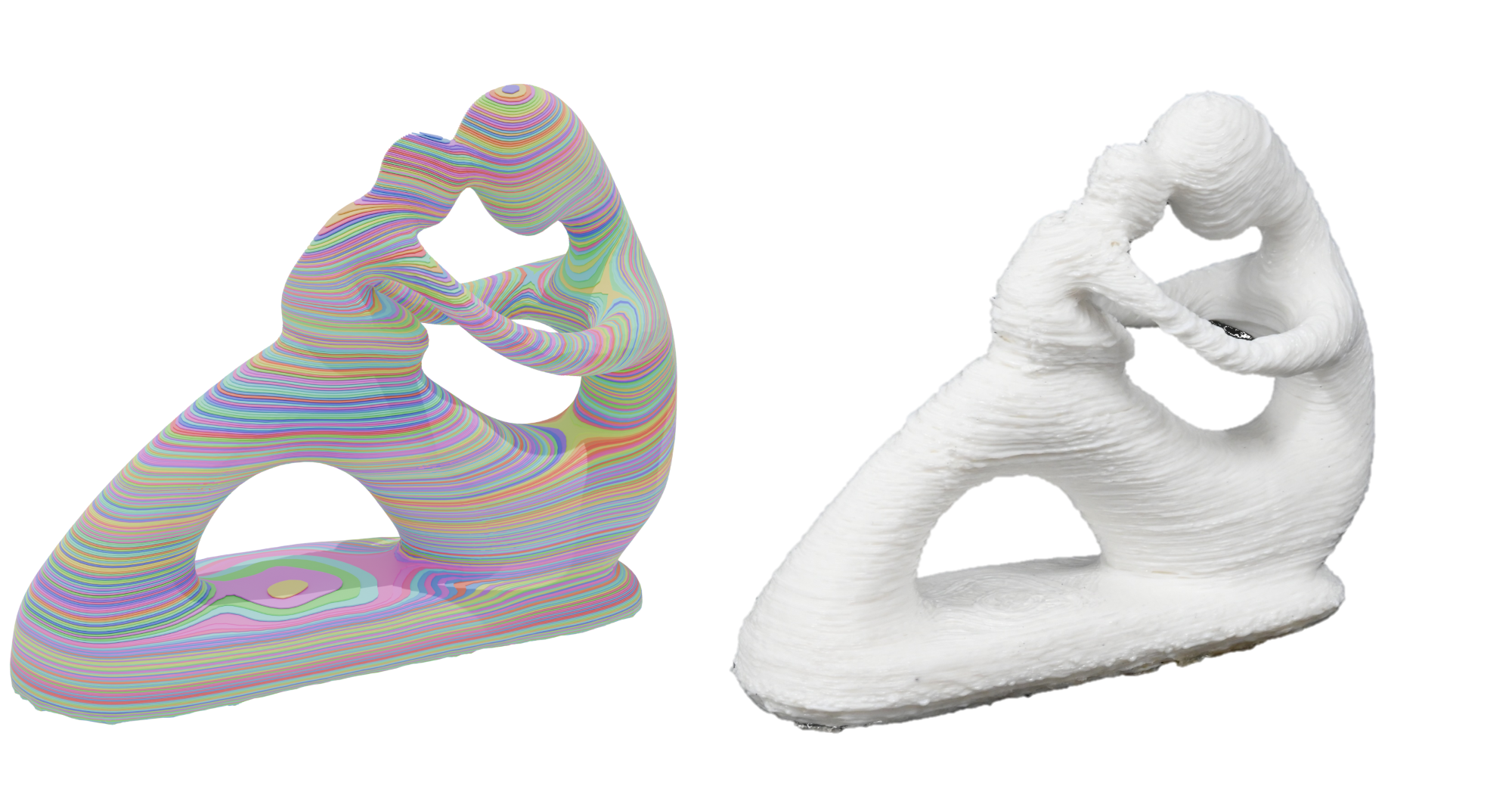}
    \put(-300,0){\footnotesize \color{black}(a)}
    \put(-100,0){\footnotesize \color{black}(b)}
    \caption{(a) Layers for the Fertility model obtained using our method (by minimizing $\mathcal{L}_{SCF}$, Eq.~\eqref{eqn:total_loss_sf}), which allows fabrication without the need for support structures while avoiding any potential collisions. The generated layers were directly employed to produce the printed part as shown in (b), without requiring any post-processing for collision removal.}
    \label{fig:result_fertility_main}
\end{figure}

\begin{figure}[t]
    \centering
    \includegraphics[width=1.00\linewidth]{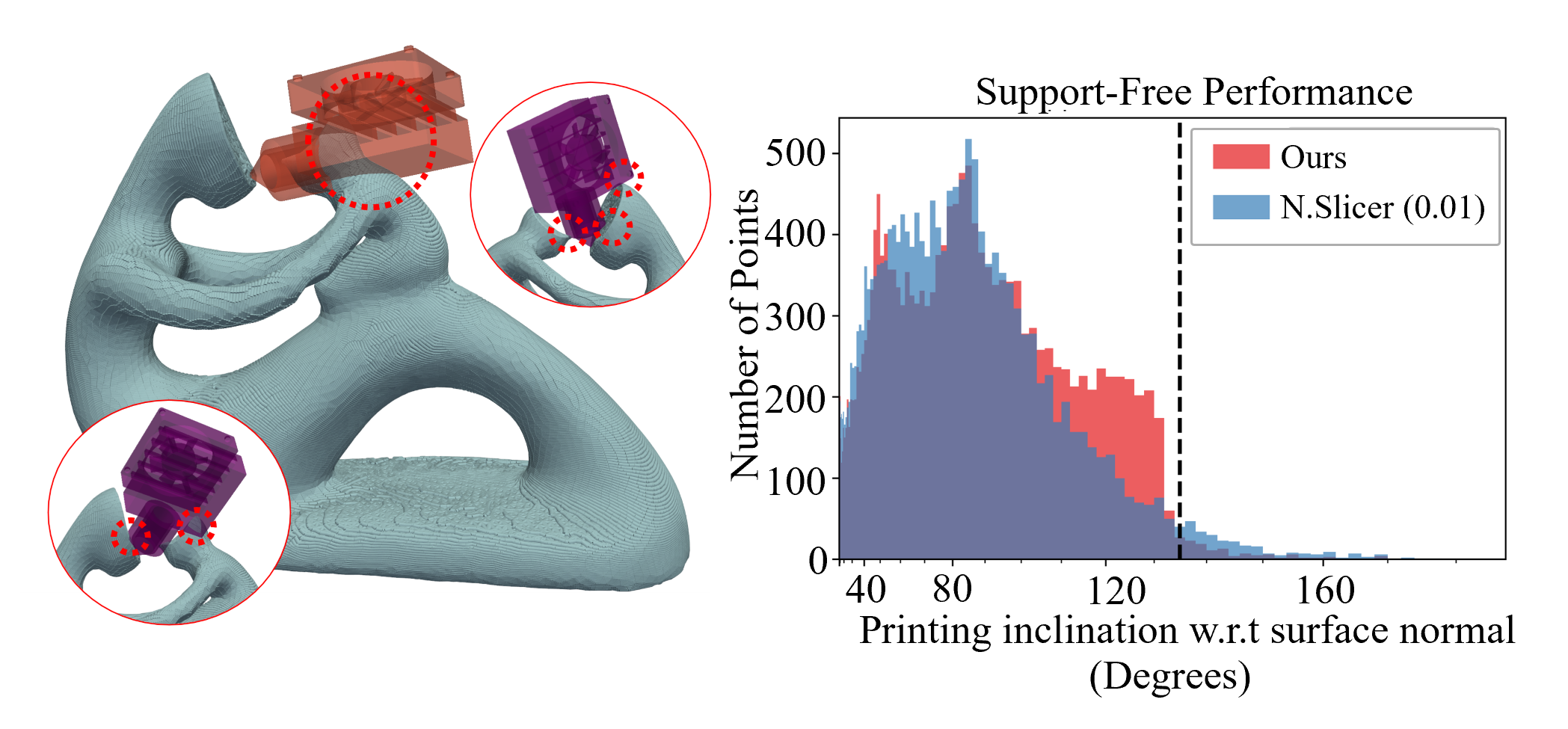}
    \put(-300,-3){\small \color{black}(a)}
    \put(-100,-3){\small \color{black}(b)}
    \put(-372,65){\small \color{black}i}
    \put(-228,136){\small \color{black}ii}
    \caption{This figure presents a comparison of the Fertility model layers generated using Neural Slicer \cite{liu_neural_2024} with the Quaternion Harmonic weight of 0.01 (a) as indirect curvature control. We select a result exhibiting support-free performance comparable to ours (b). However, as shown in (a), the generated layers lead to severe collisions and intersections with the tool. Moreover, as illustrated in the two zoom-views (a(i) \& a(ii)), such collisions cannot be resolved through post-processing because no valid configuration without collision could be obtained when using these layers. The black dashed-line on the histogram (b) shows the desired support-free threshold of $135^\circ$, which shows that although the result of the Neural Slicer does not need support, it cannot be manufactured because of collision.}
    \label{fig:collision_fertility_1}
\end{figure}

\begin{figure}[t]
    \centering
    \includegraphics[width=1.00\linewidth]{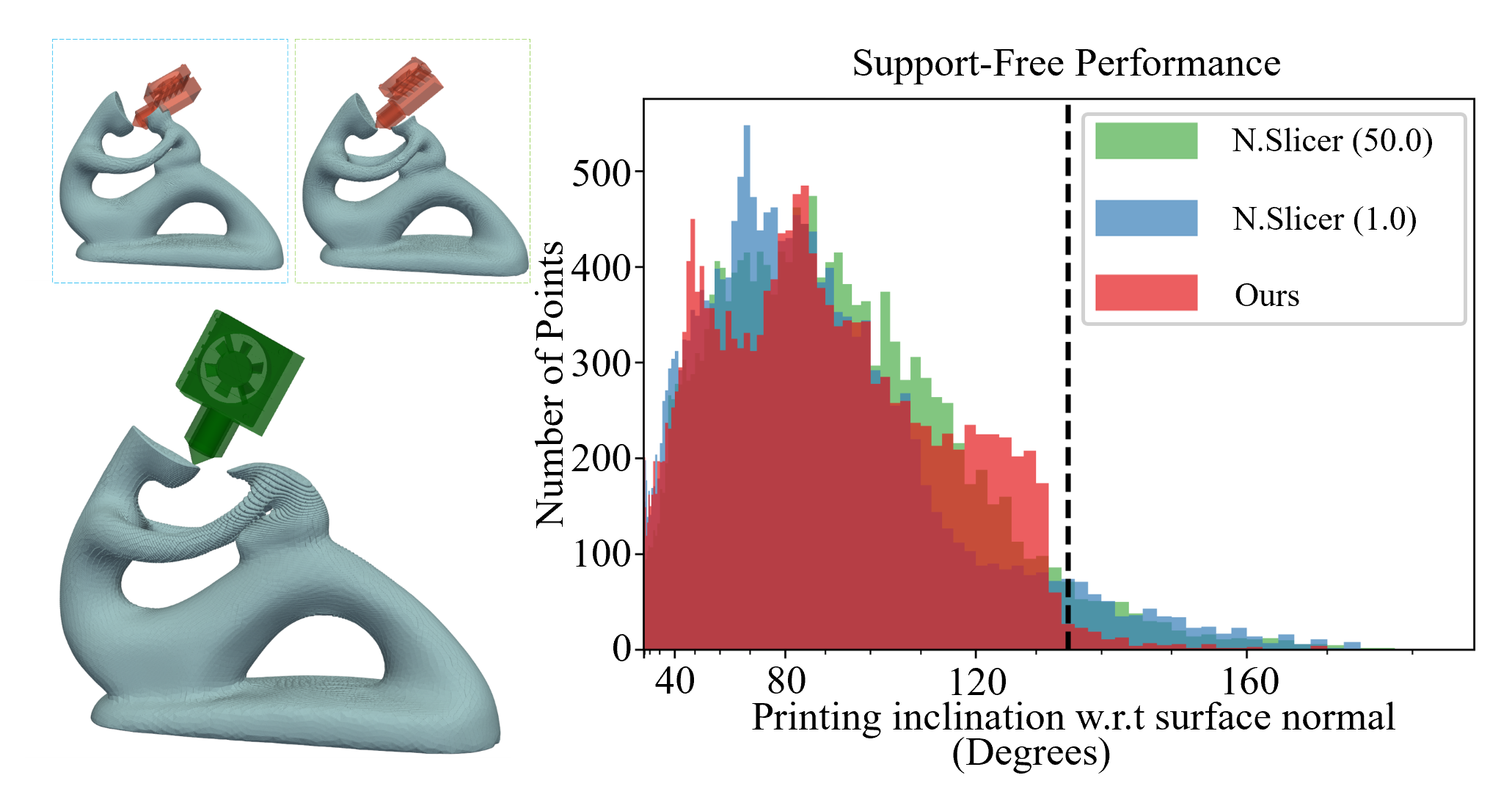}
    \put(-387,126){\small \color{black}(a1)}
    \put(-320,126){\small \color{black}(a2)}
    \put(-370,-3){\small \color{black}(a3)}
    \put(-126,-3){\small \color{black}(b)}
    \caption{This figure presents an additional comparison with the results of Neural Slicer. To mitigate collisions, we increased the Harmonic (curvature) Loss weight term in their formulation to reduce curvature. Subfigures (a1) and (a2) show the outcomes for weight values of 1.0 and 50.0, respectively, where collisions with the tool remain present. As shown in (b), this adjustment also causes a substantial deterioration of the support-free properties, while still failing to eliminate all collisions. This is because the collision region is already relatively flat locally, making curvature a poor indicator for such cases. In contrast, our result (a3) and (b) successfully avoids such collisions while maintaining support-free performance. The black dashed line in the histogram (b) indicates the desired support-free threshold of $135^\circ$. }
    \label{fig:collision_fertility_2}
\end{figure}

\begin{figure}[p]
    \centering
    \includegraphics[width=0.85\linewidth]{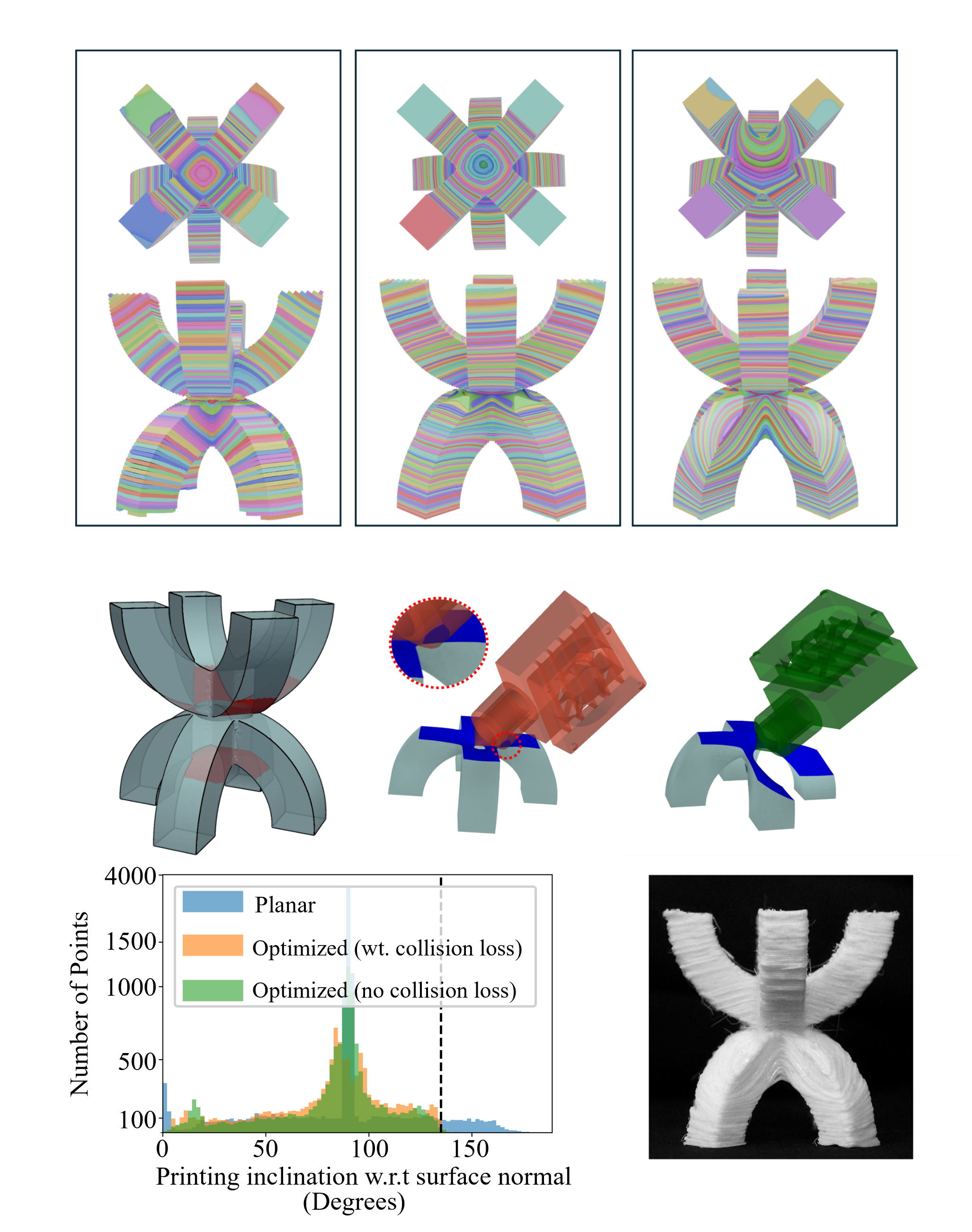}
     \put(-272,233){\footnotesize \color{black}(a)}
    \put(-172,233){\footnotesize \color{black}(b)}
    \put(-75,233){\footnotesize \color{black}(c)}
        \put(-295,133){\footnotesize \color{black}(d)}
    \put(-162,133){\footnotesize \color{black}(e)}
    \put(-52,133){\footnotesize \color{black}(f)}
         \put(-220,-10){\footnotesize \color{black}(g)}
    \put(-68,-10){\footnotesize \color{black}(h)}
 
    \caption{Results and comparisons for the 4C model (d), which we attempt to print without support. The red region in (d) indicates the area that would require support in a planar print. (a) shows the result obtained using the $S^3$-Slicer, which is similar to the result in (b) obtained by our method without applying the collision loss $\mathcal{L}_{cl}$. (c) presents the layers generated with the collision loss included. For the layers given in (b), a potential point with collision is as demonstrated in (e). (f) shows the corresponding region from (c), verifying the effectiveness of the collision loss in producing a collision-free solution while not sacrificing the support-free quality -- see (g) for the histogram for support-free. Finally, (h) shows the physically fabricated part.}
    \label{fig:results_4c}
\end{figure}

\begin{figure}
    \centering
    \includegraphics[width=0.9\linewidth]{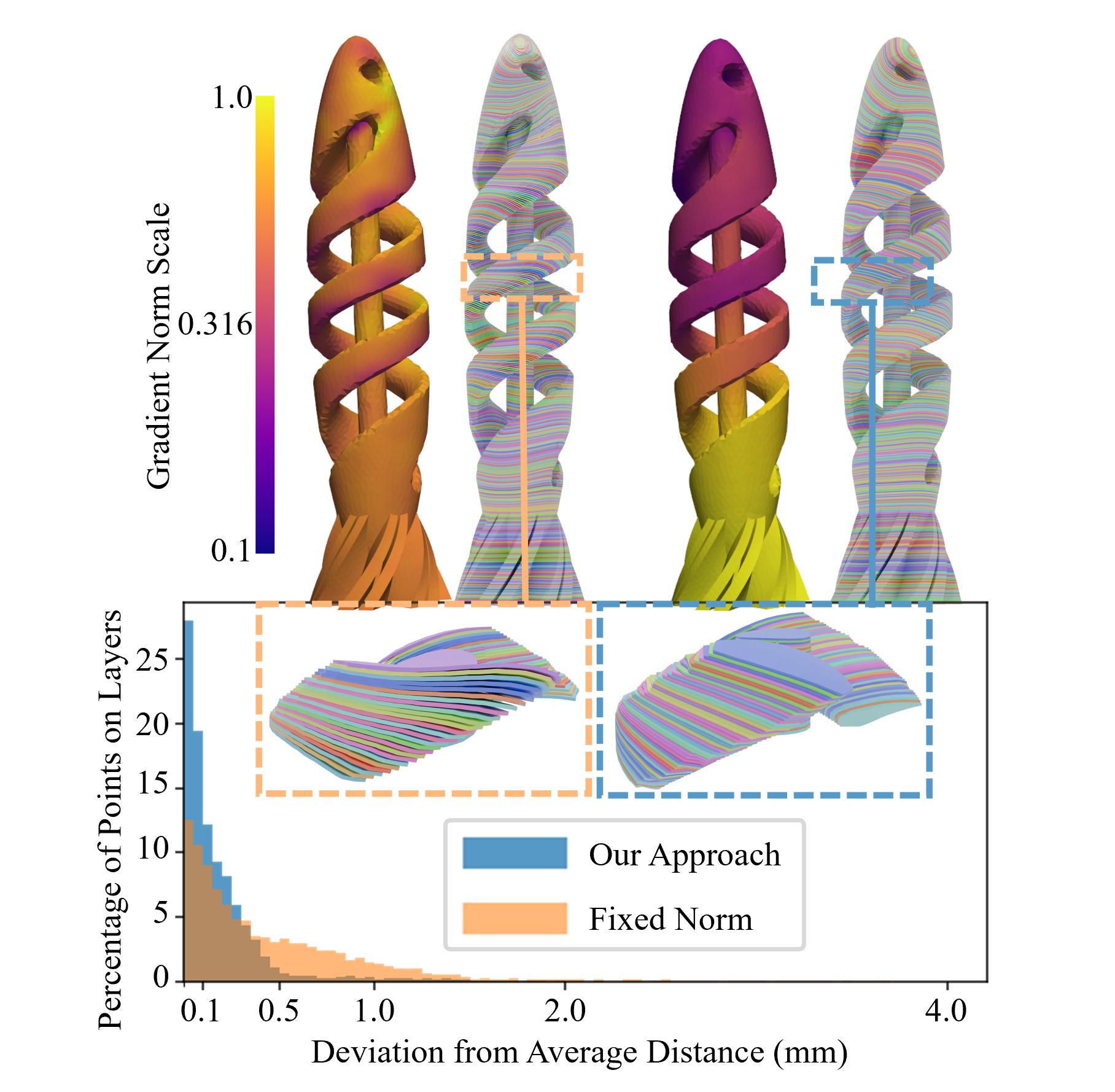}
    \put(-270,319){\footnotesize \color{black}(a1)}
    \put(-215,319){\footnotesize \color{black}(a2)}
    \put(-97,319){\footnotesize \color{black}(b2)}
    \put(-155,319){\footnotesize \color{black}(b1)}
    \put(-170,-4){\footnotesize \color{black}(c)}
    \put(-186,97){\footnotesize \color{black}(a3)}
    \put(-80,97){\footnotesize \color{black}(b3)}
    \caption{Comparison of gradient–norm control strategies for generating support-free layers in the \textit{Spiral-Fish} model. (a1–a3) Direct enforcement of a unit gradient norm ($\|\nabla f_l\| = 1$) versus (b1–b3) control of its spatial derivative $\frac{d(\|\nabla f_l\|)}{dk}$, as introduced in $\mathcal{L}_{\mathrm{lds}}$ (Eq.~\eqref{eqn:layer_density_loss}).(a1) and (b1) visualize $\|\nabla f_l\|$, showing that derivative-based control enables distinct regional growth rates of $f_l$. (a2) and (b2) depict the resulting layer structures, with corresponding zoomed-in views highlighting differences in spacing behaviors. (c) Histograms quantify deviations in layer-to-layer distances for six uniformly sliced layers within the selected regions. Overall, derivative-based control ((b2), (c)) produces helical regions with notably more uniform spacing, whereas direct norm control ((a2), (c)) fails to achieve comparable regularity.}
\label{fig:result_spiral_norm}

\end{figure}

We first examine SCF printing to evaluate the effectiveness of our simultaneous collision avoidance formulation (using $\mathcal{L}_{SCF}$, Eq.~\eqref{eqn:total_loss_sf}). We present results for the Fertility Model (Figs.~\ref{fig:result_fertility_main}–\ref{fig:collision_fertility_2}) and the 4C Model (Fig.~\ref{fig:results_4c}), along with an ablation study using the Spiral-Fish Model (Fig.~\ref{fig:result_spiral_norm}). Details are discussed below.

The Fertility Model has a spatial range of approximately 150 mm in its axis-aligned bounding box, and a normalization scaling parameter of 75 is therefore used to confine it within the  domain as $[-1, 1]$. The frequency scaling parameters of the network are defined relative to this scaling factor. For this model, we set $\omega_o = \omega = 7.0$ ($\approx$ $\text{scale}/10$). Similarly, the parameters for the 4C and Spiral-Fish models are $\omega_o = \omega = 3.0$ and $15.0$, respectively.

For optimization, we employ the total loss function $\mathcal{L}_{SCF}$, defined in Sec.~\ref{sec:total_losses}. The layer field is initialized as a planar height field (see Sec.~\ref{subsec:singularity} for the rationale).
Since the requirements are layer-specific, we use hybrid contour-parallel and zig-zag toolpaths, following Zhang et al.~\cite{zhang_toolpath_2025}.

In Fig.~\ref{fig:result_fertility_main}(a) we show the resultant layers for the Fertility model and the result of the physical fabrication in Fig.~\ref{fig:result_fertility_main}(b).

While previous approaches (e.g., \cite{zhang_zhangty019s3_deformfdm_2025, li_vector_2022, liu_neural_2024}) have demonstrated multi-axis, support-free 3D printing, they do not explicitly handle direct collision avoidance. To highlight the advantage of our method, we compare against the recently published Neural Slicer~\cite{liu_neural_2024}, which achieves similar support-free behavior but lacks an integrated collision-loss formulation (Fig.~\ref{fig:collision_fertility_1}(b)).

Previous strategies~\cite{zhang_zhangty019s3_deformfdm_2025, liu_neural_2024} often adjust tool orientations in post-processing steps to mitigate collisions. However, as illustrated in Fig.~\ref{fig:collision_fertility_1}(a), there may exist points with no feasible or accessible tool orientations. Moreover, such adjustments can compromise support-free performance.
Another common strategy restricts layer curvature to reduce local collisions. Although this helps prevent collisions caused by sharp local geometry, it does not account for non-local tool-part interactions.
As shown in Fig.~\ref{fig:collision_fertility_2}, increasing the weight of the Quaternion-Harmonic loss indeed reduces curvature globally, but this indiscriminate effect deteriorates support-free quality without fully eliminating collisions. In contrast, our approach achieves collision-free printing while preserving support-free performance, as demonstrated in Fig.~\ref{fig:collision_fertility_2}.

Figure~\ref{fig:results_4c} shows the results for the 4C Model. For comparison, we include results from the $S^3$-Slicer~\cite{zhang_zhangty019s3_deformfdm_2025} and our method without the collision-loss term ($\mathcal{L}_{cl}$), optimized under identical settings. Both of these produce layer configurations that result in tool collisions.
Our full SCF formulation, however, successfully avoids all collisions while maintaining support-free layer orientations. This improvement arises because the collision-loss term helps escape local minima corresponding to collision-prone configurations, guiding the optimization toward a valid and manufacturable solution under the same training conditions.

In Section \ref{subsub:layer_density}, we introduced a layer-distance uniformity loss that avoids enforcing a fixed norm. To further examine the effect of this layer-distance regularization, we compare the outcomes of directly constraining the gradient norm ($\|\nabla f_l\| = 1$) against those of minimizing its spatial derivative, as formulated in $\mathcal{L}_{lds}$ (Eq.~\eqref{eqn:layer_density_loss}). Fig.~\ref{fig:result_spiral_norm} presents the results for this study on the Spiral-Fish model. When the gradient norm is directly fixed to unity, the resulting layers (or the layer field) exhibit coupled variation between the central cylindrical and helical regions, preventing independent twisting of the layer in the helices. In contrast, our proposed derivative-based control allows the rate of field growth to vary spatially, thereby enabling the field in the helical regions to twist freely while maintaining uniform layer spacing (Fig.~\ref{fig:result_spiral_norm} (c)). This difference is evident from the corresponding $\|\nabla f_l\|$ (gradient-norm, Fig.~\ref{fig:result_spiral_norm}(a1,b1)) distributions, which show region-specific variations in the field’s rate of change under our formulation. Consequently, the derivative-based loss provides a more flexible and spatially adaptive control mechanism for achieving uniformly spaced, support-free layer generation.

\subsection{Direction Alignment with Collision Avoidance}

\begin{figure}[!t]
    \centering
    \includegraphics[width=0.95\linewidth]{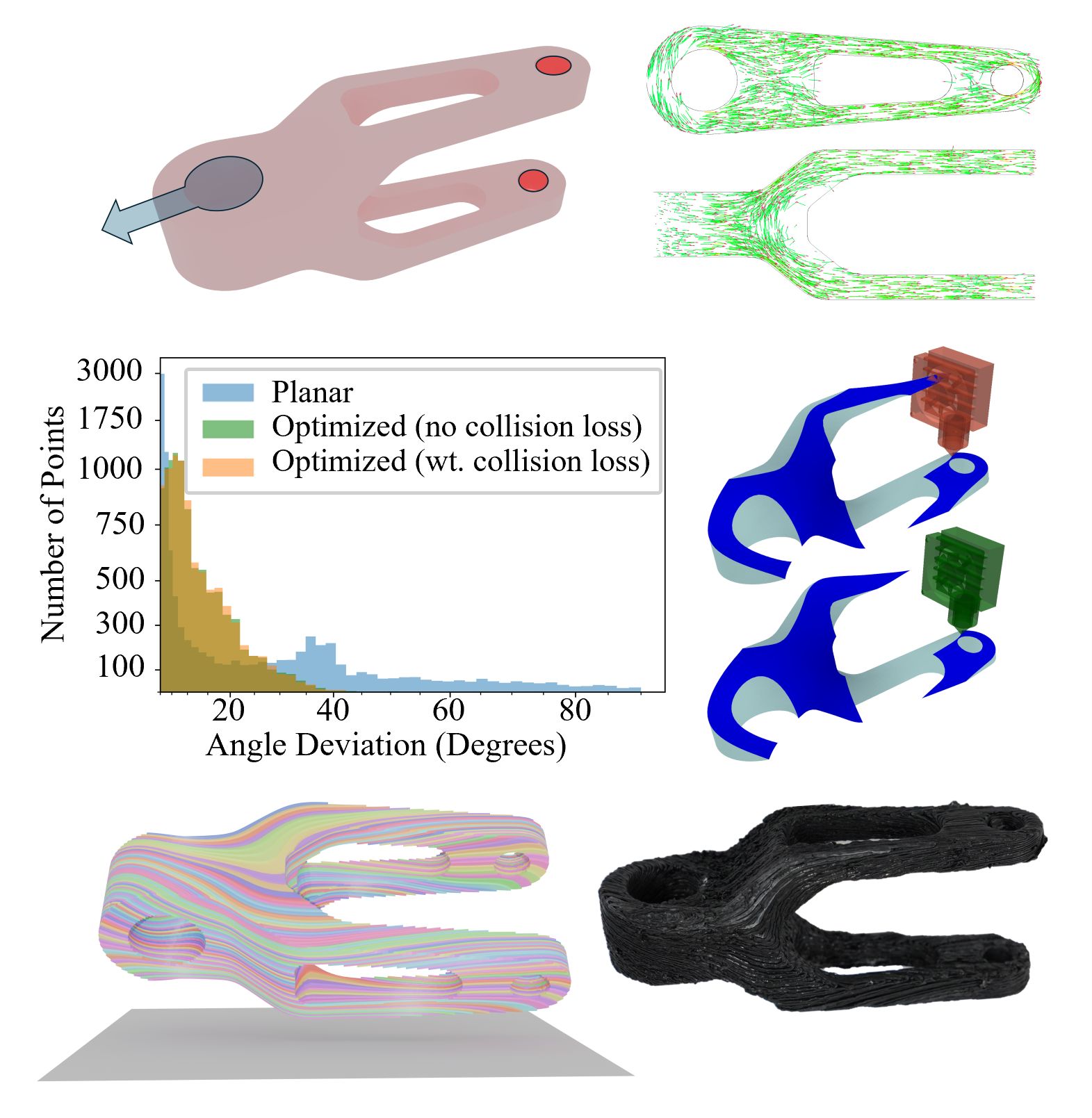}
    \put(-300,273){\footnotesize \color{black}(a)}
    \put(-132,273){\footnotesize \color{black}(b)}
    \put(-330,120){\footnotesize \color{black}(c)}
    \put(-140,175){\footnotesize \color{black}(d1)}
    \put(-140,120){\footnotesize \color{black}(d2)}
    \put(-330,20){\footnotesize \color{black}(e)}
    \put(-150,20){\footnotesize \color{black}(f)}
    \caption{Results and analysis for the Fork model. (a) Model geometry with applied loading conditions. (b) Maximum principal stress directions. (c) Direction alignment quality. Intermediate tool positions obtained (d1) without and (d2) with collision loss are shown as: (d1) the tool intersects with the part, whereas in (d2) collision loss prevents intersection without compromising alignment, proved by the histogram given in (c). (e) Optimized layer configuration with setup. (f) Fabricated part.} 
    \label{fig:results_clip_layer}
\end{figure}

We demonstrate the applicability of our collision-free multi-axis process planning with layer-direction alignment in the context of strength-reinforced slicing \cite{fang_reinforced_2020, zhang_zhangty019s3_deformfdm_2025, liu_neural_2024, li_vector_2022}. Specifically, we consider a Fork Model subjected to the loading conditions as shown in Fig.~\ref{fig:results_clip_layer}(a), where the goal is to align all tensile maximum principal stresses to the layer surfaces tangentially(Fig.~\ref{fig:results_clip_layer}(b)).

The optimization process follows the same formulation as in the support-free case, except that the support-free objectives are replaced by the direction-alignment constraint. We employ the total loss $\mathcal{L}_{DAC}$ (Eq.~\eqref{eqn:total_loss_str}), combined with the setup optimization loss $\mathcal{L}_{ST}$ (Eq.\eqref{eqn:loss_setup_combined}). The results of our slicing optimization are shown in Fig.~\ref{fig:results_clip_layer}(e), along with the corresponding optimized platform configuration. The physically fabricated part is shown in Fig.~\ref{fig:results_clip_layer}(f).

To highlight the contribution of our collision-avoidance formulation, we perform an ablation study comparing results with and without the collision-loss term $\mathcal{L}_{cl}$ (Fig.~\ref{fig:results_clip_layer}(d1,d2)).
As observed, including the collision-loss eliminates tool-part collisions while preserving high-quality direction alignment, as also evidenced by the histogram in Fig.~\ref{fig:results_clip_layer}(c).
This demonstrates that the collision-loss term effectively complements other functional objectives, such as direction alignment, without degrading their performance.

Although this part required support material, we did not explicitly generate support structures in the current pipeline. Consequently, supports were not included in the collision-avoidance optimization. Instead, we increased the tool-enclosure radius threshold to provide sufficient clearance for the anticipated supports. Combined with a small adjustment of tool orientation (minor head tilting), this gives collision-free motion even in the presence of support layers. \revision{}{The support layers were generated using the method of Zhang et al.~\cite{zhang_support_2023}, as it produces stable supports, compatible with curved layers defined by scalar fields.}
The effectiveness of this approach has been validated through physical fabrication, as shown in Fig.~\ref{fig:results_clip_layer}(f).


\subsection{Toolpath Geometry Control}
\label{subsec:result_tp_geom}
\begin{figure}[!t]
    \centering
    \includegraphics[width=1.00\linewidth]{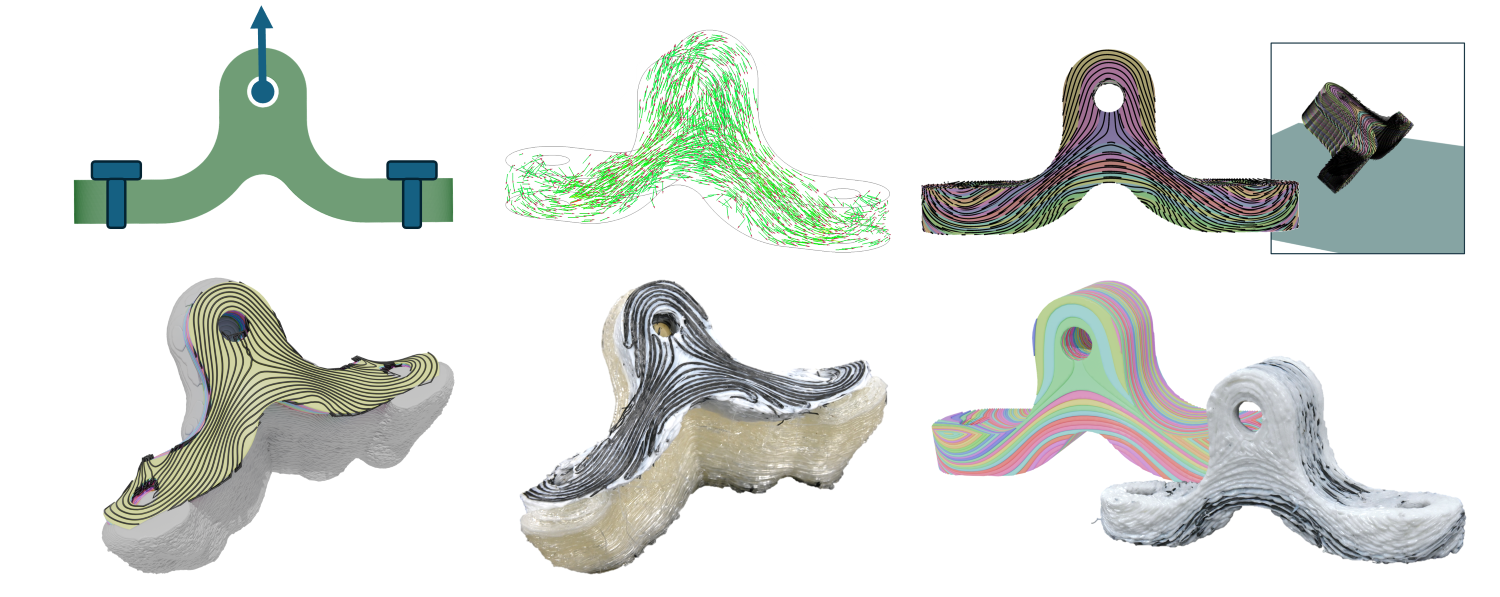}
    \put(-380,80){\footnotesize \color{black}(a)}
    \put(-266,80){\footnotesize \color{black}(b)}
    \put(-153,80){\footnotesize \color{black}(c)}
    \put(-380,3){\footnotesize \color{black}(d)}
    \put(-266,3){\footnotesize \color{black}(e)}
    \put(-153,3){\footnotesize \color{black}(f)}
    \caption{This figure presents our result for Toolpath–Geometry–driven optimization applied to the T-Bracket model. (a) shows the part under the specified loading condition, with the corresponding maximal tensile principal-stress directions as illustrated in (b). Using these directions as the target, we generate the layers and toolpaths depicted in (c), which also includes the optimised setup, where both the  orientation and the location of the model w.r.t. the building platform are optimized. (d) highlights a layer with the continuous carbon-fiber toolpath (black), and its corresponding fabricated stage is as shown in (e). \revision{}{The white color in (d) represent the support structure.} Finally, (f) presents the complete part, including both the simulated (left) and fabricated (right) results}
    \label{fig:result_t_complete}
\end{figure}

\begin{figure}
    \centering
    \includegraphics[width=.90\linewidth]{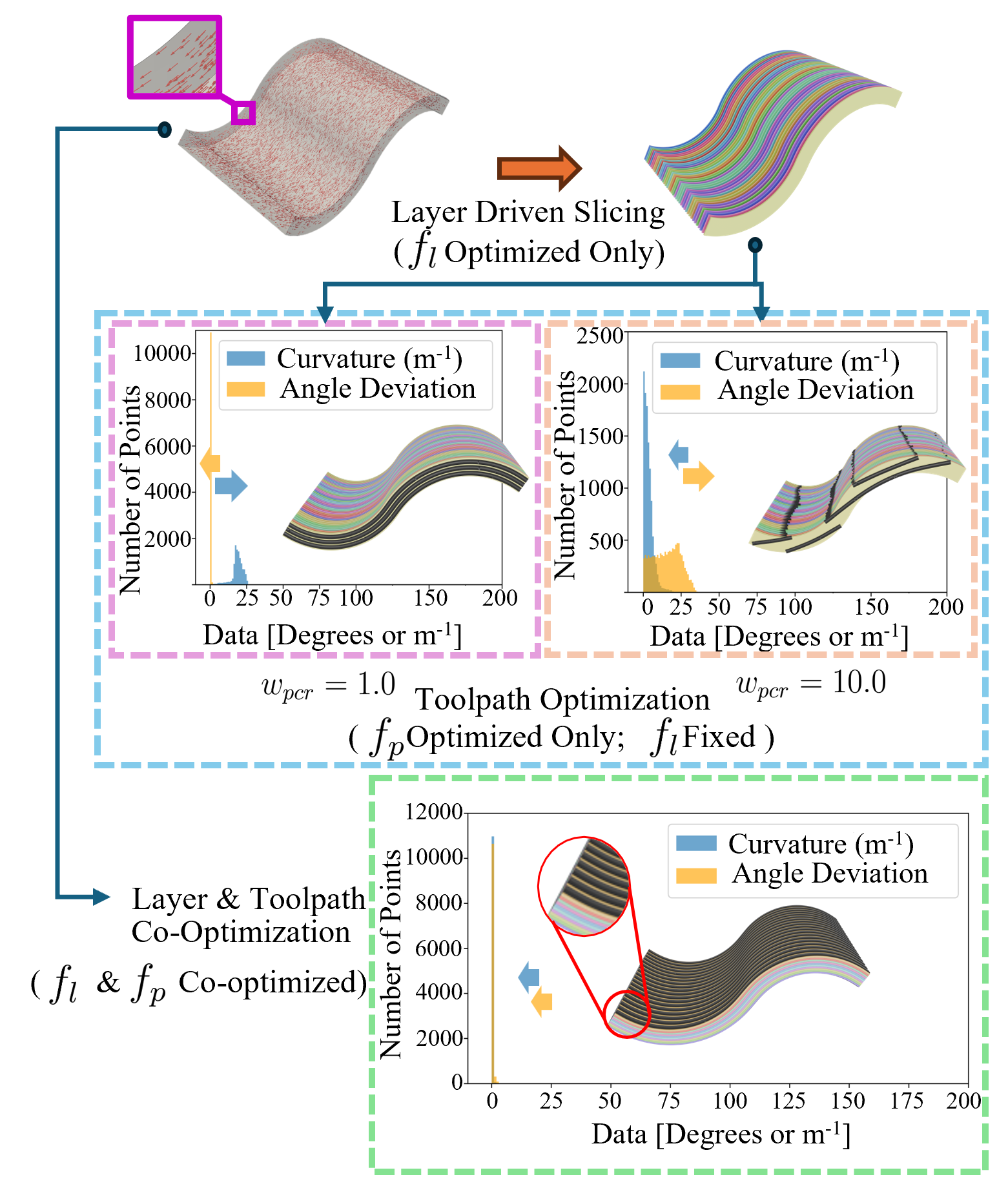}
    \put(-285,330){\footnotesize \color{black}(a)}
    \put(-308,189){\footnotesize \color{black}(c1)}
    \put(-155,189){\footnotesize \color{black}(c2)}
    \put(-80,330){\footnotesize \color{black}(b)}
    \put(-220,8){\footnotesize \color{black}(d)}
    \caption{Results for the S-Curve model. The colored layers indicate different layers, whereas the black curves indicate the fiber-toolpath on the layers. (a) Target model with the prescribed conformal direction field. (b–c2) Sequential optimization results. (b) shows the outcome of the first stage, where only the layer field is optimized. This is followed by toolpath-only optimization, with (c1) and (c2) illustrating the results for two different values of the curvature-weight parameter $w_{pcr}$. (d) presents the result obtained from our joint layer–toolpath co-optimization approach, which achieves superior alignment and curvature performance, as also reflected in the accompanying plot. }
    \label{fig:result_tp_cooptimisation}
\end{figure}

While previous cases focused solely on layer-based optimization, i.e., the shape of layers, certain applications require careful design of the toolpath as well. This is particularly important for materials with highly anisotropic properties, such as continuous carbon fibers, where performance depends not only on toolpath direction but also on geometric properties such as curvature. Existing methods typically generate toolpaths as a separate step and lack of simultaneous and explicit control over geometry. Here, we demonstrate that our method generates toolpaths in a single-step optimization while simultaneously enabling direct geometry control over both the paths and the layers. We illustrate this using a T-Bracket model (Fig.~\ref{fig:result_t_complete}(a)) and an S-Curve mode (Fig.~\ref{fig:result_tp_cooptimisation}(a)).

For the T-Bracket model, which spans 130 mm along the x-axis, the frequency parameters are set as $\omega_0[f_l] = \omega[f_l] = \omega_0[f_p] = 7.0$ and $\omega[f_p] = 10.0$ ($\approx 1.5\omega_0$, will be explained in Sec.~\ref{subsec:singularity}). The optimization employs the total loss $\mathcal{L}_{TPD}$, which includes the direction-alignment and regularization terms for the toolpath.

Fig.~\ref{fig:result_t_complete}(a) shows the T-Bracket model under the applied loading, with the resulting maximal stress directions as shown in Fig.~\ref{fig:result_t_complete}(b). Using our single-step optimization, we generate both the layers and continuous-carbon-fiber (CCF) toolpaths, along with the optimized setup (Fig.~\ref{fig:result_t_complete}(c)). The part was printed with PLA as the matrix material and CCF deposited over 26 layers, with PVA material employed as support, filling the space between the part and platform by the method of Zhang et al.~\cite{zhang_support_2023}. Fig.~\ref{fig:result_t_complete}(d,e) show a simulated in-progress state and the real fabrication process respectively, while Fig.~\ref{fig:result_t_complete}(f) presents the complete simulated and printed parts.

\begin{figure}[!t]
    \centering
    \includegraphics[width=0.95\linewidth]{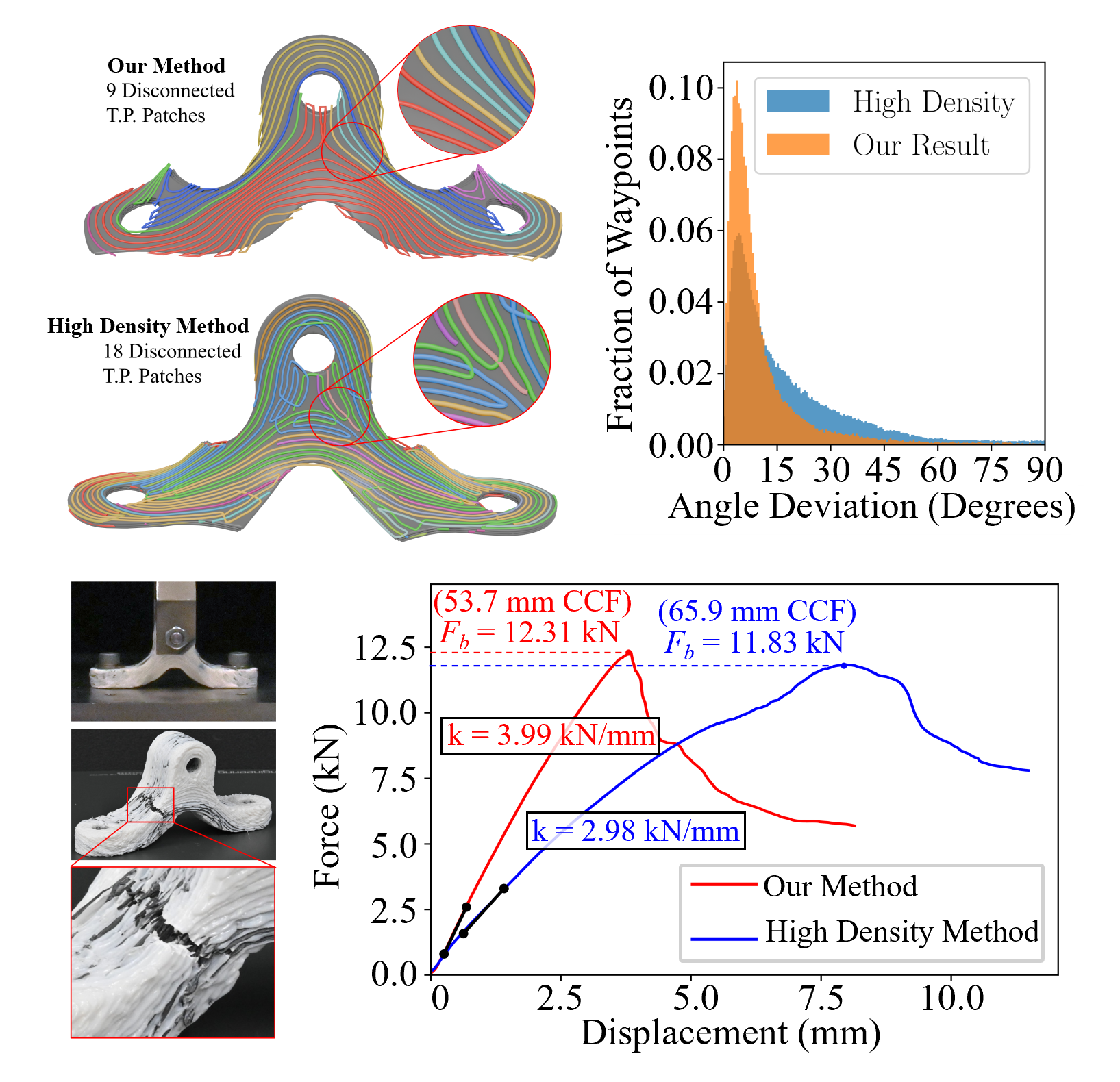}
    \put(-367,275){\footnotesize \color{black}(a1)}
    \put(-367,180){\footnotesize \color{black}(a2)}
    \put(-87,180){\footnotesize \color{black}(b)}
    \put(-327,0){\footnotesize \color{black}(c)}
    \put(-142,0){\footnotesize \color{black}(d)}
    \caption{This figure provides a comparison between our method and the High-Density (HD) Toolpath generation approach \cite{zhang_toolpath_2025} on the same part. (a1) and (a2) illustrate the fiber toolpaths on comparable layers, generated using our method and the HD method, respectively, where colors denote disconnected patches of connected toolpaths. The HD method results in nearly twice as many toolpath jumps as ours and produces numerous path ends and turns within layers, reducing continuity. (b) reports the direction-alignment statistics as angle deviation between the toolpath and the expected direction at different regions, showing that our toolpaths exhibit significantly better alignment with the maximal principal stress directions. This improved alignment directly leads to enhanced mechanical (c) performance: as shown in (d), our parts achieve markedly higher stiffness and increased breaking force, despite requiring less fiber. Finally, (c) presents the tensile test for validation and the failed specimen, from which we can clearly observe fiber fractures. 
    }\label{fig:result_T_Tensile}
\end{figure}

Our method produces continuous CCF toolpaths and curved layers. We also simultaneously controlled the printing setup.  We can control the spacing between paths, enabling the inclusion of more fibers than the methods based on tracing stress lines (e.g., \cite{fang_exceptional_2024}), while also controlling the geodesic curvature. Fig.~\ref{fig:result_T_Tensile}(a1,a2) compares a representative layer of our method with the high-density toolpath approach presented in \cite{zhang_toolpath_2025}, showing that our toolpaths have no discontinuities inside the boundary and fewer jumps, reducing fiber cuts and simplifying printing while also adhering more closely to the intended direction field (Fig.~\ref{fig:result_T_Tensile}(b)).  These improvements generate superior mechanical performance. Despite using $18.5\%$ less (length:~$53.7$m) carbon fiber versus the high-density method (length:~$65.9$m)~(ref.~\cite{zhang_toolpath_2025}), our parts achieve $33.9\%$ higher stiffness ($k$) and comparable breaking force ($F_b$) as shown by the force-displacement curve in Fig.~\ref{fig:result_T_Tensile}(d), with the tensile test setup and failure region being given in Fig.~\ref{fig:result_T_Tensile}(c).

In another experiment, we demonstrate that directly co-optimizing the geometry of toolpaths and layers within a single unified formulation yields superior results compared to the conventional (sequential) two-stage optimization strategies widely adopted in prior work \cite{zhang_toolpath_2025, fang_exceptional_2024}.

As illustrated in Fig.~\ref{fig:result_tp_cooptimisation}(a), we consider the S-Curve Model with a prescribed contour-parallel direction field. The objective is to generate toolpaths that align ($\mathcal{L}_{df1}$, $\mathcal{L}_{df2}$, $\mathcal{L}_{df3}$) with this direction field while simultaneously  satisfying geometric regularization constraints, including minimum (zero) geodesic curvature ($\mathcal{L}_{pcr}$) and uniform toolpath spacing ($\mathcal{L}_{pds}$), and the layer-regularization conditions.

Following the traditional sequential approach, we first optimize the layer field independently by minimizing the layer-based functional and regularization losses, yielding the layers shown in Fig.~\ref{fig:result_tp_cooptimisation}(b). These layers are then used as input to a subsequent toolpath optimization stage based on toolpath-level loss terms. The resulting toolpaths (Fig.~\ref{fig:result_tp_cooptimisation}(c1)) exhibit finite geodesic curvature. Increasing the weight for the curvature-regularization loss ($\mathcal{L}_{pcr}$) successfully reduces curvature (Fig.~\ref{fig:result_tp_cooptimisation}(c2)) but does so at the cost of degraded direction alignment, highlighting a trade-off inherent in the decoupled optimization strategy.

In contrast, our proposed pipeline can perfrom layer–toolpath co-optimization which enables mutual information exchange between the layer and toolpath representations during training. This coupling is particularly crucial for quantities such as geodesic curvature, which depend on both the intrinsic curve geometry and the extrinsic shape of the surface on which the curve lies. As a result, our method simultaneously achieves low geodesic curvature and strong directional alignment, as shown in Fig.~\ref{fig:result_tp_cooptimisation}(d). Such balanced performance cannot be attained by the sequential optimization approach, where the layer-generation stage lacks access to toolpath-level geometric information, thereby constraining convergence to suboptimal layer configurations. This capability is particularly valuable for applications requiring wide fiber stripes, such as Automated-Tape Laying (ATL) \cite{clancy2019study}, where minimizing turns (tow-steering) during printing is critical. \begin{wrapfigure}{r}{0.55\textwidth}
\centering
\includegraphics[width=0.55\textwidth]{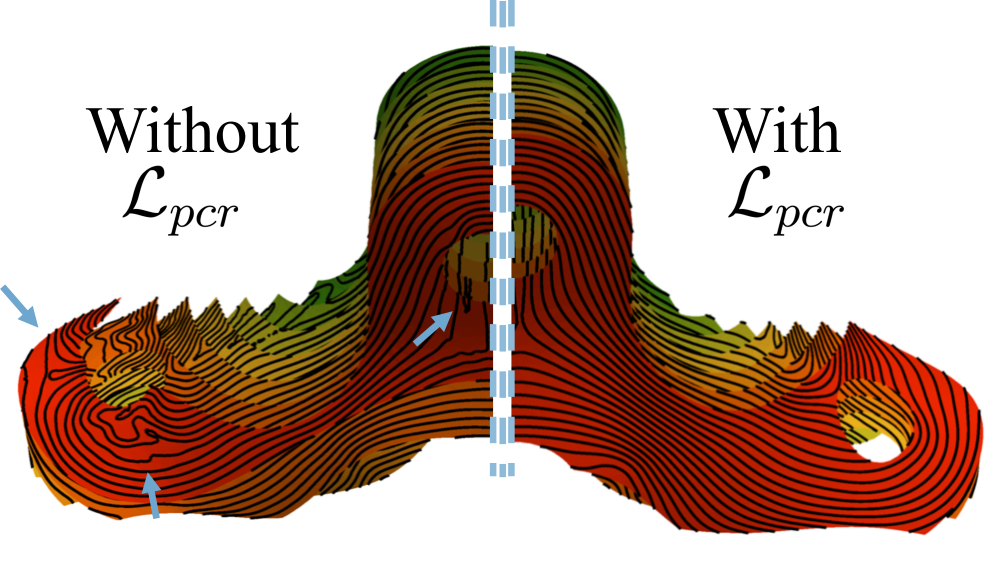}
\end{wrapfigure}This example also demonstrates the effectiveness of our geodesic-curvature control loss. However for the sake of completeness, the adjacent figure also shows the results of toolpaths for the T-Bracket obtained without (left) and with (right) the toolpath-curvature loss ($\mathcal{L}_{pcr}$) after the end of full 1100 optimization epochs.

These two examples highlight the utility and advantages of our pipeline. We further demonstrate similar toolpath-level computations on additional models, as presented in Sec.~\ref{subsec:singularity}.


\subsection{Milling and Collision}

\begin{figure}[!t]
    \centering
    \includegraphics[width=0.99\linewidth]{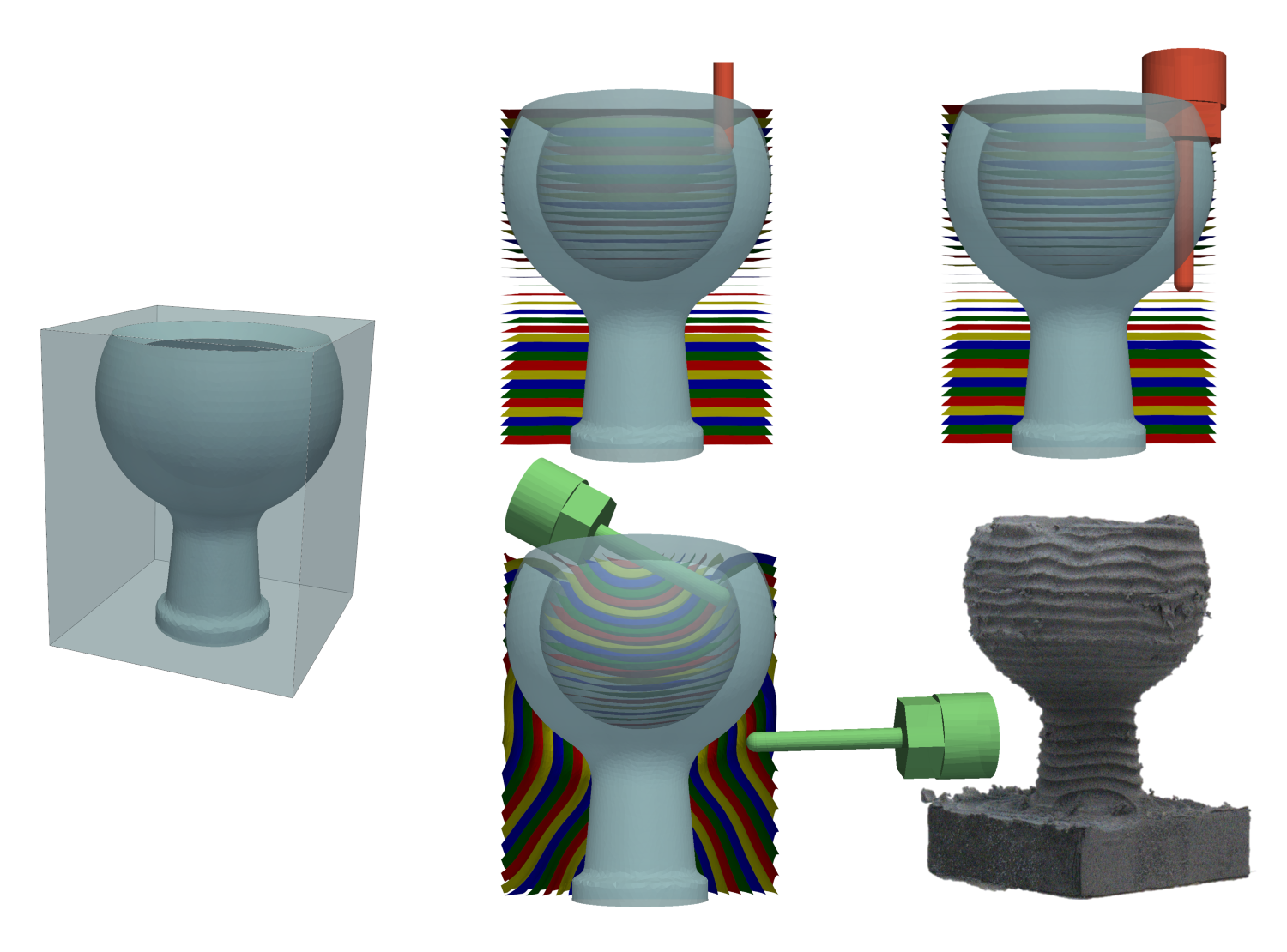}
    \put(-375,60){\footnotesize \color{black}(a)}
    \put(-263,145){\footnotesize \color{black}(b1)}
    \put(-127,145){\footnotesize \color{black}(b2)}
    \put(-263,0){\footnotesize \color{black}(c)}
    \put(-127,0){\footnotesize \color{black}(d)}
    \caption{Rough-milling planning for the Cup model. (a) Target geometry with stock material. (b1) Planar height-field strategy. (b2) Tool–part collisions arising from the planar layers. (c) Result obtained using our collision-loss–based pipeline, eliminating tool–part intersections. (d) Fabricated part after rough milling} 
    \label{fig:result_milling}
\end{figure}

We first demonstrated the pipeline in additive manufacturing. The framework is, however, general and extends to processes involving sequential material addition or removal. To illustrate this, we apply it to volume milling, ensuring collision-free accessibility on the Cup model (Fig.~\ref{fig:result_milling}(a)) from Dutta et al. \cite{dutta_vector_2023}, enabling direct comparison with their method. The optimization employs the total loss $\mathcal{L}_{MLC}$ (Eq.~\eqref{eqn:total_loss_mill}).

Figure~\ref{fig:result_milling}(a) shows the Cup model to be milled from a bounding block. Starting from planar (height-field) layers (Fig.~\ref{fig:result_milling}(b1)), the pipeline optimizes them into accessible layers (Fig.~\ref{fig:result_milling}(c)). As shown in Fig.~\ref{fig:result_milling}(b1,b2), the initial configuration leaves several regions unreachable as the tool intersects with the target part when oriented along the layer normals. The anchor-based method of Dutta et al. \cite{dutta_vector_2023} requires manual anchor placement to address these regions and does not guarantee success. By contrast, our method achieves accessibility automatically by encoding collisions as a differentiable loss, which is effectively equivalent to determining anchor placements without user guidance. Fig.~\ref{fig:result_milling}(c) shows the optimized layers with tools placed at different regions showing accessibility. The resulting rough-milled part is shown in Fig.~\ref{fig:result_milling}(d), fabricated using a contour-parallel toolpath on the optimized layers following \cite{dutta_vector_2023}.


\subsection{Singularity, Frequency and Topology}
\label{subsec:singularity}

In the preceding subsections, we presented various results obtained using our implicit neural-field based pipeline. However, several of the hyperparameter choices adopted during the optimization process were not explicitly justified. In this subsection, we outline the rationale behind these choices and provide a broader discussion on how hyperparameters and loss-function definitions can be selected to achieve specific geometric or manufacturing outcomes.

\begin{figure}[!t]
    \centering
    \includegraphics[width=0.95\linewidth]{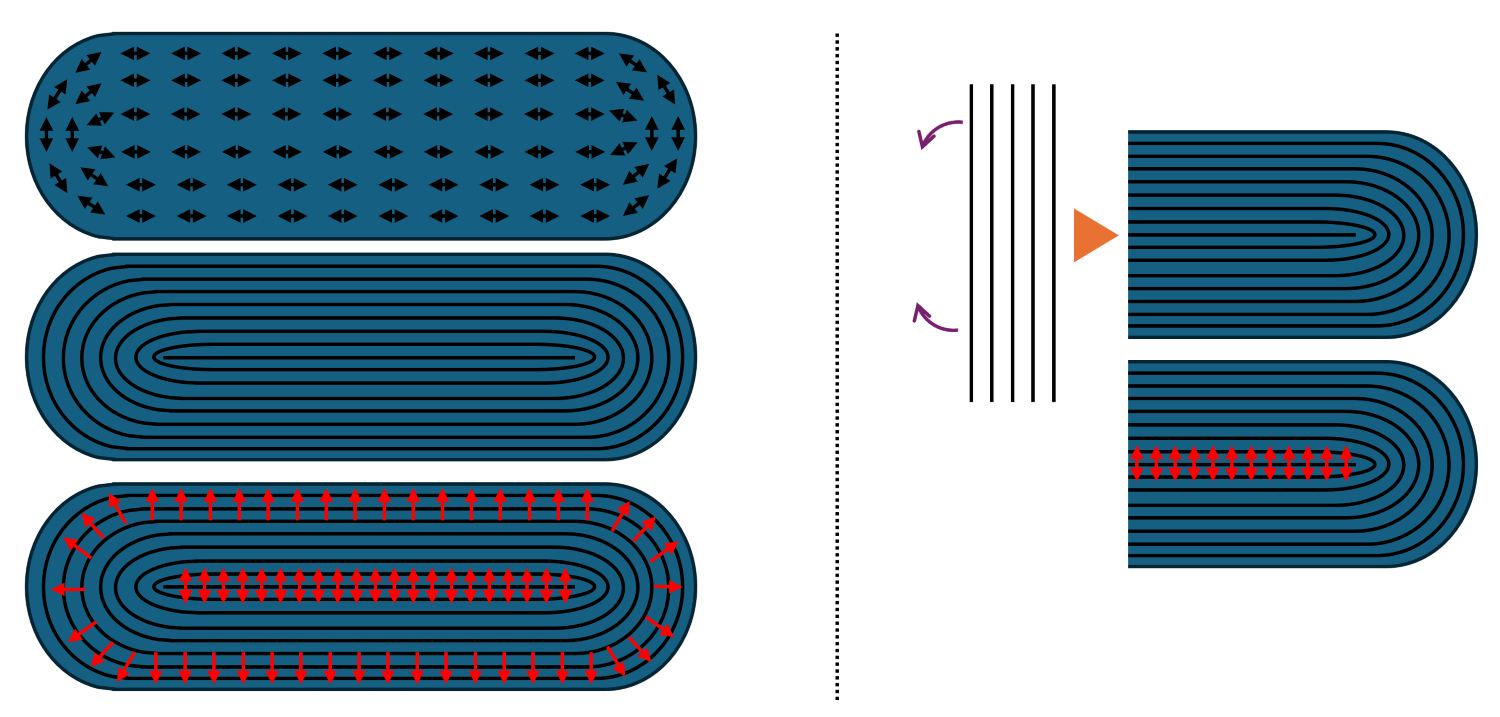}
    \put(-200,115){\footnotesize \color{black}(a1)}
    \put(-200,60){\footnotesize \color{black}(a2)}
    \put(-200,0){\footnotesize \color{black}(a3)}
    \put(-70,0){\footnotesize \color{black}(b)}
    
    \caption{This figure illustrates the necessity of singularities in the gradient of the toolpath field ($\nabla_l f_p$). (a1) shows the prescribed directional field to be followed by the toolpaths, and (a2) presents the resulting toolpaths (black curves) generated by aligning with this field. The corresponding gradient field ($\nabla_l f_p$) is visualized in (a3), where the central region exhibits an ill-defined gradient orientation, indicating the presence of an inherent singularity. In this configuration, the toolpaths form closed contours. In contrast, (b) shows a modified part consisting of only one half of the geometry from (a). Although the directional alignment still requires a singularity, the resulting toolpaths are open rather than closed. This demonstrates that the topology of the generated toolpaths inherently depends on several conditions and cannot be universally predefined}
    \label{fig:freqIllustration}
\end{figure}

Consider the direction field on a layer, illustrated in Fig.~\ref{fig:freqIllustration}(a1). A corresponding toolpath that follows this field is shown in Fig.~\ref{fig:freqIllustration}(a2). If we examine the gradient directions of these toolpaths (direction of $\nabla_lf_p$, defined as normals to the curves, oriented toward higher field values on the surface), an ambiguity emerges in the central region, as seen in Fig.~\ref{fig:freqIllustration}(a3). This ambiguity in the gradient field is referred to as a singularity. For any continuous scalar field $f$, such singularities arise in the vicinity of local scalar extrema, where the gradient magnitude vanishes and the direction becomes undefined.

In Fig.~\ref{fig:freqIllustration}(a2), we further observe that toolpath curves aligned with the direction field are closed. This underscores the influence of the topology of the field’s level sets on the resulting toolpaths. Our pipeline must therefore be capable of accommodating topological transitions, particularly when initialized from arbitrary seed points. At the same time, as shown in Fig.~\ref{fig:freqIllustration}(b), singularities can also arise without the presence of closed curves, further complicating the behavior of the field. These observations motivate the subsequent discussion in this section. Additionally, we examine how the frequency-scaling parameters of sinusoidal activations in our neural representation \cite{sitzmann_implicit_2020} affect the emergence of such features and other aspects of our result.


\subsubsection{Network Frequency}
\label{subsub:network_frequency}
 Following is the mathematical representation of the SIREN architecture \cite{sitzmann_implicit_2020} employed in our implementation:
\begin{equation}
    f(\mathbf{x}) = \mathbf{W_n}(\phi_{n-1}\circ\phi_{n-2}...\circ\phi_0)(\mathbf{x}) + \mathbf{b_n}, \, \phi_i(\mathbf{x}) = sin(\omega_i\cdot\mathbf{W_i\mathbf{x} +\mathbf{b_i})}
\end{equation}
\begin{wrapfigure}{r}{0.45\textwidth}
    \centering
    \includegraphics[width=0.40\textwidth]{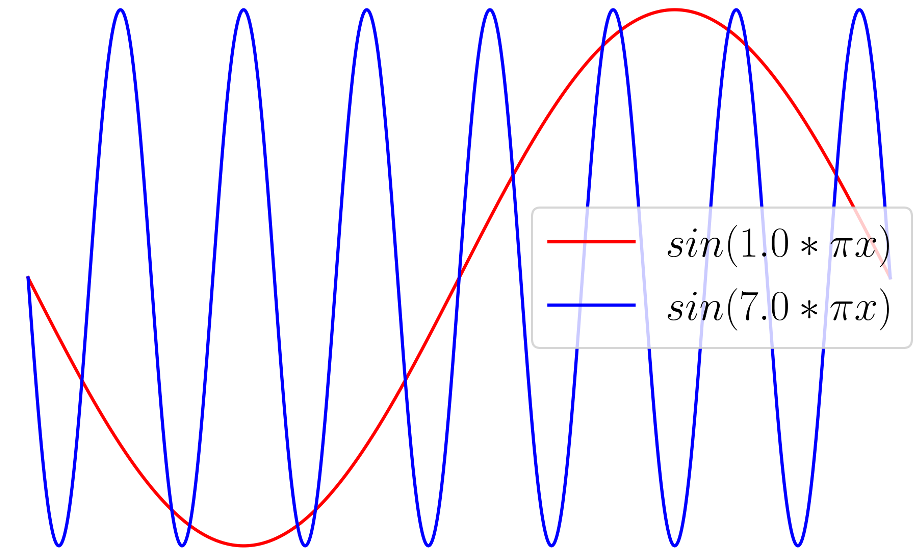}
\end{wrapfigure}

The parameter $\omega_i$ serves as a hyperparameter that determines the number of sinusoidal periods within a fixed domain (see the adjacent inset figure). Hence, when other network parameters are fixed, $\omega_i$ directly controls the frequency of variation of the field and its derivatives, and therefore the number of local extrema (or singularities) present in the scalar field.

We adopt the network initialization scheme proposed by Sitzmann et al.~\cite{sitzmann_implicit_2020}, where the input-layer frequency $\omega_0$ does not influence the initialization of the trainable weights, while the frequency-scaling $\omega_i$ of all subsequent layers, hereafter referred to simply as $\omega$, does (scaled by that factor and hence cancels out its direct effect on the initialized state). Accordingly, at the initialized  state, we can assume that $\omega_0$ takes the major role of the frequency-scaling factor \cite{yeom_fast_2024}. Despite this, there is no explicit decoupling in the effect and we keep the values of $\omega$, similar to $\omega_0$, as also used by Sitzmann et al.\cite{sitzmann_implicit_2020}. As described in Sec.~\ref{subsub:optimisationScheme}, all models are scaled to lie within the $[-1,1]^3$ bounding cube. Under this normalization, the model’s geometric scale interacts multiplicatively with the frequency-scaling parameter. To ensure that the physical size of a model does not affect the characteristics of the layer field or the resulting toolpaths, we define $\omega_i$ as a function of the model scaling factor such that this dependency is cancelled. Specifically, $\omega_i = c(\text{geometric scaling factor})$ where $c$ is a constant. Thus, when the part is scaled by this factor, the product 
\begin{center}
$\omega_i\mathbf{x}=[c(\text{geometric scaling factor})\frac{\mathbf{x}'}{\text{geometric scaling factor}}]$, 
\end{center}
with $\mathbf{x}'$ denoting the unscaled coordinate, ensures that the frequency behavior of the network remains invariant to model size.

\begin{figure}
    \centering
    \vspace{-75pt}
    \includegraphics[width=0.90\linewidth]{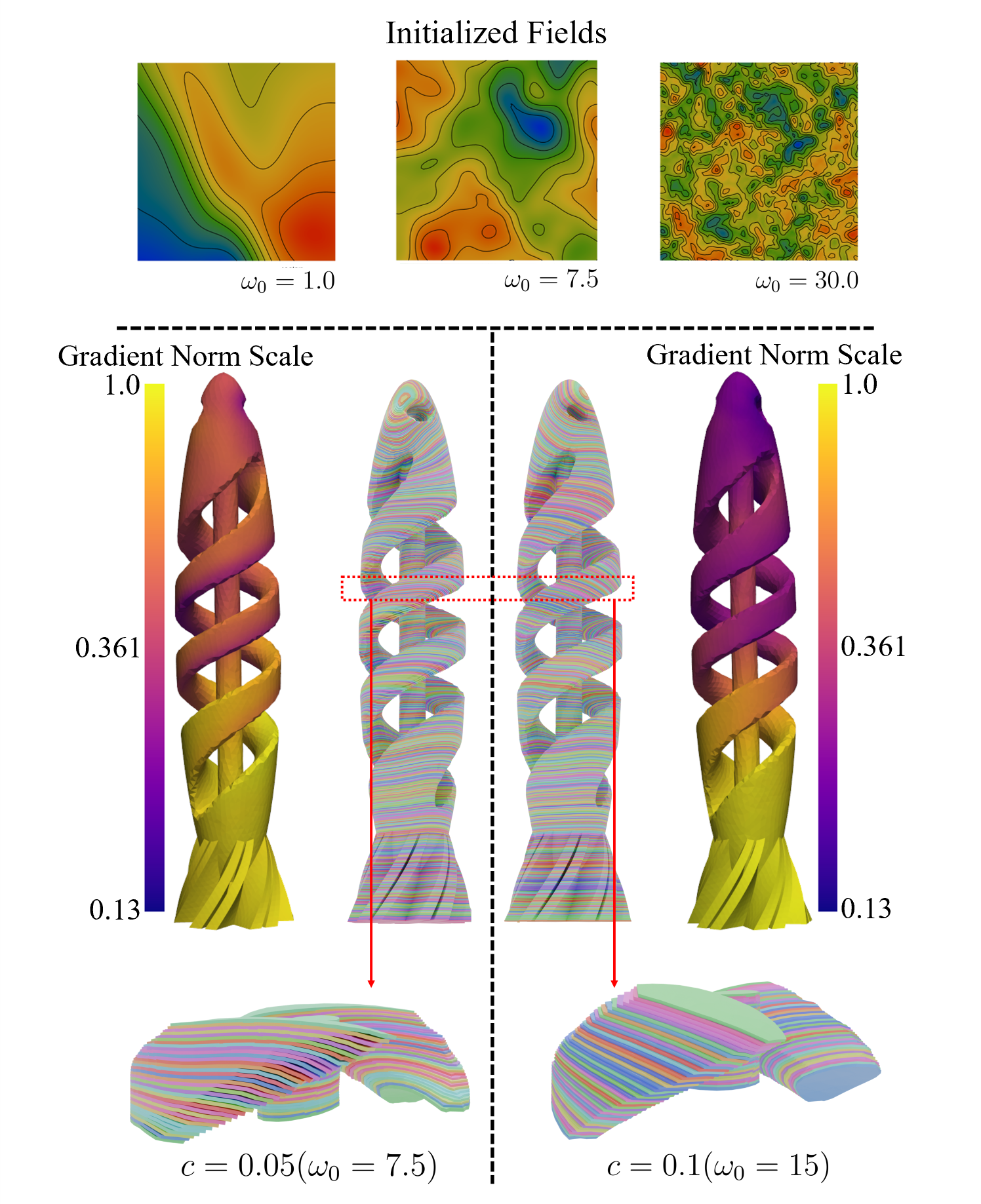}
    \put(-276,318){\footnotesize \color{black}(a1)}
    \put(-175,318){\footnotesize \color{black}(a2)}
    \put(-78,318){\footnotesize \color{black}(a3)}
    \put(-279,90){\footnotesize \color{black}(b1)}
    \put(-213,90){\footnotesize \color{black}(b2)}
    \put(-250,-12){\footnotesize \color{black}(b3)}
    \put(-150,90){\footnotesize \color{black}(c1)}
    \put(-85,90){\footnotesize \color{black}(c2)}
    \put(-89,-12){\footnotesize \color{black}(c3)}
    \caption{Effect of the frequency-scaling factor ($\omega_i$) in SIREN networks.
(a1), (a2) and (a3) show cross-sections of the initialized layer field $f_l$ for $\omega_0=\omega=1.0$, $\omega_0=\omega=7.5$ and $\omega_0=\omega=30$, respectively. \revision{}{The colors show the scalar values and black curves show contour lines}. A lower frequency-scaling value produces a more regularized initial field with fewer singularities, whereas a higher value introduces finer oscillations and greater local variation. (b1)–(c3) illustrate the influence of the empirical constant $c$ in the frequency-scaling formulation $\omega_0 = c(\text{geometric scaling factor})$ for support-free slicing of the Spiral-Fish. For $c=0.05$ [(b1)–(b3)], the field exhibits slower spatial variation, causing the layers to align more horizontally under the influence of the central cylindrical region. This inter-regional coupling leads to non-uniform inter-layer spacing, as highlighted in (b3) and reflected in the gradient magnitude ($\|\nabla f_l\|$) distribution shown in (b1). In contrast, $c=0.1$ [(c1)–(c3)] allows more rapid variation in both the scalar field and its gradient norm $\|\nabla f_l\|$, thereby reducing inter-regional coupling and improving layer uniformity and adaptability.}
    \label{fig:spiral_abalation_frequency}
\end{figure}

The constant $c$ thus acts as an effective frequency-scaling parameter that defines the network’s frequency bias. Fig.~\ref{fig:spiral_abalation_frequency}(a1),(a2) and (a3) illustrate the initialized $f_l$ fields within a $[-1,1]^2$ domain for  $\omega_0 = 1.0$, $\omega_0 = 7.5$ and $\omega_0 = 30.0$. As expected, higher $\omega_0$ values produce fields with more singularities and higher spatial oscillations. This property must therefore be carefully considered when selecting $c$, as it determines the field’s representational expressiveness (ability of capturing a variety of shapes, patterns, or details) and smoothness.

Another perspective on $\omega_i$ concerns the rate of change of both the scalar field and its gradient. As shown in the previous inset figure illustrating the sine-function, a higher frequency results in faster variation of both quantities, effectively reducing the inherent smoothness or regularization in the representation. We use this insight to empirically select an appropriate value of $c$.

Fig.~\ref{fig:spiral_abalation_frequency}(b1)-(c3) demonstrates the influence of different $c$ values on the layer field $f_l$ for the Spiral-Fish model under the support-free printing condition. As anticipated, lower-frequency values inhibit rapid changes in the field, thereby coupling the behavior of distant regions. For instance, with low $c$ values, the central cylindrical layers restrict the twisting of layer field in the two helices, forcing the support-free constraint to be satisfied primarily near the outer boundaries. This leads to non-uniform layer spacing along the helices. In contrast, higher frequencies enable faster spatial variation and thus stronger local decoupling between the central and helical regions, resulting in more uniform layers. This observation is similar to our earlier ablation results on the same model (Fig.~\ref{fig:result_spiral_norm}), where a stronger constraint was shown to create inter-regional coupling and generate more non-uniform layer spacing.

The effect is also visible in the magnitude of the field gradient $\|\nabla f_l\|$ [(b1),(c2)], where higher frequencies produce distinct gradient magnitudes in the central region, corresponding to different local rates of change for $f_l$. From all of our experiments, and as depicted in Fig.~\ref{fig:spiral_abalation_frequency}[(c1)-(c3)], we found that $c = 0.1$ provides a balanced trade-off between field smoothness and local adaptability. While larger values of $c$ could further enhance the ability of the field to vary spatially and decouple local regions, they also introduce a stronger bias toward high-frequency features, thereby increasing susceptibility to noise (Figs.~\ref{fig:spiral_abalation_frequency}(a3) and~\ref{fig:abalationSurfCurv}).

Therefore, throughout this work, we use $c = 0.1$ for $f_l$ as the default choice. Nonetheless, determining an optimal frequency that balances representational expressiveness and robustness remains an interesting topic for further study.


\subsubsection{Singularities and Curvature control}

In the beginning of this subsection, we presented an illustration (Fig.~\ref{fig:freqIllustration}) showing toolpaths exhibiting different topologies and singularities in the gradient of the toolpath field, $f_p$. A similar type of singularity (local extremum) can also occur in the layer field, $f_l$. However, unlike in the case of toolpaths, singularities in $f_l$ are undesirable because they lead to non-manufacturable layers \cite{dutta_vector_2023}. Therefore, for the layer scalar field, $f_l$, we assume a monotonic growth from a base surface outward. This implies that $f_l$ should have a topology comparable to that of conventional planar slicing, which we use as the initial configuration for optimizing the field. As will be shown later in this subsection, this choice also yields additional benefits.

In contrast, as illustrated in Fig.~\ref{fig:freqIllustration}, such an assumption cannot be made for the toolpath field, $f_p$, where the presence of singularities is often necessary to satisfy directional or geometric constraints. The ease of formation and stability of these singularities depend on both the network properties and the defined loss functions. In Sec.~\ref{subsub:network_frequency}, we described how the frequency bias of the SIREN network can be tuned to facilitate the emergence of such singularities. However, near these regions, the gradient (first derivative) becomes ill-defined, and measures involving curvature (second derivatives) tend to produce very large values. Consequently, curvature-based losses should only be activated in regions where the gradient is well defined.

To achieve this, we introduce a masking/filtering scheme that restricts the computation of curvature loss to regions where the norm of the projected gradient of the toolpath field exceeds a small threshold, $\delta$. The implementation is given by:
\begin{equation}
    \mathcal{L}= \sum_{\mathbf{x}\in\{\Omega\}} (10\,ReLU(|\kappa_{geo}| - \kappa_T))^2\cdot\eta(\mathbf{x})\cdot\frac{1}{|\{\eta\circ\Omega\}|}
\end{equation}
where we define $\eta$ as:
\begin{equation}
    \eta({\mathbf{x}}) = 
    \begin{cases}
        1, & \text{if $\| \nabla_lf_p\| \geq \delta$}\\
        0, & \text{if $\| \nabla_lf_p\| < \delta$}\\
    \end{cases}
\end{equation}

The implementation of $\eta$ can be approximated smoothly using a sigmoid function. The $\eta$ reduces the domain $\Omega$ to a subdomain $\Omega'$ as used in Eq.~\eqref{eqn:loss_curvature_TP}. When combined with the path-distance loss $\mathcal{L}_{pds}$ (Eq.~\eqref{eqn:toolpath_density_loss}) and appropriately chosen weights and thresholds, this scheme enables the generation of toolpaths with the desired geometric properties, without explicitly defining the locations of the singularities.

\begin{figure}[!t]
    \centering
    \includegraphics[width=0.95\linewidth]{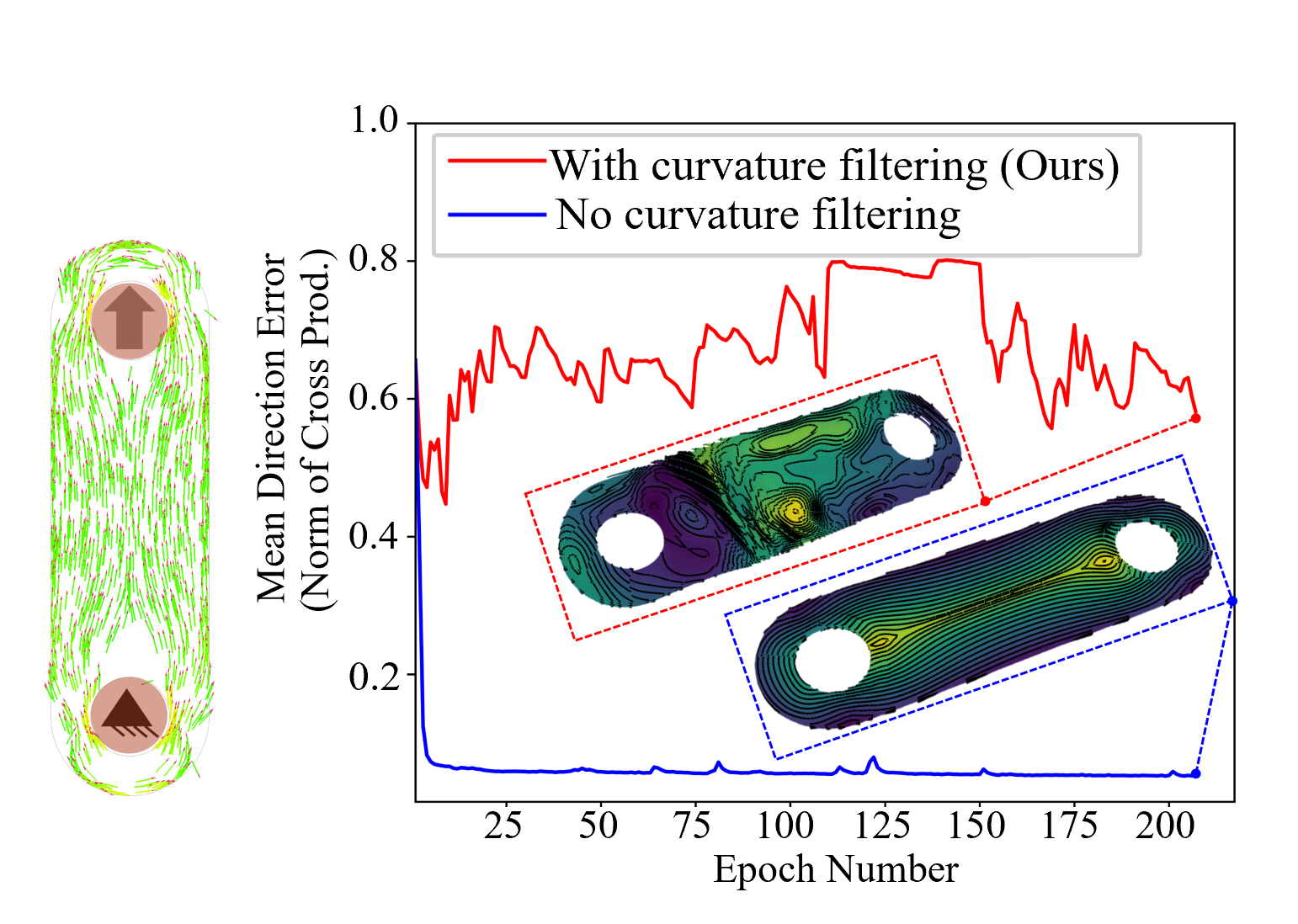}
    \put(-345,-5){\footnotesize \color{black}(a)}
    \put(-155,-5){\footnotesize \color{black}(b)}
    \caption{Figure illustrating the need for our toolpath curvature filtering method. (a) shows the loading and principal stress field on the Flat-Bracket model. (b) compares results with and without filtering in the path-curvature loss $\mathcal{L}_{pcr}$. Notice that applying the path-curvature loss naively prevents the field from converging to the directionally aligned result.}
    \label{fig:curvatureFiltering}
\end{figure}
We demonstrate the effect of this masking strategy in Fig.~\ref{fig:curvatureFiltering} for a Flat Bracket model subjected to the loading condition shown in Fig.~\ref{fig:curvatureFiltering}(a). This setup is analogous to the case shown earlier in Fig.~\ref{fig:freqIllustration}. Without the filtering scheme, we observed that the optimization process fails to converge because the extremely high curvature values at singularities dominate the total loss, preventing the optimizer from resolving other regions effectively.

\begin{figure}[!t]
    \centering
    \includegraphics[width=1.00\linewidth]{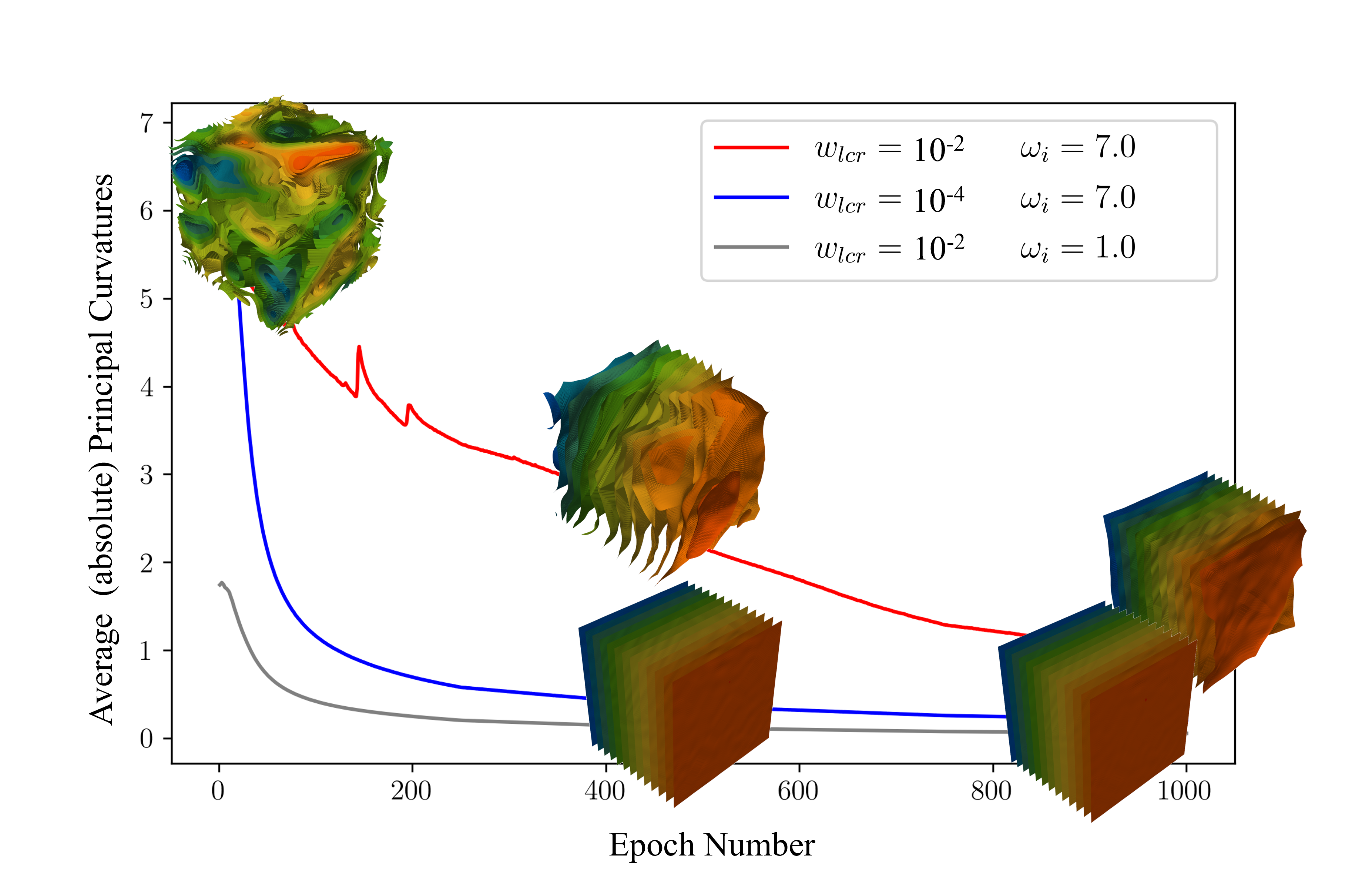}
    \caption{In this figure we illustrate the effect of the weight of the layer curvature loss $\mathcal{L}_{lcr}$ for various frequency scaling values. The target is to reach planar layers from an initial state. Higher frequencies are initialized with many singularities and convergence to planar solution require multiple topology changes. As seen in the figure, higher value of $w_{pcr}$ restricts such convergence.
}
    \label{fig:abalationSurfCurv}
\end{figure}

In general, the requirements on the curvature have been observed to slow down or prevent the convergence. We present a related example for the layer field $f_l$ in Fig.~\ref{fig:abalationSurfCurv}. Notice how a higher weight on the surface curvature loss slows down the convergence. This can be attributed to the strong loss on the curvature (or similarly for a normalized Laplacian) which hampers change of the topology of the field, due to the presence of a norm in the denominator which drives the solution towards a larger norm. This provides additional motivation for initializing the layer field $f_l$ from a planar configuration, which allows the use of higher-frequency representations without the adverse effects of excessive curvature penalization.

For toolpath fields, on the other hand, we employ a gradually increasing weighting strategy for the curvature loss, starting from a small value and progressively raising it during training, to mitigate its negative effects while still achieving smooth, geometrically consistent toolpaths.

\subsubsection{Singularity and Direction Alignment}
In Sec.~\ref{subsec:direction_loss}, we discussed that Eq.~\eqref{eqn:direction_follow_main} represents the true formulation of the directional alignment error. However, we proposed to substitute this formulation with three alternative conditions (Eqs.~\eqref{eqn:direcion_follow_true}–\eqref{eqn:direction_path_only}). Here, we present a comparative study to validate the claims made in Sec.~\ref{subsec:direction_loss} and to justify the adopted formulation. For clarity, we refer to the original formulation (Eq.~\eqref{eqn:direction_follow_main}) as the normalized loss, and to the combined set of alternative formulations as the non-normalized loss.

\begin{figure}[!t]
    \centering
    \includegraphics[width=1.0\linewidth]{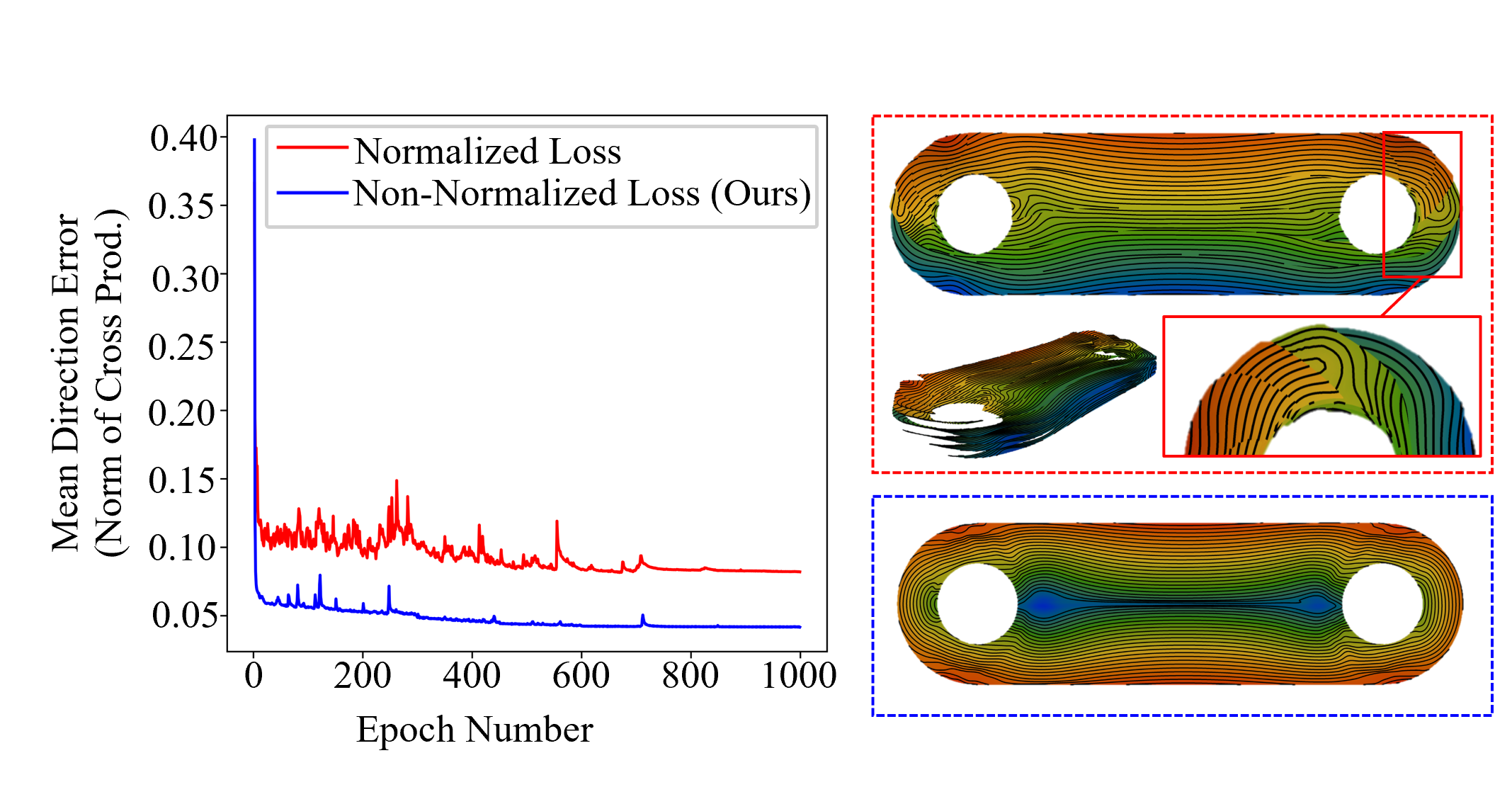}
    \put(-270,-7){\footnotesize \color{black}(a)}
    \put(-167,83){\footnotesize \color{black}(b1)}
    \put(-167,17){\footnotesize \color{black}(b2)}
    \put(-100,-7){\footnotesize \color{black}(b)}
    \caption{This figure illustrates the effect of directly using Eq.\eqref{eqn:direction_follow_main} (normalized version) to optimize the toolpaths versus our proposed version (Eqs.\eqref{eqn:direcion_follow_true}-\eqref{eqn:direction_path_only}, non-normalized version). The presence of singularities renders the normalized version unstable, potentially causing convergence to a suboptimal solution. Subfigure (b1) shows the result obtained using the normalized loss, whereas the non-normalized loss leads to convergence to a better solution, as shown in (b2). The convergence behavior for both cases is also presented in the plot (a).}
    \label{fig:ablation_normalized_direction}
\end{figure}
Figure~\ref{fig:ablation_normalized_direction} shows the results for the Flat-Bracket model used previously in Fig.~\ref{fig:curvatureFiltering}. With the normalized loss (Eq.~\eqref{eqn:direction_follow_main}), the presence of singularities often leads to unstable convergence, resulting in solutions that deviate significantly from the target directions (Fig.~\ref{fig:ablation_normalized_direction}(a)) and, consequently, produce distorted layer structures (Fig.~\ref{fig:ablation_normalized_direction}(b1)). In contrast, our non-normalized formulation allows the formation of stable singularities (Fig.~\ref{fig:ablation_normalized_direction}(b2)) and achieves better convergence toward the target field (Fig.~\ref{fig:ablation_normalized_direction}(a)), consistent with our observations in Fig.~\ref{fig:freqIllustration}

In practice, the input field that guides the toolpaths is not necessarily a well-defined gradient field. This implies that the desired toolpath field should have the ability to change its topology and, where necessary, admit singularities. When the loss is formulated in normalized form, i.e., all vectors are normalized, the optimizer tends to stagnate in a suboptimal local minimum (Fig.~\ref{fig:ablation_normalized_direction}). This can be explained by the mathematical structure of the loss; the normalization introduces vector norms in the denominator. For a gradient field to undergo a topological change and have singularities, there must exist regions where the gradient norm approaches zero. However, such a case is not encouraged by the normalized version, thereby impeding topology changes and sustainment of singularities. Conceptually, this corresponds to a field that can only rotate its local direction but not modulate its magnitude. The non-normalized formulation \revision{}{(Fig.~\ref{fig:ablation_normalized_direction} (b2))}, on the other hand, allows the magnitude of the gradients to vary, providing additional degrees of freedom to reach a lower-energy configuration.

Furthermore, since the curvature loss inherently penalizes high-curvature regions,thus suppressing the formation of singularities, the presence of such singularities can be encouraged by initializing the toolpath field from a state that already exhibits them. So, for the toolpath field $f_p$, we initiate the field from the random state described in Sitzmann et al.~\cite{sitzmann_implicit_2020}. As shown in Fig.~\ref{fig:spiral_abalation_frequency}, a higher-frequency initialization increases the likelihood of singularities in the initial state.  Besides, we also increase the number of neuron-layers for $f_p$ to 15 because of the more complex fields necessary for the toolpath compared to the layers as described in this section. Because we want to co-optimize both $f_l$ and $f_p$ together with $f_l$ starting from a good (planar) field, we set the frequency scaling factors for the toolpath as: 
\begin{center}
    $\omega_0[f_p]=\frac{\omega[f_p]}{1.5} = 0.1 ({\text{geometric scaling factor}})$. 
\end{center}
This helps for faster convergence of $f_p$ (owing to larger gradients, ref.~\cite{sitzmann_implicit_2020}) and the higher frequency also contributes to a higher expressiveness. 

\revision{}{
\subsection{Analysis: Convergence, Weight and Scale}}

\revision{}{In the previous section, we analyzed the network hyperparameters and provided a rationale for the design choices. In this subsection, we present additional analysis to further support the applicability of the proposed approach.}

\begin{figure}
    \centering
    \includegraphics[width=1.0\linewidth]{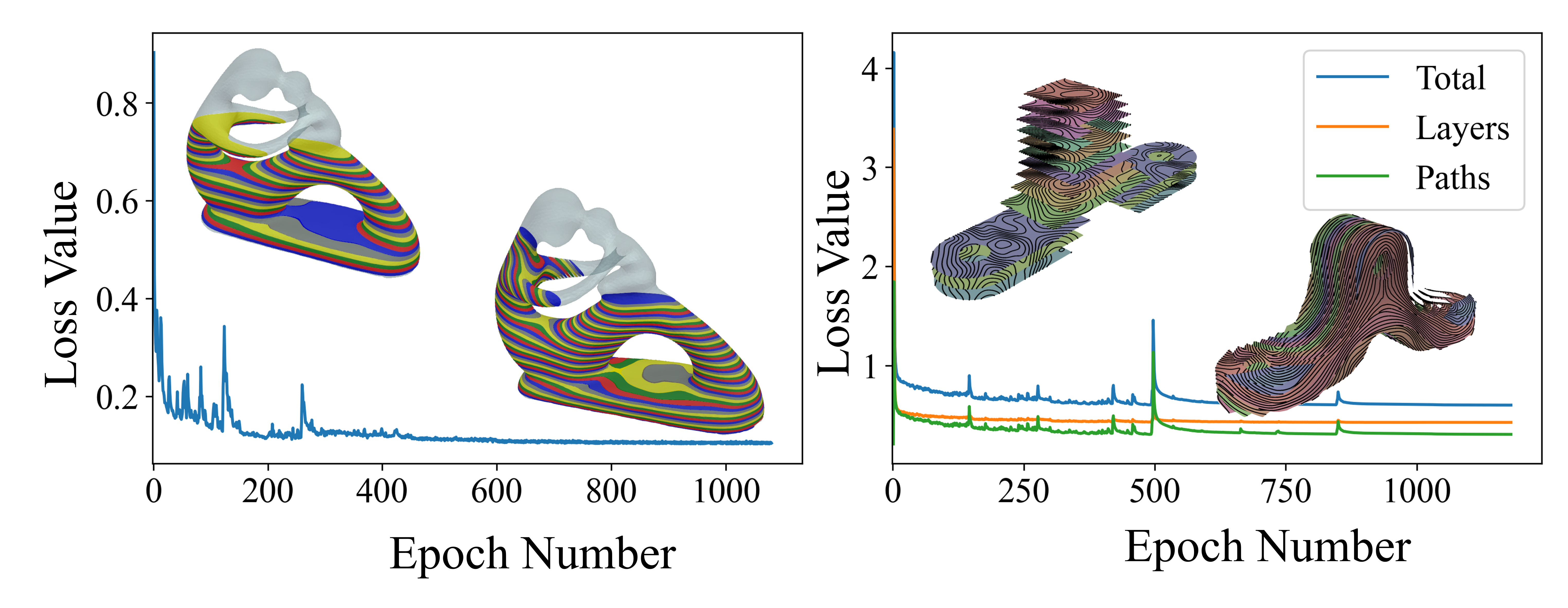}
    \put(-250,-5.0){\small \color{black}(a)}
    \put(-100,-5.0){\small \color{black}(b)}
    \caption{Evolution of loss distributions for the (a) Fertility and (b) T-bracket over training epochs. In addition to the total loss, the aggregated layer-related and toolpath-related loss components are reported in (b). The loss converges to a stable plateau after approximately 1100 epochs, which is adopted as the termination criterion in our optimization. Insets illustrate the initial and final configurations of layers and toolpaths, highlighting the progression toward the optimized state.}
    \label{fig:convergenceTbr}
\end{figure}

\revision{}{The first result concerns the convergence behaviour of the optimization process. The optimization is performed for 1100 epochs; although this choice is not critical, Fig.~\ref{fig:convergenceTbr} show that, across different applications, the loss values consistently converge to a stable plateau within this range. }

\revision{}{A satisfactory solution is defined as one in which all requirements are met within manufacturable tolerances. All requirements, including collision avoidance, are formulated as soft constraints. To enable collision-free configurations, enclosure units are slightly offset; consequently, the collision loss is not expected to reach zero. Therefore, full requirement satisfaction cannot be inferred solely from loss values. Instead, explicit geometric, functional, or collision-based indicators are recommended as auxiliary criteria to terminate the optimization.}

\begin{figure}
    \centering
    \includegraphics[width=0.95\linewidth]{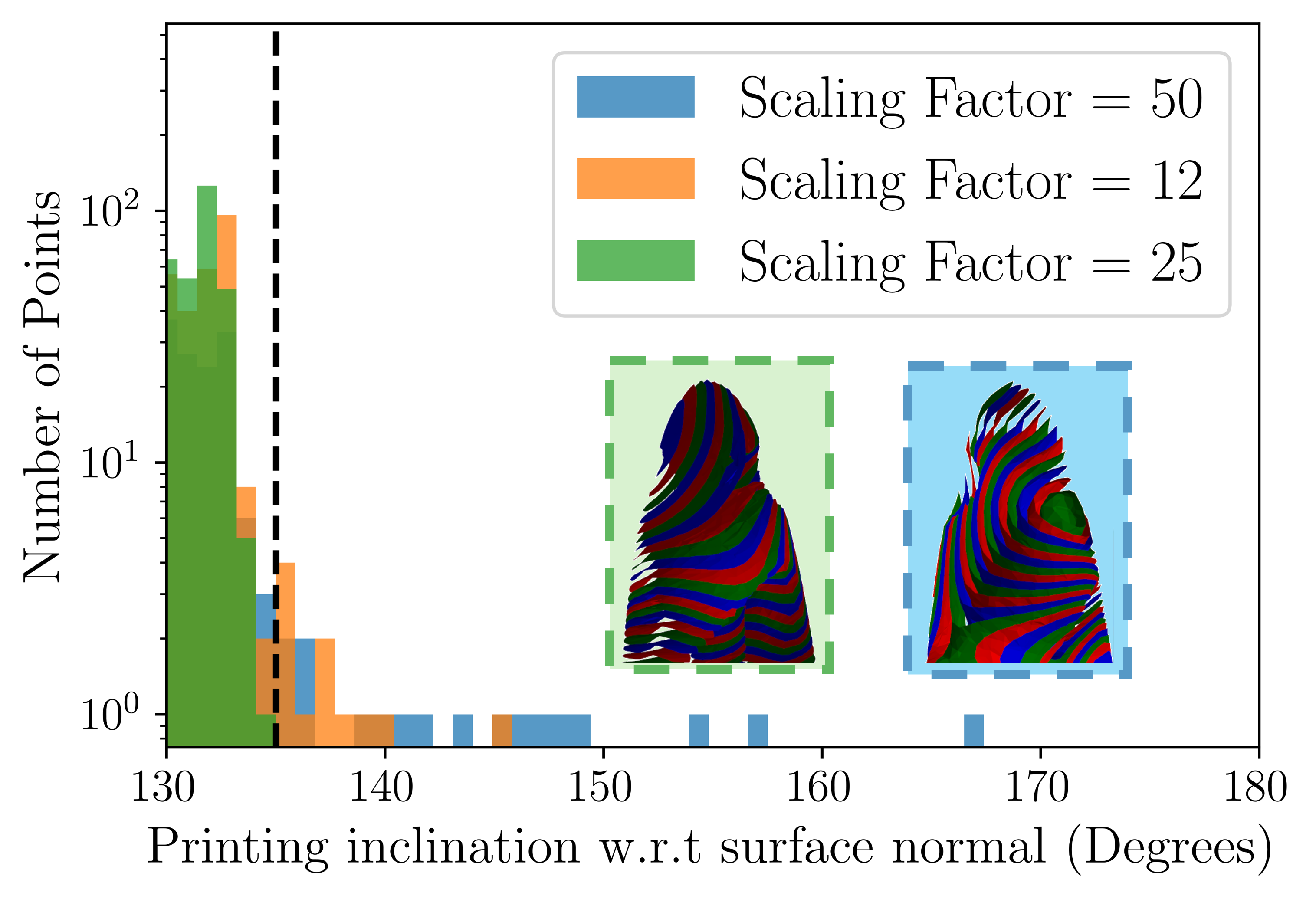}
    \caption{Histogram of the angle between the surface normal and local printing directions for Spiral Fish Model, restricted to regions exceeding $130^\circ$ for visual clarity. The black dashed line at $135^\circ$ denotes the threshold for support-free printing. The distributions are shown for varying scaling factors in $\mathcal{L}_{\mathrm{lsf}}$, with all other terms held constant. Increasing the scaling factor from 12 to 25 improves compliance with the support-free criterion; however, a further increase to 50 induces large gradients, resulting in numerical instability and degraded performance, as also evidenced by cusp formation  (singularities) in the corresponding layers.}
    \label{fig:scalingFactor}
\end{figure}

\revision{}{In Sec.~\ref{sec:losses_objectives}, intrinsic scaling parameters were introduced to control the slope of each loss term, thereby determining its contribution to gradient-based updates. Given the high expressiveness of SIREN networks and their tendency to form singular regions, the network is initialized as planar surfaces to facilitate convergence. The scaling parameters must therefore be chosen to balance stability and enforcement of functional behaviour. Fig.~\ref{fig:scalingFactor} illustrates their effect on support-free loss functions.}

\revision{}{Although intrinsic scaling is mathematically equivalent to weighting loss terms, we distinguish them conceptually. Intrinsic scaling defines the baseline magnitude and gradient behaviour of each loss term, whereas weights adjust their relative importance in a task-dependent manner. For functional losses, scaling factors are incorporated directly into the loss definition, as they remain consistent across different examples and simplify the overall formulation. In contrast, for losses such as collision, both scaling and additional weighting are employed, with scaling defining the baseline and weights providing further adjustment. This distinction improves interpretability without altering the optimization. The influence of weighting on convergence is also illustrated in Fig.~\ref{fig:abalationSurfCurv}. Similar analyses can therefore be used to guide the selection of appropriate scaling factors and weights.}

\revision{}{To evaluate scalability, additional experiments are conducted using the Fertility model sampled at three resolutions—4137, 9897, and 33,696 points for the regularization samples ($\Omega$), also used in collision computations. The corresponding runtimes are 60.95 min, 161.80 min, and 544.99 min, respectively. These results complement Table~\ref{tab:result_summary} and highlight the computational scaling behaviour. However, as the current implementation does not fully utilize GPU acceleration, the reported runtimes likely underestimate the framework’s scalability potential.}

\vspace{15pt}

\section{Conclusion and Discussion}
In this work, we presented a universal differentiable framework for multi-axis process planning, capable of addressing a range of objectives across both additive and subtractive manufacturing contexts. By representing layers, toolpaths, and auxiliary quantities as implicit neural fields, the proposed pipeline enables direct and simultaneous optimization of multiple geometric and manufacturing constraints at both layer and toolpath levels. Through a set of experiments, we demonstrated that the method effectively handles diverse fabrication scenarios while maintaining geometric fidelity and manufacturability.

Beyond the practical demonstrations, we provided an in-depth analysis of the sinusoidally activated neural network (SIREN), examining its inherent properties such as frequency scaling and field regularization. We showed how these characteristics can be systematically controlled to influence field smoothness, local coupling, and representation bias, thus linking network-level parameters with geometric behavior in manufacturing applications.

Despite these advances, several opportunities remain for future exploration.
First, while we considered support structures as a post-processing step, it could be directly integrated into our differentiable framework, given that existing scalar-field based support generation techniques~\cite{zhang_support_2023} are compatible with our implicit representation. This would enable joint optimization of layers, toolpaths, and supports within a single formulation.

Second, although we formulated multiple functional objectives such as direction alignment, collision avoidance, and support-free manufacturability, our framework can be naturally extended to include additional geometric or process-level losses, such as total layer area or toolpath length. Furthermore, our current implementation assumes axially symmetric tool geometries; incorporating the full inverse kinematics of the hardware would allow handling of general tool shapes and machine configurations, further expanding the scope of collision-aware planning.

\begin{figure}
    \centering
    \includegraphics[width=1.0\linewidth]{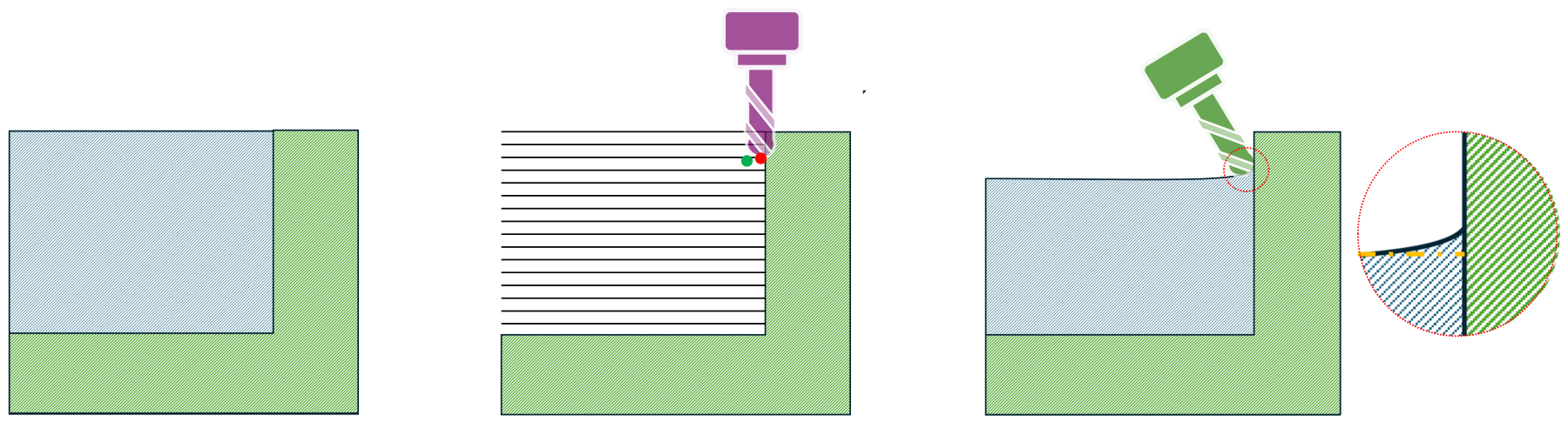}
    \put(-360,-10){\small \color{black}(a)}
    \put(-225,-10){\small \color{black}(b)}
    \put(-110,-10){\small \color{black}(c)}
    \caption[Over-conservative collision removal example]{Over-conservative collision removal example.(a) Target part (green) obtained by milling the initial stock (blue).(b) Planar layering configuration. A collision is detected when the tool center is located at the red point. However, the same material region can be removed from an alternative collision-free tool position (green point) due to the finite cutter radius. (c) Enforcing collision avoidance at all sampled tool positions, including unnecessary ones such as the red point, results in artificial deformation of the roughing layers.}
    \label{fig:collMillLimit}
\end{figure}

\revision{}{As discussed previously, when considering only inter-layer collisions, a trivial collision-free solution in the form of planar layers exists. However, in the context of milling, where collisions between the tool and the part itself must also be considered, a solution that is potentially collision-free (given appropriate tool length) may still not be classed as collision-free. This situation is illustrated in Fig.~\ref{fig:collMillLimit}. Consider a target part (green-shaded) obtained by removing excess material (blue-shaded). Fig.~\ref{fig:collMillLimit}(b) shows a planar layering configuration. When the tool is positioned at the red point, a collision with the part occurs. However, due to the finite radius of the milling tool, the material at that red point can instead be removed by positioning the tool at a different location (e.g., the green point), thereby avoiding intersection with the part geometry. In the current implementation, we do not verify whether tool positions that cause collisions are actually necessary for material removal. Consequently, to eliminate such collisions, the optimization process modifies the layers locally so that they become nearly parallel to the part surface in those regions. This results in undesirable curvature in the layers, as shown in Fig.~\ref{fig:collMillLimit}(c). Furthermore, the final collision-free configuration depends on the initialization, since the gradient-based optimization approach converges to local minima.}

\revision{}{Another concern related to the collision loss is the reliable detection of collisions involving thin or small-scale objects. Since the collision loss is computed over points sampled on the tool surface, a uniform sampling strategy may fail to capture interactions with objects whose dimensions fall below the sampling resolution. To mitigate this limitation, we introduce a perturbed sampling scheme, as detailed in \ref{appendix:tool_sampling}, which reduces the probability of missing such small objects. Nevertheless, we do not conduct extreme-case evaluations in this work, and a systematic assessment of these scenarios remains an important direction for future research.}


Furthermore, our exploration of the SIREN-based architecture has revealed promising avenues for deeper theoretical analysis. While we introduced a scaling-invariant frequency formulation, a more rigorous study is needed to understand how frequency, sampling interval, and model geometry interact with various optimization constraints. Moreover, it remains to be investigated whether controlling network parameters such as frequency should be treated as an explicit design choice or as part of the optimization objective itself.

\revision{}{As noted earlier, the SIREN network exhibits high representational capacity, which can lead to instability or convergence to suboptimal local minima in the absence of sufficient regularization. To mitigate this issue, we initialize the network layers to represent a planar configuration, thereby biasing optimization toward a nearby and well-behaved minimum. Furthermore, as illustrated in Fig.~\ref{fig:results_4c}, the inclusion of task-specific constraints in the form of loss functions helps guide the optimization toward solutions that satisfy the desired requirements. Future work may therefore focus on exploring additional regularization strategies or functional loss formulations to further reduce sensitivity to initialization and improve the robustness of the optimization process.}

Finally, although the implicit field formulation provides a continuous and differentiable representation over the entire domain, it lacks explicit topological connectivity information. Such information could be beneficial in cases involving multiple, closely spaced substructures, such as in the Spiral-Fish model, where limited local coupling may constrain the search space. Incorporating topological or graph-based priors into the implicit representation could therefore enhance the flexibility of the optimization and enable localized discontinuities when necessary.

In addition to these methodological improvements, optimizing the computational implementation of the current pipeline remains an important direction for future work, with the potential to significantly accelerate the optimization process.

Overall, the proposed framework constitutes a step forward in process planning for multi-axis manufacturing. It presents an implicit representation of the processes, parts and objectives, allowing collision handling, toolpath geometry, and layer generation to be co-optimized within a single differentiable system. To the best of our knowledge, this work represents the first approach that directly incorporates true collision avoidance within the implicit field generation process itself, rather than addressing it through post-processing or heuristic correction. Using a common representation for both additive and subtractive multi-axis processes can open up the possibility of integrating these workflows in the future. For optimization, we employ a stochastic gradient-based solver; since our representations and formulations are fully differentiable, the pipeline can be seamlessly integrated with other differentiable shape optimization or learning frameworks (e.g., \cite{park_deepsdf_2019,liu_neural_2025}) to provide complete control over the process from design to manufacturing.

%
%
%
%

\section*{Declaration of competing interest}
The authors declare that they have no known competing financial interests or personal relationships that could have appeared to influence the work reported in this paper.


\section*{Acknowledgments}
To be added upon the acceptance of the paper.


\appendix
\section{Derivation of Geodesic Curvature}
\label{appendix:geodesic}
\begin{flalign}
    \begin{split}
                &{\mathbf{t}} = \frac{\nabla f_l \times \nabla f_p}{\|\nabla f_l \times \nabla f_p\|} = \frac{\mathbf{T}}{\|\mathbf{T}\|} = \sum_{w=\{x,y,z\}}\frac{d w}{d s}\mathbf{\hat{w}}\\
       &{\vec{\kappa}_{curve}} = \frac{d{\mathbf{t}}}{ds}\\
        &{\vec{\kappa}_{curve}} = \sum_{w=\{x,y,z\}}\frac{\partial {\mathbf{t}}}{\partial w}\frac{d w}{d s}\\
        &\frac{d w}{d s} = \mathbf{{t}}_w, \text{($w$ component of the unit normal, $w$=$\{x,y,z\}$)}\\
        &\frac{\partial {\mathbf{t}}}{\partial w} = \frac{1}{\|\mathbf{T}\|}\frac{\partial \mathbf{T}}{\partial w} - \frac{\mathbf{T}}{\|\mathbf{T}\|^3}(\mathbf{T}\cdot\frac{\partial \mathbf{T}}{\partial w}),\text{where } w=\{x,y,z\}\\
        &\frac{\partial \mathbf{T}}{\partial w} = \frac{\partial\nabla f_l}{\partial w}\times \nabla f_p +  \nabla f_l \times \frac{\partial\nabla f_p}{\partial w},\text{where } w=\{x,y,z\}\\  
    \end{split}
\end{flalign}
and $\frac{\partial\nabla f_{k=\{l,p\}}}{\partial w}$ is the row/column of the corresponding Hessian matrix.
\\Now,
\begin{flalign}
    \begin{split}
        &{\kappa_{geodesic}} = \|{\vec{\kappa}_{curve}} - ({\vec{\kappa}_{curve}}\cdot \mathbf{{n}}_{f_l})\mathbf{{n}}_{f_l}\|
    \end{split}
\end{flalign}
We can now compute the Geodesic curvature using only the outputs of the field networks without re-computing new derivatives (backpropagation).

\section{Sampling The Tool}
\label{appendix:tool_sampling}
In Section~\ref{sec:implementation}, we discussed our implementation strategy for the collision avoidance scheme introduced in Section~\ref{subsec:collision}. Here, we provide the exact sampling strategy used in our tests to ensure reproducibility. The adjacent figure illustrates our printing tool, modeled as a combination of tool–enclosure units.

\begin{wrapfigure}{r}{0.40\textwidth}
\centering
\includegraphics[width=0.40\textwidth]{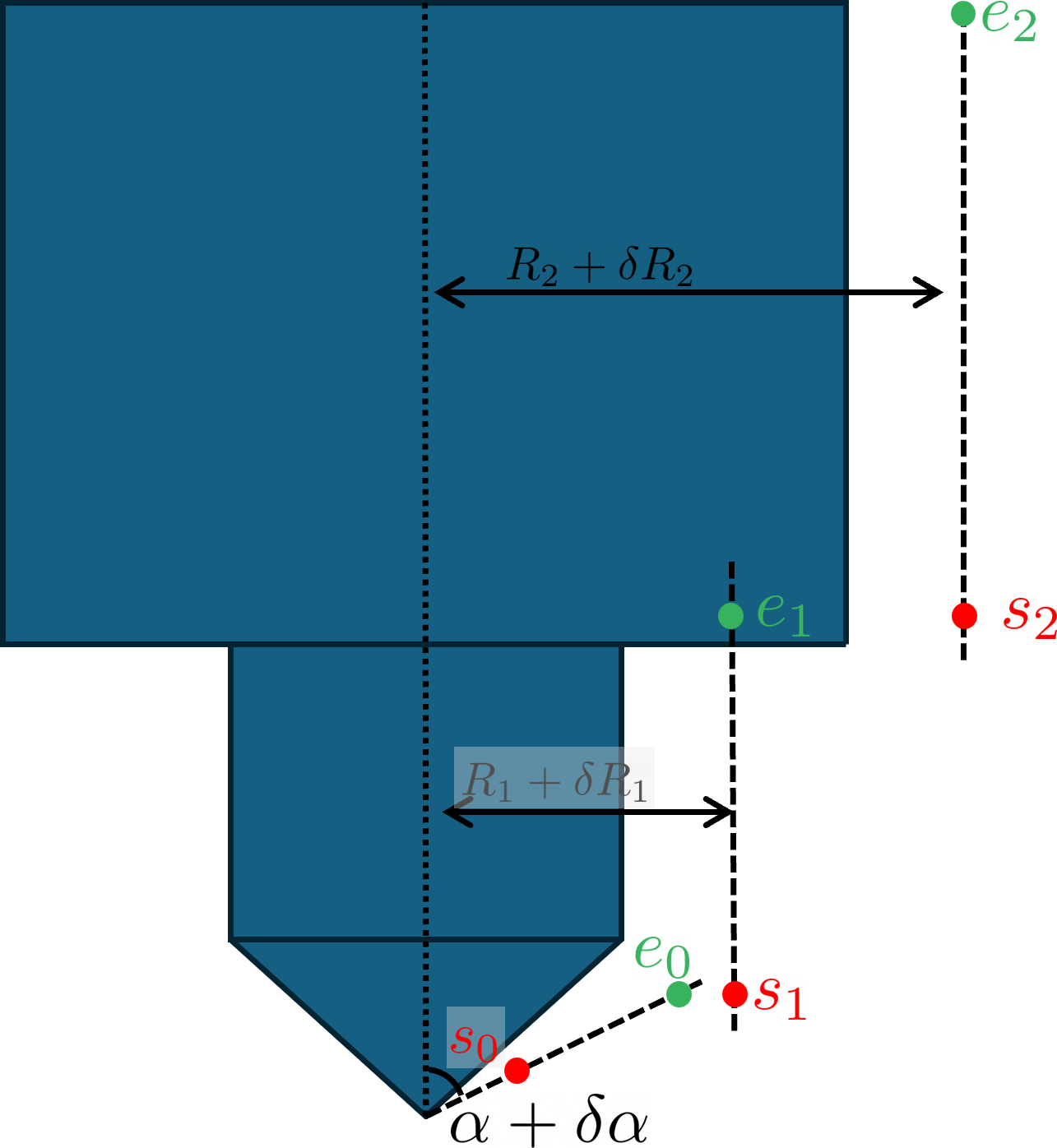}
\end{wrapfigure}

For our tool, we used $\alpha = 60^\circ$, $R_1 = 7.25,\text{mm}$, and $R_2 = 20,\text{mm}$. We obtained good results with $\delta({\text{length}}) = 4\text{mm}$ and $\delta({\text{angle}}) = 1^\circ$. As discussed earlier, the larger cylindrical unit ($R_2$) requires more points; to mitigate this, we employed a hybrid strategy. Axially, we started at $s_2 = 18.5,\text{mm}$ (measured from the base point) and sampled at fixed 5mm intervals up to $e_2 = 58.5\text{mm}$. At each axial level, we sampled 10 uniformly spaced points along the boundary. Additional sample points were then generated by placing new points between the existing ones, both axially and angularly, and perturbing them across iterations to increase coverage.

For the conical unit, we sampled along 20 directions distributed as follows: one along the axis, five uniformly spaced at half of the cone angle, and the remainder uniformly spaced around the full cone angle. Along each direction, points were sampled at 3 mm intervals from $s_0 = 3.75,\text{mm}$ to $e_0 = 12.75,\text{mm}$.

For the smaller cylindrical unit ($R_1$), sampling was carried out axially from $s_1 = 4.5 \text{mm}$ in steps of 3.75 mm. A small number of axial samples (five) were additionally placed in the region overlapping with the larger cylinder. Again, at each axial level, 10 boundary points were uniformly spaced. While the exact dimensions will change based on the tool, a similar procedure can be followed to generate the samples for other tools. For the milling tool, we applied a similar strategy but modeled it using only cylindrical units. Furthermore, depending on the available computational resources, the sampled points can be processed in batches of different sizes.


 \bibliographystyle{elsarticle-num} 
 \bibliography{references}






\end{document}